\documentclass{bmvc2k}

\vspace{-0.2em} 
\title{Why Multi-Source Audio Fails to Compose: \\ A Geometric Diagnosis and Its Remedy \vspace{-0.5em}}

\usepackage[table,xcdraw]{xcolor}
\usepackage{multirow}
\usepackage{url}
\usepackage{booktabs}
\usepackage{amsfonts}
\usepackage{nicefrac}
\usepackage{microtype}
\usepackage{amsmath,graphicx}
\usepackage{amssymb}
\usepackage{xspace}
\usepackage{enumitem}
\usepackage[accsupp]{axessibility}
\usepackage[ruled,vlined]{algorithm2e}
\usepackage{listings}
\usepackage{tikz}
\usepackage{subcaption}
\usetikzlibrary{arrows.meta,positioning,calc,shapes.geometric,fit,backgrounds}
\usepackage{caption}

\definecolor{bestcol}{HTML}{7B68EE}    %
\definecolor{ourrow}{HTML}{F2EFFF}     %

\addauthor{Anurag Singh}{singh.anurag@iitgn.ac.in}{1}
\addauthor{Prajwal Singh}{singh_prajwal@iitgn.ac.in}{1}
\addauthor{Shanmuganathan Raman}{shanmuga@iitgn.ac.in}{1}

\addinstitution{
 CVIG Lab, IIT Gandhinagar\\
 Gandhinagar, Gujarat, India
}
\runninghead{Singh \emph{et.al}}{Why Multi-Source Audio Fails to Compose}

\newcommand{\Lae}{\mathcal{L}_{\text{AE}}}
\newcommand{\Lastar}{\mathcal{L}_{\text{A}^{*}}}
\newcommand{\Lmix}{\mathcal{L}_{\text{mix}}}
\newcommand{\Lseg}{\mathcal{L}_{\mathrm{seg}}}
\newcommand{\Lret}{\mathcal{L}_{\mathrm{ret}}}

\begin{document}

\maketitle

\vspace{-2em}

\begin{abstract}
Multi-source audio-to-image generation exhibits consistent compositional failures when audio tokens occupy geometrically displaced, semantically unanchored positions in the CLIP text-token space, where the diffusion U-Net lacks the prior needed to attend to them reliably or ground them in distinct spatial regions. We formalize this as the \textit{geometric displacement} phenomenon and make it explicit through a two-level clustering analysis comparing BEATs and CLAP embeddings at both the raw encoder level and the projected CLIP token level. Three geometric properties emerge as important: intra-class compactness, angular alignment with the CLIP text manifold, and inter-token spatial separation. We show that CLAP, whose contrastive language-audio pretraining aligns embeddings with natural language semantics, substantially outperforms BEATs across all three properties, improving the silhouette score by more than 3× and the Audio-Image Content score as well. Building on this geometric analysis, we introduce a multi-token extension that requires no diffusion-backbone retraining, assigns each audio source to a distinct placeholder token in the CLIP tokenizer, and complements it with a mixed inference-time objective that combines Attend-and-Excite with an A$^\ast$ spatial separation loss to resolve residual cross-token interference during denoising. To support evaluation, we construct compositional benchmarks from VGGSound with multiple input audios, VGG2, VGG3, and VGGN, and extend six standard audio-image metrics to the multi-source setting. Across five benchmarks and against four strong baselines, our method achieves strong results on the metrics most directly tied to compositional faithfulness, with AIC improvements of up to 2.25$\times$ over the strongest baseline and FID reduced by nearly 2$\times$ on two-source generation. We propose geometric displacement as a diagnostic framework for cross-modal generation systems that project learned embeddings into a pretrained contrastive space.
\end{abstract}

\begin{figure}[!t]
\centering
\includegraphics[width=0.92\linewidth]{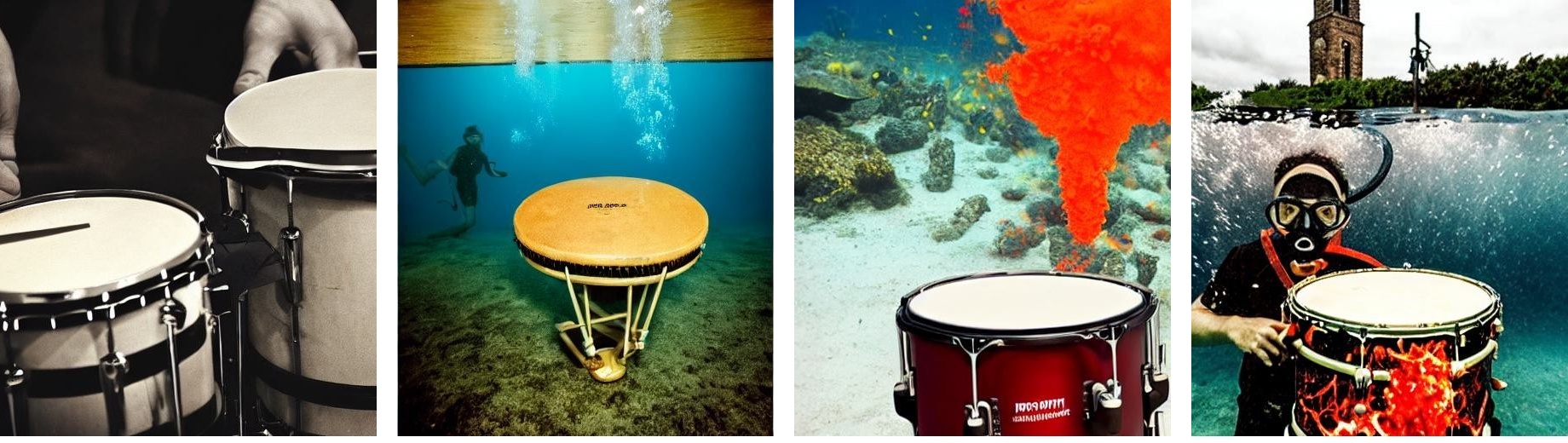}\vspace{-0.5mm}\\
\noindent{\scriptsize\itshape%
  \makebox[0.25\linewidth][c]{snare drum}%
  \makebox[0.25\linewidth][c]{+ underwater}%
  \makebox[0.25\linewidth][c]{+ volcano}%
  \makebox[0.25\linewidth][c]{+ church bell}\par}
\vspace{4pt}
\includegraphics[width=0.92\linewidth]{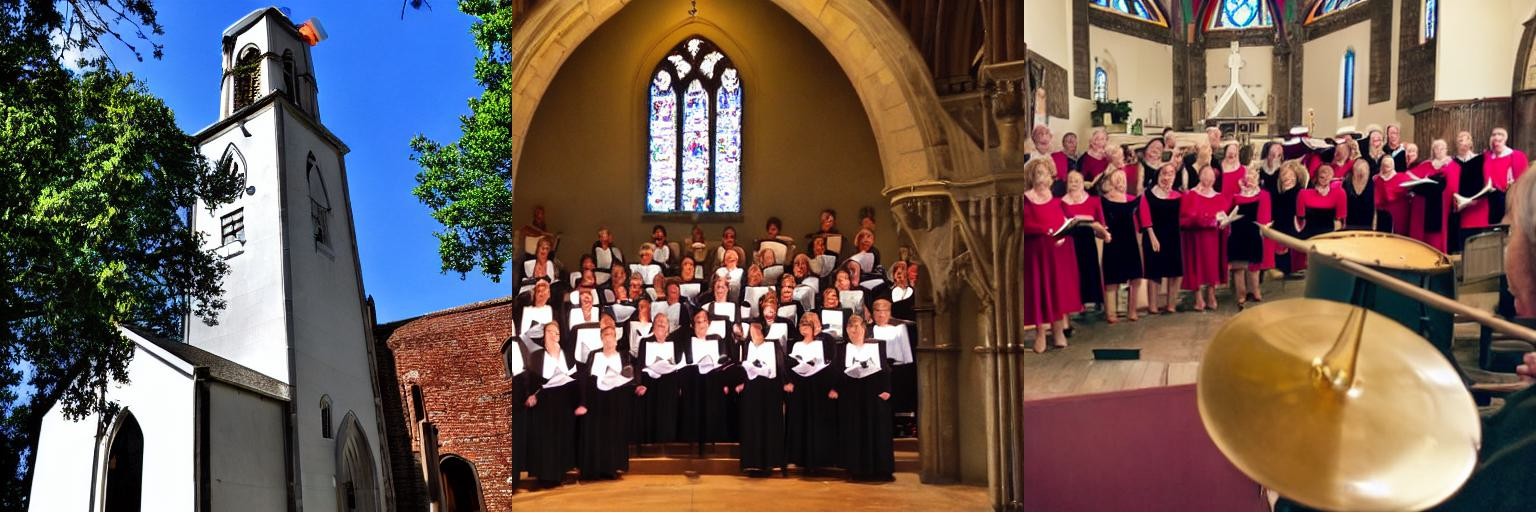}\\[1pt]
\noindent{\scriptsize\itshape%
  \makebox[0.330\linewidth][c]{church bell}%
  \makebox[0.330\linewidth][c]{+ singing choir}%
  \makebox[0.330\linewidth][c]{+ drum kit}\par}\vspace{-1mm}
\caption{\textbf{Compositional audio-to-image generation with our method.} The \textit{top row} progressively adds four audio sources (\textit{snare drum, underwater bubbling, volcano explosion, church bell ringing}), while the \textit{bottom row} shows a three-source example (\textit{church bell, singing choir, drum kit}). Each new object is added to a separate, spatially consistent region while preserving earlier concepts. Our analysis suggests that this behavior is strongly influenced by the geometry of audio tokens in CLIP space, which we quantify in Section~\ref{sec:clap}, use to redesign the embedder in Section~\ref{sec:method}, and further refine through our mixed inference-time objective.}
\label{fig:teaser}
\vspace{-1.5em}
\end{figure}

\section{Introduction}
\label{sec:intro}

Sound in real-world environments is naturally multi-source. A street scene may contain traffic, conversation, and birdsong, while a kitchen scene may include running water, footsteps, and appliance noise. Despite recent progress in audio-conditioned image synthesis~\cite{audiotoken,sonicdiffusion,sound2scene}, most existing methods operate on only a \emph{single} audio clip and therefore generate single-object scenes. A natural extension is to generate a coherent image from multiple audio inputs, where each sound source is represented in the final scene. We refer to this task as \textit{compositional audio-to-image generation} (Figure~\ref{fig:teaser}).

\begin{figure}[!t]
\centering
\setlength{\tabcolsep}{1pt}
\renewcommand{\arraystretch}{1.05}
\begin{tabular}{cccc}
\textbf{\scriptsize AT} & \textbf{\scriptsize AT + A$^{*}$} &
\textbf{\scriptsize AT + AnE} & \textbf{\scriptsize AT + AnE + A$^{*}$ \emph{(ours)}} \\
\includegraphics[width=0.245\linewidth]{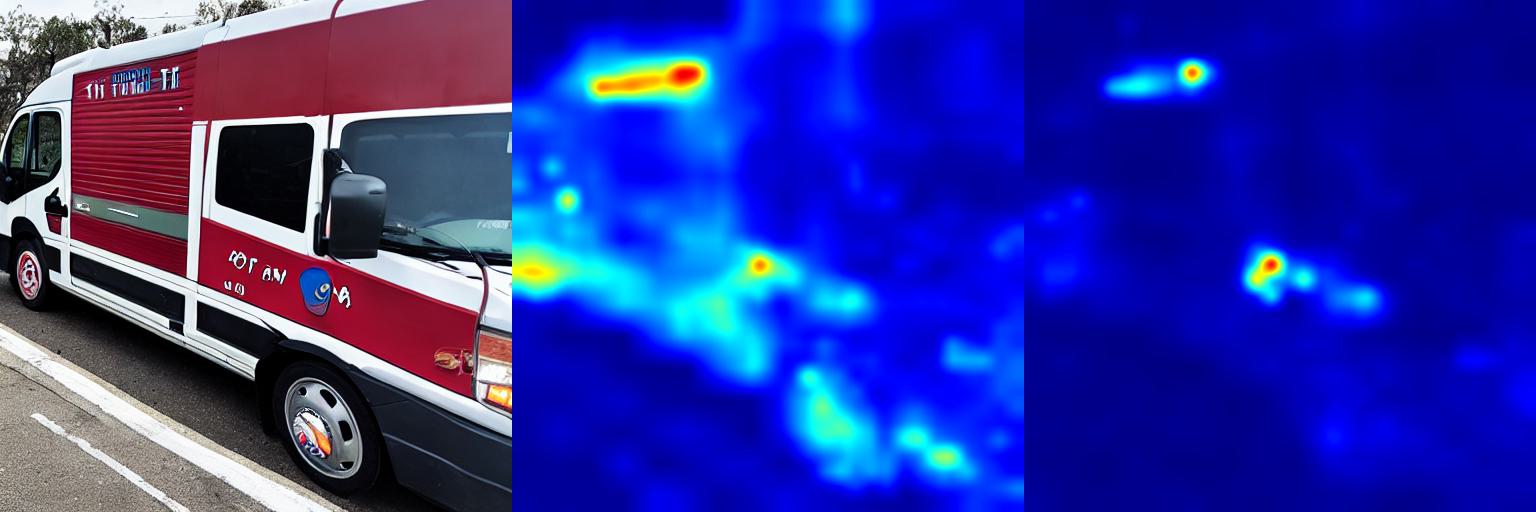} &
\includegraphics[width=0.245\linewidth]{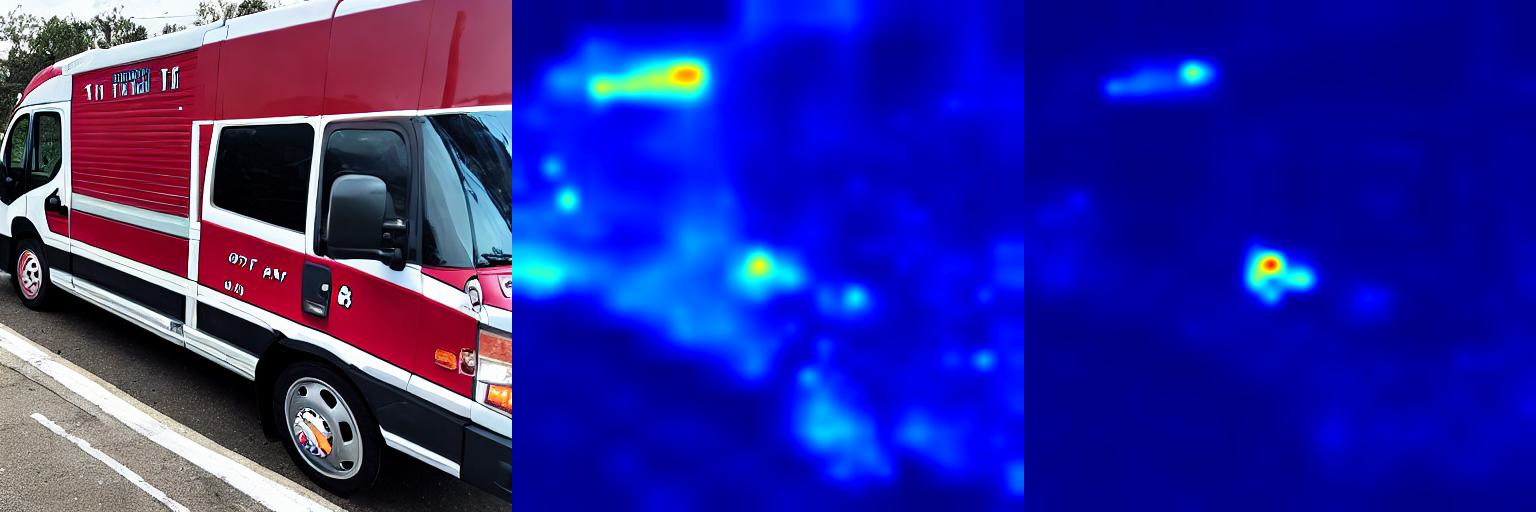} &
\includegraphics[width=0.245\linewidth]{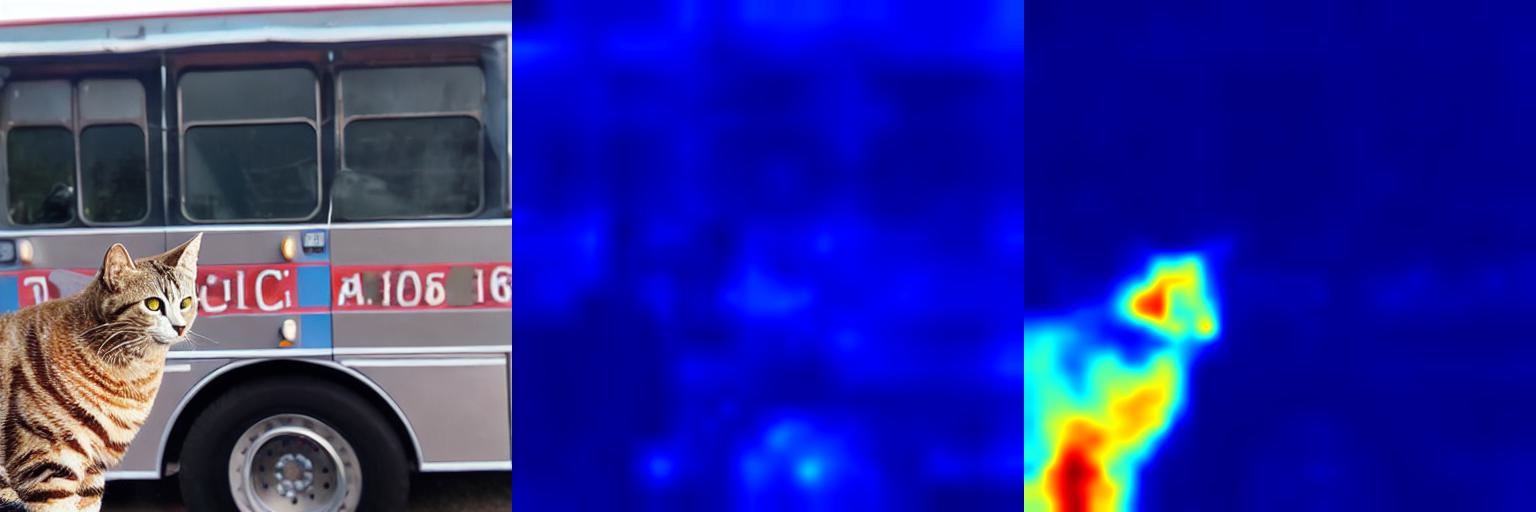} &
\includegraphics[width=0.245\linewidth]{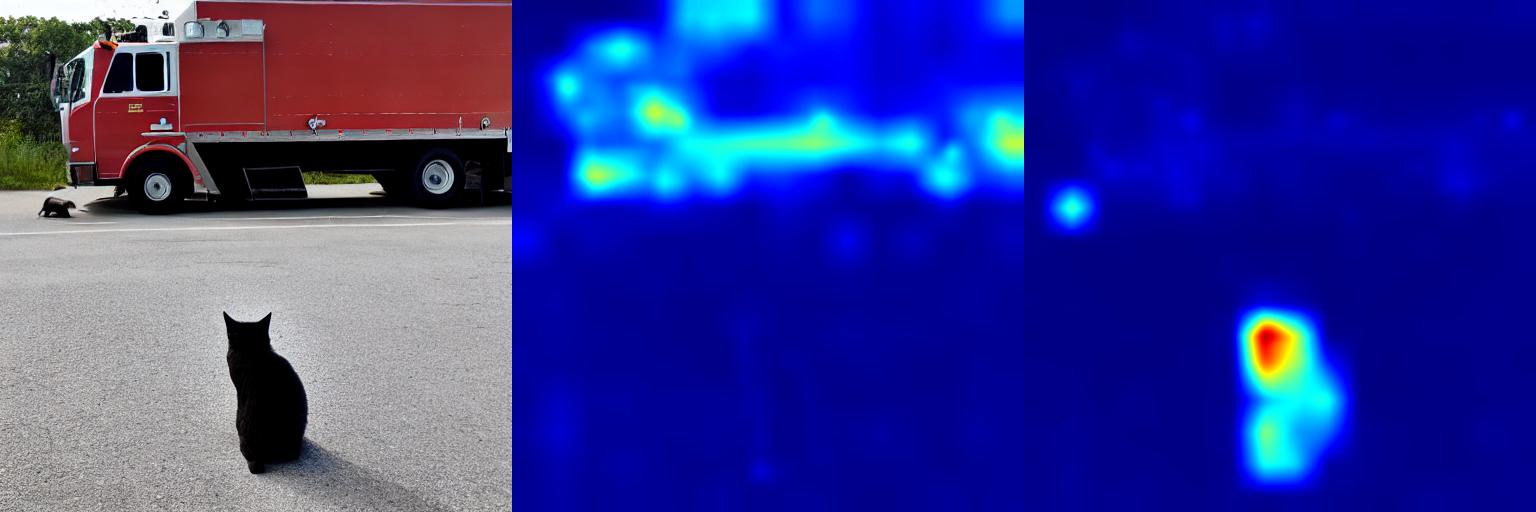} \\
\end{tabular}\vspace{-0.5mm}
\caption{\textbf{Cross-attention failure modes in compositional generation.} Final denoising step ($t{=}50$) for the prompt \textit{``a photo of a $\langle\!*\!\rangle$ and a $\langle\hat{\;}\rangle$''}, where $\langle\!*\!\rangle$ represents \textit{fire truck siren} and $\langle\hat{\;}\rangle$ represents \textit{cat purring}. The retrained AT (column 1) produces overlapping attention maps. A$^{*}$ alone (column 2) separates the attention regions but weakens activation. Attend-and-Excite (column 3) increases token activation but still produces overlap. The mixed objective (column 4) generates strong, spatially separated attention maps.}
\label{fig:concept}
\vspace{-1.5em}
\end{figure}

\vspace{1mm} \noindent
Extending single-audio methods directly to multiple audio sources results in consistent failure modes. Concatenating multiple audio embeddings into a single conditioning token collapses the sources into an averaged representation. Using multiple audio-derived tokens in the prompt introduces a different problem: the diffusion model often ignores all but one token, generating only one object or merging multiple objects into a semantically unclear result. This behavior is related to the \emph{catastrophic neglect} problem observed in text-conditioned diffusion models~\cite{attendexcite}. Figure~\ref{fig:concept} illustrates this behavior in AudioToken~\cite{audiotoken}, where the cross-attention maps of two audio tokens collapse into the same spatial region during denoising. Our mixed objective explicitly separates these attention maps.

\vspace{1mm} \noindent
Our analysis suggests that these failures are strongly influenced by the geometry of the conditioning space. In diffusion models, cross-attention computes similarities between image features and conditioning tokens, which are typically CLIP text embeddings drawn from a well-structured semantic space~\cite{clip}. When audio-derived tokens are inserted into this space, three geometric properties become important: i) \textbf{Intra-class compactness.} Audio embeddings from the same semantic class should form compact clusters. Otherwise, the diffusion model receives inconsistent conditioning signals across samples. ii) \textbf{Angular alignment with the CLIP text manifold.} Cross-attention depends more strongly on cosine similarity than Euclidean distance. Tokens that preserve angular alignment with the CLIP text space can still behave like text tokens even if their absolute magnitude differs. iii) \textbf{Spatial separation of attention maps.} When multiple audio tokens are present, each should activate strongly, and their cross-attention maps should occupy different image regions. Weak activation causes neglect, while overlapping maps cause object fusion (Figure~\ref{fig:concept}).

\vspace{1mm} \noindent
These properties can be directly measured and used to guide model design. We show that the BEATs encoder used in prior work~\cite{audiotoken} yields weak cluster geometry, with low intra-class compactness and poor separation in both the raw embedding space and the projected CLIP token space. In contrast, CLAP~\cite{clap} produces more compact and better-aligned embeddings, improving silhouette score by more than $3\times$. More importantly, CLAP embeddings preserve angular alignment with the CLIP text manifold, a key property for stable cross-attention. Based on this geometric view, we make the following contributions: \textbf{i)} We provide a geometric analysis of how existing audio-to-image models fail in the compositional setting, using clustering in both the raw audio embedding space and the projected CLIP token space. \textbf{ii)} We introduce a CLAP-based embedder redesigned around CLAP's global $512$-dimensional representation, producing significantly improved cluster geometry compared to BEATs. \textbf{iii)} We propose a multi-token extension (MTE) that assigns each audio source to a separate placeholder token, enabling compositional conditioning without retraining the diffusion backbone. \textbf{iv)} We introduce a mixed inference-time objective that combines Attend-and-Excite with the A$^{*}$ spatial loss, jointly enforcing strong token activation and spatial separation of attention maps. \textbf{v)} We construct three new compositional benchmarks, VGG2, VGG3 and VGGn, together with multi-source extensions of standard audio-to-image evaluation metrics.

\vspace{1mm} \noindent
Across five benchmarks (VGG, IS3, VGG2, VGG3, and VGGn) and against four strong baselines (AudioToken, SonicDiffusion, Sound2Scene, and MACS), our variants achieve the best performance on the metrics most closely related to compositional generation, including AIC, IIS, and FID. The CLAP variant is strongest in distribution, improving AIC by up to $2.25\times$, while the BEATs variant leads on the out-of-distribution IS3 benchmark.

\vspace{1mm} \noindent
Our method operates on separate audio sources, which naturally align with settings such as instrument recordings, studio audio tracks, or labeled datasets. When only mixed audio is available, any source separation method such as Demucs~\cite{demucs} can be used as a preprocessing step. Our framework is independent of the specific source separation method used.
\section{Related Work}
\label{sec:related}

\paragraph{Audio-to-image generation.} Early audio-to-image (A2I) methods were mainly based on conditional GANs. Sound2Scene~\cite{sound2scene} trains a conditional GAN on a self-supervised image latent space and then aligns an audio encoder with that space. More recent diffusion-based approaches achieve much higher image quality. SonicDiffusion~\cite{sonicdiffusion} maps CLAP audio features to tokens and injects them into a frozen Stable Diffusion U-Net via gated audio-image cross-attention layers. AudioToken~\cite{audiotoken}, which our work builds upon, formulates A2I generation as textual inversion~\cite{textualinversion}, projecting BEATs~\cite{beats} audio features into a learned token inserted into the CLIP text prompt of a frozen Stable Diffusion model~\cite{ldm}. The most closely related prior work is MACS~\cite{macs}, which tackles multi-source A2I by first separating mixed audio using a weakly supervised CLAP-aligned module and a ranking loss, and then mapping each separated signal to the generator via a lightweight adapter. In contrast, we assume separate audio streams and instead address the geometry of audio tokens in CLIP space, thereby repairing cross-attention behavior directly within the denoising loop.

\vspace{1mm} \noindent
\textbf{Compositional text-to-image generation.} Diffusion models are known to suffer from \emph{catastrophic neglect}~\cite{attendexcite}, where one or more subjects in a multi-object prompt are ignored during generation. Attend-and-Excite~\cite{attendexcite} addresses this issue by monitoring the cross-attention activation of subject tokens during denoising and updating the latent representation when activations become weak. Structured Diffusion Guidance~\cite{structured} decomposes prompts into separate noun phrases routed through different attention pathways. Other approaches, such as A$^{*}$~\cite{astar} and BoxDiff~\cite{boxdiff}, introduce spatial constraints that reduce overlap between attention maps or enforce bounding-box consistency. Our work extends this line of research to the audio-conditioned setting. Audio-derived tokens can occupy a different geometric region in CLIP space from natural text tokens, which can make compositional conditioning more challenging (Section~\ref{sec:clap}).

\vspace{1mm} \noindent
\textbf{Audio representation learning.} The effectiveness of audio-conditioned generation can depend strongly on the geometry of the underlying audio embeddings. BEATs~\cite{beats} is a self-supervised audio transformer trained using masked acoustic-token prediction and produces frame-level embeddings of dimension $768$. In contrast, CLAP~\cite{clap} is a contrastive language-audio model trained on approximately $633$K audio-text pairs, producing a global $512$-dimensional embedding aligned with natural language semantics. We use CLAP because its embeddings preserve important geometric properties such as compact clustering and angular separation, which remain stable after projection into the CLIP text-token space and improve cross-attention behavior.

\vspace{1mm} \noindent
\textbf{Audio source separation.} Audio source separation~\cite{demucs} is a separate and well-studied problem. Our method assumes that individual audio sources are already separated and does not rely on any specific source-separation model. When only mixed audio is available, any existing separation method can be used as a preprocessing step.
\section{Method}
\label{sec:method}

\begin{figure}[!t]
\centering

\definecolor{frozenFill}{HTML}{DCE7F4}   %
\definecolor{frozenEdge}{HTML}{6B89B2}
\definecolor{trainFill}{HTML}{FCE3C2}    %
\definecolor{trainEdge}{HTML}{C58A3C}
\definecolor{tokenFill}{HTML}{E4DCEF}    %
\definecolor{tokenEdge}{HTML}{7E6BA0}
\definecolor{promptFill}{HTML}{FAF1D6}   %
\definecolor{promptEdge}{HTML}{B89B4B}
\definecolor{outFill}{HTML}{ECECEC}      %
\definecolor{outEdge}{HTML}{888888}
\definecolor{lossFillA}{HTML}{F4D6D2}    %
\definecolor{lossFillB}{HTML}{E9B4AD}    %
\definecolor{lossEdge}{HTML}{B45A4F}

\resizebox{\linewidth}{!}{%
\begin{tikzpicture}[
  font=\small,
  >=Latex,
  every node/.style={align=center},
  encoder/.style={draw=frozenEdge,line width=0.5pt,rounded corners=2pt,fill=frozenFill,
                  minimum width=1.4cm,minimum height=0.95cm,inner sep=2pt},
  embed/.style={draw=trainEdge,line width=0.6pt,rounded corners=2pt,fill=trainFill,
                minimum width=1.6cm,minimum height=0.95cm,inner sep=2pt},
  token/.style={draw=tokenEdge,line width=0.5pt,circle,fill=tokenFill,
                minimum size=0.55cm,inner sep=0pt},
  unet/.style={draw=frozenEdge,line width=0.6pt,rounded corners=4pt,fill=frozenFill,
               minimum width=2.4cm,minimum height=1.5cm,inner sep=2pt},
  prompt/.style={draw=promptEdge,line width=0.5pt,rounded corners=2pt,fill=promptFill,
                 minimum width=3.0cm,minimum height=2.6cm,inner sep=4pt},
  outbox/.style={draw=outEdge,line width=0.5pt,rounded corners=2pt,fill=outFill,
                 minimum width=1.8cm,minimum height=1.5cm,inner sep=3pt},
  loss/.style={draw=lossEdge,line width=0.5pt,rounded corners=2pt,
               minimum width=1.7cm,minimum height=0.8cm,inner sep=2pt},
  arr/.style={->,thick,gray!40!black},
  lossarr/.style={->,thick,dashed,lossEdge},
  dimlbl/.style={font=\scriptsize\itshape,gray!55!black},
  vd/.style={font=\small,gray!55!black,inner sep=1pt}
]

\node (a1) at (0, 1.4)  {\scriptsize $a_1$};
\node (a2) at (0, 0.0)  {\scriptsize $a_2$};
\node[vd] at (0,-1.0) {$\vdots$};
\node (an) at (0,-2.0) {\scriptsize $a_n$};

\node[encoder] (clap1) at (1.8, 1.4) {CLAP\\ \scriptsize encoder};
\node[encoder] (clap2) at (1.8, 0.0) {CLAP\\ \scriptsize encoder};
\node[vd] at (1.8,-1.0) {$\vdots$};
\node[encoder] (clapn) at (1.8,-2.0) {CLAP\\ \scriptsize encoder};
\node[dimlbl] at (1.8,-2.85) {shared, frozen};

\node[embed] (emb1) at (4.1, 1.4) {Embedder $E_{\theta}$\\ \scriptsize Eq.~(\ref{eq:embedder})};
\node[embed] (emb2) at (4.1, 0.0) {Embedder $E_{\theta}$};
\node[vd] at (4.1,-1.0) {$\vdots$};
\node[embed] (embn) at (4.1,-2.0) {Embedder $E_{\theta}$};
\node[dimlbl] at (4.1,-2.85) {shared, $\sim$8.3M params};

\node[token] (t1) at (6.1, 1.4) {$e^{(1)}$};
\node[token] (t2) at (6.1, 0.0) {$e^{(2)}$};
\node[vd] at (6.1,-1.0) {$\vdots$};
\node[token] (tn) at (6.1,-2.0) {$e^{(n)}$};

\node[prompt] (mte) at (8.2,0.0)
  {\scriptsize \textbf{Multi-Token}\\\scriptsize \textbf{Prompt}\\\scriptsize (Eq.~\ref{eq:mte})\\[3pt]
   \scriptsize\textit{``a photo of a}\\\scriptsize\textit{$\langle\!*\!\rangle$ and a $\langle\hat{\;}\rangle$''}};

\node[unet] (unet) at (11.4,0.0)
  {\textbf{Stable Diffusion}\\\scriptsize U-Net (frozen)\\\scriptsize cross-attn};

\node[loss,fill=lossFillA] (lae)  at (11.4,2.05) {$\Lae$ \scriptsize Eq.~(\ref{eq:lae})\\\tiny activation};
\node[loss,fill=lossFillB] (last) at (11.4,3.1)  {$\Lastar$ \scriptsize Eq.~(\ref{eq:lastar})\\\tiny separation};
\node[draw=lossEdge,dashed,rounded corners=3pt,inner sep=3pt,fit=(lae)(last),
      label={[font=\scriptsize\bfseries,text=lossEdge]above:Mixed objective $\Lmix$ \scriptsize(Eq.~\ref{eq:lmix})}] {};

\node[outbox] (img) at (14.1,0.0) {\scriptsize generated\\ \scriptsize image $I$};

\draw[arr] (a1) -- (clap1);
\draw[arr] (a2) -- (clap2);
\draw[arr] (an) -- (clapn);
\draw[arr] (clap1) -- node[above,dimlbl]{\scriptsize 512} (emb1);
\draw[arr] (clap2) -- node[above,dimlbl]{\scriptsize 512} (emb2);
\draw[arr] (clapn) -- node[above,dimlbl]{\scriptsize 512} (embn);
\draw[arr] (emb1) -- node[above,dimlbl]{\scriptsize 768} (t1);
\draw[arr] (emb2) -- node[above,dimlbl]{\scriptsize 768} (t2);
\draw[arr] (embn) -- node[above,dimlbl]{\scriptsize 768} (tn);
\draw[arr] (t1) to[bend left=10]  (mte.west);
\draw[arr] (t2) --                (mte.west);
\draw[arr] (tn) to[bend right=10] (mte.west);
\draw[arr] (mte.east) -- (unet.west);
\draw[lossarr] (lae)  edge[bend right=10] (unet.north);
\draw[lossarr] (last) edge[bend right=20] (unet.north);
\draw[arr] (unet.east) -- (img.west);

\node[font=\scriptsize,anchor=west] at (0.0,-3.5) {%
  \tikz\node[draw=frozenEdge,fill=frozenFill,rounded corners=1pt,
             minimum width=0.32cm,minimum height=0.20cm,inner sep=0pt]{};\,frozen \quad
  \tikz\node[draw=trainEdge,fill=trainFill,rounded corners=1pt,
             minimum width=0.32cm,minimum height=0.20cm,inner sep=0pt]{};\,trained \quad
  \tikz\node[draw=tokenEdge,fill=tokenFill,circle,
             minimum size=0.22cm,inner sep=0pt]{};\,audio token \quad
  \tikz\draw[lossarr](0,0)--(0.5,0);\,inference-time gradient on the latent};

\end{tikzpicture}}
\caption{\textbf{Method overview.} Each audio clip $a_{i}$ is encoded using a frozen CLAP~\cite{clap} encoder to obtain a global $512$-dimensional embedding. A shared embedder $E_{\theta}$ projects this embedding into a $768$-dimensional CLIP token $e^{(i)}$ (Eq.~\ref{eq:embedder}), which is assigned to a unique placeholder token in the multi-token prompt (Eq.~\ref{eq:mte}). The prompt is then used to condition the frozen Stable Diffusion U-Net. During denoising, two inference-time losses improve cross-attention behavior: $\Lae$ (Eq.~\ref{eq:lae}) increases token activations, while $\Lastar$ (Eq.~\ref{eq:lastar}) encourages spatial separation between tokens. Their combined objective $\Lmix$ (Eq.~\ref{eq:lmix}) addresses the failure modes shown in Figure~\ref{fig:concept}. The embedder is the only trainable component in the framework.}
\label{fig:method}
\vspace{-1.0em}
\end{figure}
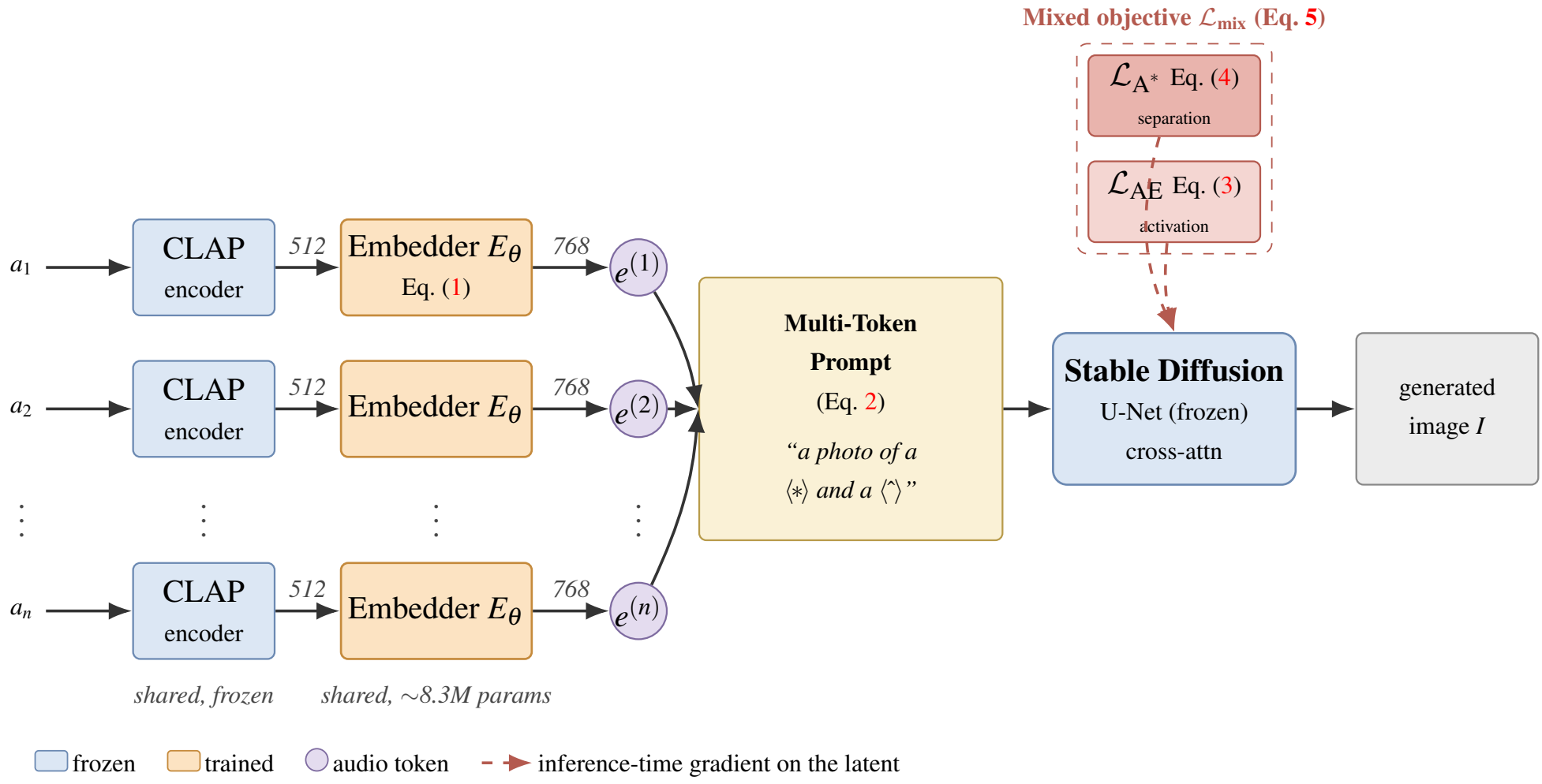

\subsection{Problem Setting}
Let $\mathcal{A}=\{a_{1},\dots,a_{n}\}$ be a set of $n$ audio waveforms, where each $a_{i}\in\mathbb{R}^{T}$ and all inputs are standardised to $T=10\text{s}$. Let $p$ denote a prompt template such as ``\textit{a photo of a $\langle\!*\!\rangle$ and a $\langle\hat{\;}\rangle$ and a $\langle!\rangle$\,\dots}'', containing $n$ placeholder tokens corresponding to the $n$ audio sources. The goal of compositional audio-to-image generation is to learn a function $f:\mathcal{A}\times p\to\mathcal{I}$ that produces an image $I\in\mathbb{R}^{H\times W\times 3}$ that is faithful to every audio source. We formalize this through three desiderata. The first is \textit{audio fidelity} (C1), which requires the object generated for each source $a_{i}$ to match its semantic content. The second is \textit{completeness} (C2), which requires all $n$ sources to be visibly represented in the image. The third is \textit{spatial separation} (C3), which requires the generated objects to occupy distinct, non-overlapping regions. Our method addresses these three desiderata through three corresponding geometric mechanisms, summarised in Figure~\ref{fig:method}.

\subsection{Background}

AudioToken~\cite{audiotoken} performs single-audio synthesis through a textual inversion pipeline: a frozen encoder $\phi(\cdot)$ produces an acoustic feature sequence $\phi(a)$, a small embedder (attention pooling followed by an MLP, $\sim$7M parameters) projects this sequence to a single embedding $e_{\text{audio}}\in\mathbb{R}^{768}$ in the CLIP text-token space, and a frozen Stable Diffusion backbone is conditioned on a prompt with $e_{\text{audio}}$ in the subject slot. The embedder is trained with the standard
diffusion denoising loss while keeping both $\phi$ and the U-Net frozen.

\subsection{The Geometry of Audio Tokens in CLIP Space}
\label{sec:clap}

Before describing our method, we first present the main observation that motivates it. In the diffusion backbone, all conditioning tokens are processed identically. Cross-attention is computed using inner products between visual queries and conditioning keys. As a result, whether an audio token is attended to depends on where that token lies in the CLIP text-token space, relative to both (i) other tokens from the same semantic class and (ii) the natural text embedding manifold on which the diffusion model was trained. We study these relationships at two levels.

\vspace{1mm}\noindent
\underline{Level 1: Raw Audio Embeddings.}
We first extract embeddings from a $50$-class subset of VGGSound (Section~\ref{sec:training}) using BEATs and CLAP. We then evaluate the $\ell_{2}$-normalized embeddings using standard clustering metrics: \emph{within-cluster variance} (WCV, lower is better), \emph{Davies--Bouldin index} (DB, lower is better), \emph{silhouette score} (higher is better), and \emph{mean intra-cluster cosine similarity} (higher is better). 

As shown in Table~\ref{tab:clustering}, CLAP produces significantly better cluster structure than BEATs. The silhouette score improves by more than $3\times$ ($0.0295$ vs.\ $0.0090$), DB decreases by $21\%$, and WCV decreases from $0.545$ to $0.470$. These results indicate that BEATs places audio clips from the same class in less consistent geometric locations, which can make the downstream embedding mapping more difficult to learn.

\vspace{1mm}\noindent
\underline{Level 2: Projected Tokens in CLIP Space.}
Next, we pass the same embeddings through the trained embedder and analyze the resulting projected tokens in CLIP space (on the right side of Table~\ref{tab:clustering}). The advantage of CLAP becomes even stronger after projection. The silhouette score improves from $0.16$ (BEATs) to $0.35$ (CLAP), while DB decreases from $2.49$ to $1.28$.

\begin{figure}[!t]
\centering
\includegraphics[width=0.55\linewidth]{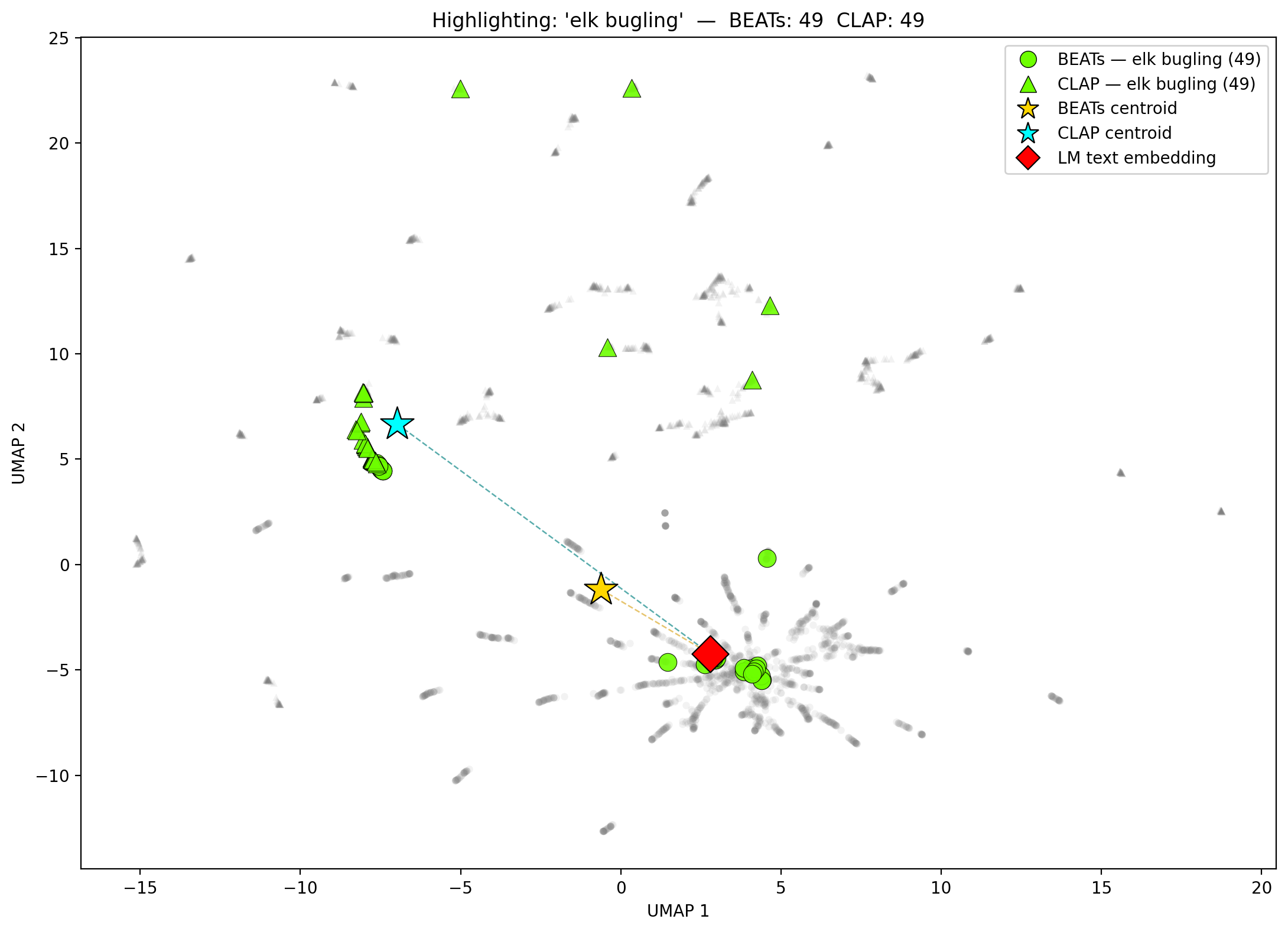}
\caption{\textbf{UMAP visualization of audio tokens in CLIP space.} For \emph{Elk Bugling} (additional classes are shown in supplementary Figure~$S9$), circles and triangles represent BEATs and CLAP tokens, respectively. Colored points denote the highlighted class, while grey points denote the remaining $49$ classes. Gold and cyan stars show the BEATs and CLAP class centroids, respectively, and the red diamond shows the CLIP text embedding; dashed lines connect each centroid to the text embedding. CLAP tokens form a tighter, more separated cluster, consistent with Table~\ref{tab:clustering}, while its centroid is farther from the text embedding in L2 distance but similarly aligned in direction to BEATs, which is more important for cross-attention.}
\label{fig:umap_main}
\vspace{-1.0em}
\end{figure}

We further measure the distance between each class centroid and the CLIP text embedding of the corresponding class label. Two observations are important. First, CLAP tokens are farther from the text embeddings in $\ell_{2}$ distance ($0.825$ vs.\ $0.298$), which may initially suggest weaker alignment. However, the cosine distance remains nearly identical ($0.639$ vs.\ $0.661$). Since cross-attention depends on normalized inner products, angular alignment is more important than Euclidean distance. This suggests that CLAP tokens retain comparable angular alignment with the text manifold while also being more clearly separated from each other. Figure~\ref{fig:umap_main} visualizes this behavior.

\vspace{1mm}\noindent
This geometric property offers a useful explanation for the consistent advantage of CLAP over BEATs in our experiments. More generally, it suggests that an audio encoder for diffusion conditioning does not need to exactly match text embeddings in $\ell_{2}$ distance. Instead, it should produce class-consistent tokens that preserve angular alignment with the text embedding.

\begin{table}[!t]
\centering
\small
\setlength{\tabcolsep}{4pt}
\vspace{0.5em}
\begin{minipage}[t]{0.485\columnwidth}
\centering
\resizebox{0.7\linewidth}{!}{
\def\arraystretch{1.1}
\begin{tabular}{lcc}
\toprule
\textbf{Raw embeddings} & \textbf{CLAP} & \textbf{BEATs} \\
\midrule
\multicolumn{3}{l}{\textit{Global}} \\
Silhouette $\uparrow$            & \textcolor{bestcol}{$\mathbf{0.0295}$} & $0.0090$ \\
Davies--Bouldin $\downarrow$     & \textcolor{bestcol}{$\mathbf{3.9079}$} & $4.9387$ \\
\midrule
\multicolumn{3}{l}{\textit{Per-class (mean $\pm$ std)}} \\
WCV $\downarrow$                 & \textcolor{bestcol}{$\mathbf{0.470}$} & $0.545$ \\
Intra cos.\ sim.\ $\uparrow$     & \textcolor{bestcol}{$\mathbf{0.520}$} & $0.450$ \\
Silhouette $\uparrow$            & \textcolor{bestcol}{$\mathbf{0.020}$} & $0.005$ \\
\bottomrule
\end{tabular}}
\end{minipage}
\hfill
\begin{minipage}[t]{0.485\columnwidth}
\centering
\resizebox{0.7\linewidth}{!}{
\def\arraystretch{1.1}
\begin{tabular}{lcc}
\toprule
\textbf{Projected tokens} & \textbf{CLAP} & \textbf{BEATs} \\
\midrule
\multicolumn{3}{l}{\textit{Global}} \\
Silhouette $\uparrow$            & \textcolor{bestcol}{$\mathbf{0.3483}$} & $0.1609$ \\
Davies--Bouldin $\downarrow$     & \textcolor{bestcol}{$\mathbf{1.2840}$} & $2.4922$ \\
\midrule
\multicolumn{3}{l}{\textit{Centroid $\to$ text emb.}} \\
L2 dist.\ (informative only)     & $0.8249$ & \textcolor{bestcol}{$\mathbf{0.2975}$} \\
Cos.\ dist.\ $\downarrow$        & \textcolor{bestcol}{$\mathbf{0.6392}$} & $0.6611$ \\
\bottomrule
\end{tabular}}
\end{minipage}
\vspace{0.3em}
\caption{\textbf{Cluster geometry of raw embeddings} (left) \textbf{and projected CLIP tokens} (right) on the $50$-class VGGSound subset. $\uparrow$/$\downarrow$ indicate higher/lower is better, and \textcolor{bestcol}{violet} marks the better result in each row. CLAP consistently produces more compact clusters and higher silhouette scores than BEATs. Although its centroids are farther from text embeddings in L2 distance, CLAP preserves cosine alignment, which is more relevant for cross-attention. These geometric improvements are preserved and further strengthened after projection into CLIP space.}\vspace{-1.2em}
\label{tab:clustering}
\end{table}

\subsection{CLAP-Based Embedder}
\label{sec:embedder}

CLAP produces a single global $512$-dimensional embedding for each audio clip, unlike BEATs, which outputs a frame-level feature sequence. As a result, the original BEATs-based embedder design (attention pooling followed by an MLP) is not directly suitable for CLAP. We therefore redesign the embedder to align with CLAP’s global representation. Given an audio clip $a$, the CLAP encoder $\phi_{\text{CLAP}}(a)$ first produces a $512$-dimensional embedding. This embedding is projected into a $2304$-dimensional hidden space using $W_{\text{in}}\in\mathbb{R}^{2304\times 512}$. The projected feature is then passed through a residual block consisting of LayerNorm, Linear, GELU, and Dropout layers. Finally, the feature is projected into the $768$-dimensional CLIP token space using 
$W_{\text{out}}\in\mathbb{R}^{768\times 2304}$ and $\ell_{2}$-normalized to match the scale of natural CLIP text-token embeddings. A lightweight attention refinement module (FGA) is then applied for final refinement in CLIP space:\vspace{-1.0em}

\begin{equation}
\label{eq:embedder}
  e_{\text{audio}} \;=\; \mathrm{FGA}\!\left(
    \frac{W_{\text{out}}\,\mathrm{ResBlock}(W_{\text{in}}\,\phi_{\text{CLAP}}(a))}
         {\lVert W_{\text{out}}\,\mathrm{ResBlock}(W_{\text{in}}\,\phi_{\text{CLAP}}(a))\rVert_{2}}
  \right).
\end{equation}
\vspace{-0.5em}

\noindent
The $\ell_2$-normalization step is important because cross-attention depends on normalized inner products and is therefore primarily sensitive to angular similarity rather than feature magnitude, as discussed in Section~\ref{sec:clap}. The embedder is the only trainable component in the pipeline (approximately $8.3$M parameters), while both CLAP and the diffusion backbone remain frozen (Figure~\ref{fig:method}).

\subsection{Multi-Token Extension (MTE)}
\label{sec:mte}

To support compositional generation from multiple audio inputs, we introduce a Multi-Token Extension (MTE). We define a set of unique placeholder tokens 
$\mathcal{T}=\{\langle\!*\!\rangle,\langle\hat{\,}\rangle,\langle!\rangle,\langle@\rangle,\langle\$\rangle\}$ 
and add them to the CLIP tokenizer by resizing the text-encoder embedding table. For each audio input $a_i$, the embedder produces an audio token 
$e^{(i)}_{\text{audio}}$, which is assigned to the corresponding placeholder token $\mathcal{T}[i]$. The inference prompt is then constructed as:

\vspace{-1em}
\begin{equation}
\label{eq:mte}
  p_{n} \;=\; \text{``a photo of a } \mathcal{T}[1] \text{ and a } \mathcal{T}[2] \text{ and \dots\ and a } \mathcal{T}[n]\text{''}.
\end{equation}
\vspace{-1em}

\noindent
This allows the text encoder to process a sequence of standard text tokens together with multiple audio-derived subject tokens. MTE does not require retraining the diffusion backbone or modifying its architecture, while the embedder remains the only trainable component.

\subsection{Repairing Cross-Attention Geometry at Inference Time}
\label{sec:loss}

Even when multiple audio tokens are well separated in CLIP space, standard diffusion sampling still suffers from two common problems during generation (Figure~\ref{fig:concept}): \emph{neglect}, where some tokens receive very weak cross-attention, and \emph{overlap}, where multiple tokens attend to the same spatial region, causing objects to merge together. We address these two issues during denoising by combining two complementary losses into a single mixed objective.

\vspace{1mm}\noindent
\textbf{Attend-and-Excite activation ($\Lae$).}
Let $\bar{A}^{(s)}\in[0,1]^{h\times w}$ denote the cross-attention map for token $s$, averaged across attention heads and selected U-Net layers. For the set of audio tokens
$\mathcal{S}=\{\mathcal{T}[1],\dots,\mathcal{T}[n]\}$, the Attend-and-Excite loss~\cite{attendexcite} encourages every token to strongly activate at least one spatial location:

\vspace{-1.0em}
\begin{equation}
\label{eq:lae}
  \Lae \;=\; \max_{s\in\mathcal{S}}\,\max\!\left(0,\;1-\max_{x,y}\bar{A}^{(s)}_{x,y}\right)
\end{equation}
\vspace{-1em}

\noindent
During sampling, the noisy latent is updated as
$z_{t}\leftarrow z_{t}-\alpha_{t}\,\nabla_{z_{t}}\Lae$,
which increases the activation strength of weak tokens. This ensures that all audio sources are represented in the generated image, addressing criterion (C2).

\vspace{1mm}\noindent
\textbf{A$^{*}$ losses ($\Lastar$).}
Strong activation alone is not sufficient. Different tokens may attend to the same region, merging objects (Figure~\ref{fig:concept}, column 3), and high-response regions from early denoising may drift by the end. We use soft-mask variants of the two A-STAR losses~\cite{astar}. For each token $s$ we form a soft spatial mask $m^{(s)}$ by Gaussian-smoothing its attention map, subtracting the spatial mean, and applying a temperature-scaled sigmoid, $m^{(s)}=\sigma\!\big(\gamma\,(\tilde{A}^{(s)}-\mu^{(s)})\big)$, where
$\tilde{A}^{(s)}=G(\bar{A}^{(s)})$ is the Gaussian-smoothed attention map,
$\mu^{(s)}$ its spatial mean, and $\gamma=20$. A \textit{segregation} term $\Lseg$ penalizes overlap between the masks of different tokens, and a \textit{retention} term $\Lret$ keeps each token's mask consistent across consecutive timesteps:

\vspace{-1.5em}
\begin{equation} \label{eq:lastar}
\Lseg = \frac{1}{\binom{n}{2}}\sum_{i<j} \frac{\sum_{x,y} m^{(i)}_{x,y}\,m^{(j)}_{x,y}} {\sum_{x,y}\big(m^{(i)}_{x,y}+m^{(j)}_{x,y}\big)+\varepsilon}
\quad;\quad
\Lret = \frac{1}{n}\sum_{s}\!\left( 1-\frac{\sum_{x,y}\min\!\big(m^{(s)}_{t-1,x,y},\,m^{(s)}_{t,x,y}\big)} {\sum_{x,y}\max\!\big(m^{(s)}_{t-1,x,y},\,m^{(s)}_{t,x,y}\big)+\varepsilon} \right).
\end{equation}

\vspace{-0.5em}

\noindent
Together these two terms, which we denote $\Lastar$, address C3 and reduce region collapse.

\vspace{1mm}\noindent
\textbf{Mixed objective.} The two losses serve different purposes. $\Lae$ encourages strong activation of every token, while $\Lastar$ controls where each token appears. We combine them as:

\vspace{-0.75em}
\begin{equation}
\label{eq:lmix}
  \Lmix \;=\; \lambda_{\text{AE}}\,\Lae \;+\; \lambda_{\text{seg}}\,\Lseg \;+\; \lambda_{\text{ret}}\,\Lret
\end{equation}
\vspace{-1.25em}

\noindent
where $(\lambda_{\text{AE}},\lambda_{\text{seg}},\lambda_{\text{ret}})=(1.0,\,2.0,\,0.5)$ in all experiments. Optimization is applied during the first $T_{\text{max}}=25$ of $50$ denoising steps, since early denoising stages determine the global scene layout, whereas later stages primarily refine texture details. As shown in Section~\ref{sec:ablation}, the mixed objective consistently performs better than using either loss alone.

\subsection{Training on a Focused 50-Class Subset}
\label{sec:training}

We train the embedder on a curated $50$-class subset of VGGSound ($\sim$25K audio--image pairs) instead of the full $309$-class dataset. The main reason is representational capacity. The embedder contains only $\sim$8.3M trainable parameters, and learning mappings for all $309$ classes, many of which are visually similar or acoustically ambiguous, can make optimization unstable and lead to weaker token separation. By restricting training to $50$ acoustically and visually distinct classes, the task becomes easier for the embedder to learn. This helps produce cleaner gradients, more stable optimization, and tighter token clusters in CLIP space. The effect is also reflected quantitatively in the clustering results shown in Table~\ref{tab:clustering}.
\section{Experiments}
\label{sec:experiments}

\subsection{Datasets}

\textbf{VGGSound}~\cite{vggsound} is a large-scale audio-visual dataset containing approximately $200$K audio clips of length $10$ seconds across $309$ classes, each paired with a sampled image frame. We train our embedder on a curated $50$-class subset and use its test split for single-audio evaluation, which we refer to as \textit{VGG}. Classes outside this training vocabulary are treated as out-of-distribution. \textbf{IS3}~\cite{audiotoken} (Interactive Synthetic Sound Source) is a compositional benchmark that includes two audio clips from different classes, along with a synthesized composite reference image. We use IS3 as an out-of-distribution multi-audio benchmark because its compositions are drawn from a broader, partially non-overlapping vocabulary.

\vspace{1mm}\noindent
\textbf{New compositional benchmarks (VGG2, VGG3, VGGn).}
To evaluate compositional generation under controlled settings, we construct three multi-audio benchmarks from the VGGSound test split. \textbf{VGG2} contains $2500$ groups, each consisting of exactly two audio-image pairs sampled from different classes. \textbf{VGG3} contains $1000$ groups of three audio-image pairs and supports a more detailed analysis of the three-audio setting. \textbf{VGGn} contains $500$ groups with variable group size $n\in\{2,3,4,5\}$. To ensure balanced class coverage, we sample a group size uniformly, then select acoustically distinct classes without replacement, giving preference to classes that have not yet appeared in earlier groups. Audio-image pairs within each class are drawn sequentially, ensuring all samples are used before being reused. All splits are generated with a fixed random seed.

\subsection{Metrics}
We extend six standard audio-to-image metrics to the multi-source setting. For groups with $n$ audio inputs, each metric is aggregated across all $n$ objects in the group. \textbf{IIS} (Image--Image Similarity) measures how often the generated image is more similar, in CLIP ViT-L/14 space, to its ground-truth reference than to random references; higher is better. \textbf{AIS} (Audio--Image Similarity) is
the analogous win rate between the generated image (CLIP ViT-B/32) and the wav2CLIP~\cite{wav2clip} embedding of each input audio, ranked against random audios. \textbf{AIC} (Audio--Image Content) is the recall of the $n$ ground-truth labels within the top-$n$ predictions of a zero-shot CLIP ViT-L/14 classifier over the $50$ candidate classes, reducing to top-1 accuracy when $n=1$. \textbf{CLS} (Caption-Label Similarity) is the geometric mean, over labels, of the cosine
similarity between a BLIP~\cite{blip}-generated caption and each class label. \textbf{CCS} (Caption--Caption Similarity) is the same aggregation between captions of the generated and reference images. \textbf{FID}~\cite{fid} measures the distributional distance between generated and real images in InceptionV3 feature space, where lower values are better. 
Five of these six metrics use CLIP- or BLIP-derived representations, which our method also aligns to, so we report FID alongside them as a CLIP-independent check. Moreover, the coupling is indirect, since our inference-time objective acts on cross-attention maps and never optimizes CLIP image--text similarity.

\subsection{Baselines and Variants}

We compare our method with four existing approaches: \textbf{AudioToken}~\cite{audiotoken} trained on the full $309$-class VGGSound dataset (denoted as \textit{AT (309)}), \textbf{SonicDiffusion}~\cite{sonicdiffusion}, \textbf{Sound2Scene}~\cite{sound2scene}, and \textbf{MACS}~\cite{macs}. For a fair comparison, each baseline receives the same pre-separated audio sources as our method, so that differences reflect the generation stage rather than source-separation quality. For MACS, which includes a source separation stage, we bypass this stage and provide the same separated sources directly to its generation module. \textit{AT (309)} uses the released $309$-class AudioToken weights with our multi-token extension at inference, so it also produces one combined image per group and remains directly comparable. We denote our variants as follows: \textbf{AT}, which uses the multi-token extension with the BEATs encoder trained on the curated $50$-class subset; \textbf{AT Mixed}, which adds the mixed inference objective $\Lmix$; and \textbf{AT Mixed (Clap)}, which further replaces BEATs with the CLAP encoder. This final variant (AT Mixed (Clap)) is referred to as ``Ours'' in the figures. In all tables, the best result in each column is highlighted in \textcolor{bestcol}{violet}, while the second-best result is shown in \textbf{bold}. Rows corresponding to our variants are shaded.

\subsection{Main Results}

\begin{table}[!t]
\centering
\small
\setlength{\tabcolsep}{4.5pt}
\renewcommand{\arraystretch}{1.05}
\resizebox{0.85\linewidth}{!}{
\begin{tabular}{lcccccc}
\toprule
\textbf{Method} & \textbf{IIS}$\uparrow$ & \textbf{AIS}$\uparrow$ & \textbf{AIC}$\uparrow$ & \textbf{CLS}$\uparrow$ & \textbf{CCS}$\uparrow$ & \textbf{FID}$\downarrow$ \\
\midrule
\multicolumn{7}{l}{\textit{VGG --- single audio (in-distribution)}} \\
SonicDiffusion~\cite{sonicdiffusion} & 0.4955 & 0.5137 & 0.0574 & 0.4719 & 0.4348 & 86.0037 \\
Sound2Scene~\cite{sound2scene} & 0.4873 & 0.4877 & 0.0794 & 0.5242 & 0.4776 & 112.3871 \\
AudioToken (309)~\cite{audiotoken} & 0.6310 & 0.6830 & 0.1394 & 0.5286 & 0.5067 & 93.6600 \\
MACS~\cite{macs} & \textbf{0.7780} & \textcolor{bestcol}{0.7940} & 0.2616 & 0.5615 & \textbf{0.5344} & \textcolor{bestcol}{58.6611} \\
\rowcolor{ourrow}
AT (ours) & 0.7330 & 0.7200 & 0.4337 & 0.5837 & 0.5343 & 70.0351 \\
\rowcolor{ourrow}
AT Mixed (ours) & 0.7372 & 0.7215 & \textbf{0.4477} & \textbf{0.5846} & 0.5330 & 62.3452 \\
\rowcolor{ourrow}
AT Mixed (Clap) (ours) & \textcolor{bestcol}{0.8047} & \textbf{0.7484} & \textcolor{bestcol}{0.5904} & \textcolor{bestcol}{0.5878} & \textcolor{bestcol}{0.5496} & \textbf{60.0736} \\
\midrule
\multicolumn{7}{l}{\textit{VGG2 --- two audios (in-distribution compositional)}} \\
AudioToken (309)~\cite{audiotoken} & 0.5654 & 0.5957 & 0.1214 & 0.5005 & 0.4740 & 93.9700 \\
MACS~\cite{macs} & 0.6476 & \textbf{0.6625} & 0.2126 & 0.5122 & 0.4917 & 57.8110 \\
\rowcolor{ourrow}
AT (ours) & 0.6553 & 0.6550 & 0.3822 & \textbf{0.5312} & \textbf{0.4929} & 60.4968 \\
\rowcolor{ourrow}
AT Mixed (ours) & \textbf{0.6701} & 0.6611 & \textbf{0.3926} & \textcolor{bestcol}{0.5315} & \textcolor{bestcol}{0.4938} & \textbf{55.5968} \\
\rowcolor{ourrow}
AT Mixed (Clap) (ours) & \textcolor{bestcol}{0.6993} & \textcolor{bestcol}{0.6677} & \textcolor{bestcol}{0.4332} & 0.5237 & 0.4902 & \textcolor{bestcol}{54.1177} \\
\midrule
\multicolumn{7}{l}{\textit{IS3 --- two audios (out-of-distribution, broader vocab.)}} \\
AudioToken (309)~\cite{audiotoken} & 0.6815 & 0.6103 & 0.1313 & \textcolor{bestcol}{0.5240} & 0.4915 & 77.8841 \\
MACS~\cite{macs} & 0.7251 & \textcolor{bestcol}{0.6516} & 0.1337 & 0.5116 & 0.4908 & 89.3090 \\
\rowcolor{ourrow}
AT (ours) & \textbf{0.7490} & 0.6174 & \textbf{0.2110} & \textbf{0.5205} & \textcolor{bestcol}{0.5101} & 69.6100 \\
\rowcolor{ourrow}
AT Mixed (ours) & \textcolor{bestcol}{0.7520} & 0.6188 & \textcolor{bestcol}{0.2131} & 0.5079 & \textbf{0.5035} & \textcolor{bestcol}{66.4075} \\
\rowcolor{ourrow}
AT Mixed (Clap) (ours) & 0.7291 & \textbf{0.6221} & 0.1688 & 0.4983 & 0.4914 & \textbf{69.0440} \\
\midrule
\multicolumn{7}{l}{\textit{VGG3 --- three audios (in-distribution compositional)}} \\
AudioToken (309)~\cite{audiotoken} & 0.5447 & 0.5615 & 0.1233 & 0.4867 & 0.4395 & 120.7110 \\
MACS~\cite{macs} & 0.5944 & 0.6098 & 0.1957 & 0.4978 & \textbf{0.4764} & 77.8314 \\
\rowcolor{ourrow}
AT (ours) & 0.6237 & \textbf{0.6215} & 0.3523 & \textcolor{bestcol}{0.5147} & \textcolor{bestcol}{0.4783} & 76.3441 \\
\rowcolor{ourrow}
AT Mixed (ours) & \textbf{0.6298} & 0.6210 & \textbf{0.3527} & \textbf{0.5111} & 0.4742 & \textbf{74.8436} \\
\rowcolor{ourrow}
AT Mixed (Clap) (ours) & \textcolor{bestcol}{0.6528} & \textcolor{bestcol}{0.6266} & \textcolor{bestcol}{0.3703} & 0.5026 & 0.4730 & \textcolor{bestcol}{74.0027} \\
\midrule
\multicolumn{7}{l}{\textit{VGGn --- variable $n\in\{2,3,4,5\}$ (in-distribution, scaling)}} \\
AudioToken (309)~\cite{audiotoken} & 0.5351 & 0.5546 & 0.1411 & 0.4787 & 0.4326 & 122.0682 \\
MACS~\cite{macs} & 0.5731 & 0.5930 & 0.1926 & 0.4982 & \textcolor{bestcol}{0.4748} & 77.2468 \\
\rowcolor{ourrow}
AT (ours) & 0.6094 & \textbf{0.5946} & \textbf{0.3366} & \textcolor{bestcol}{0.5040} & \textbf{0.4697} & 78.1004 \\
\rowcolor{ourrow}
AT Mixed (ours) & \textbf{0.6124} & \textcolor{bestcol}{0.5973} & \textcolor{bestcol}{0.3381} & \textbf{0.5026} & 0.4684 & \textcolor{bestcol}{75.8900} \\
\rowcolor{ourrow}
AT Mixed (Clap) (ours) & \textcolor{bestcol}{0.6183} & 0.5898 & \textcolor{bestcol}{0.3381} & 0.4931 & 0.4661 & \textbf{76.9760} \\
\bottomrule
\end{tabular}}
\vspace{0.4em}
\caption{\textbf{Quantitative Comparison.} Each block reports one benchmark
(VGG, VGG2, IS3, VGG3, VGGn); rows are methods (baselines above, our variants
shaded below) and columns are the six metrics, with $\uparrow$/$\downarrow$
indicating whether higher or lower is better. In each column the best result is
in \textcolor{bestcol}{violet} and the second-best in \textbf{bold}.}
\label{tab:main_eval}
\vspace{-1.8em}
\end{table}

\begin{figure}[!t]
\centering
\setlength{\tabcolsep}{1pt}
\renewcommand{\arraystretch}{1.05}
\begin{tabular}{ccccc}
\textbf{\scriptsize Ours} & \textbf{\scriptsize AT Mixed} &
\textbf{\scriptsize AT} & \textbf{\scriptsize AT (309)} &
\textbf{\scriptsize MACS}~\cite{macs} \\\vspace{-0.2em}
\includegraphics[width=0.17\linewidth]{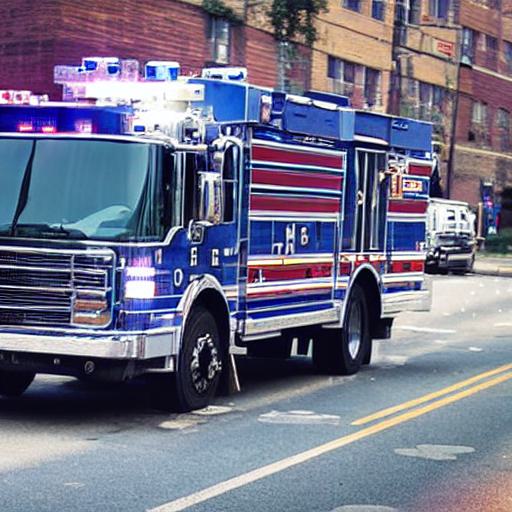} &
\includegraphics[width=0.17\linewidth]{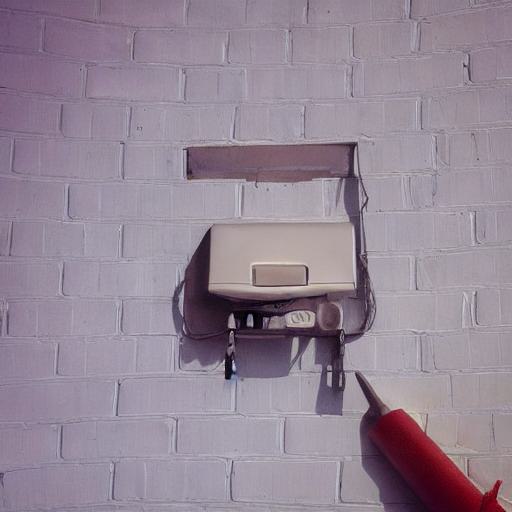} &
\includegraphics[width=0.17\linewidth]{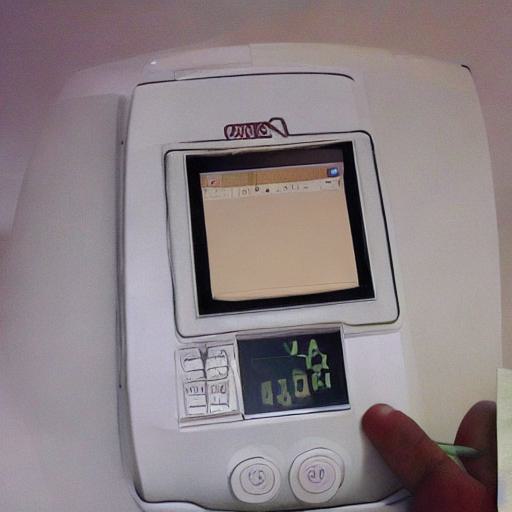} &
\includegraphics[width=0.17\linewidth]{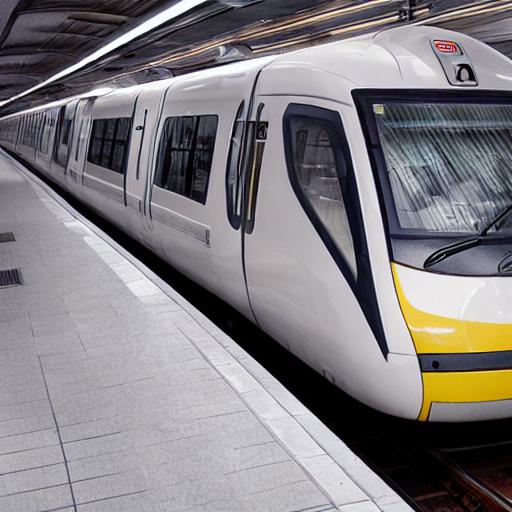} &
\includegraphics[width=0.17\linewidth]{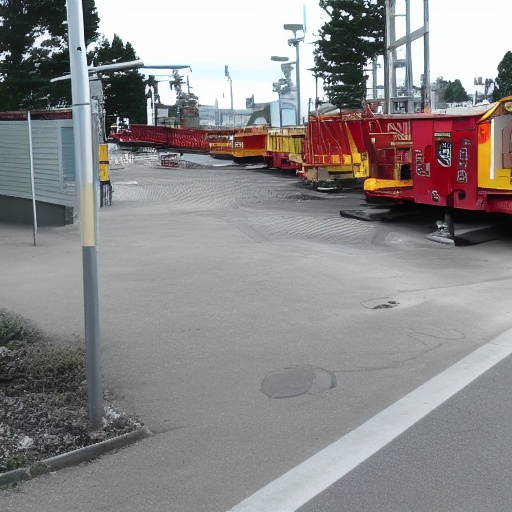} \\\vspace{-0.2em}
\includegraphics[width=0.17\linewidth]{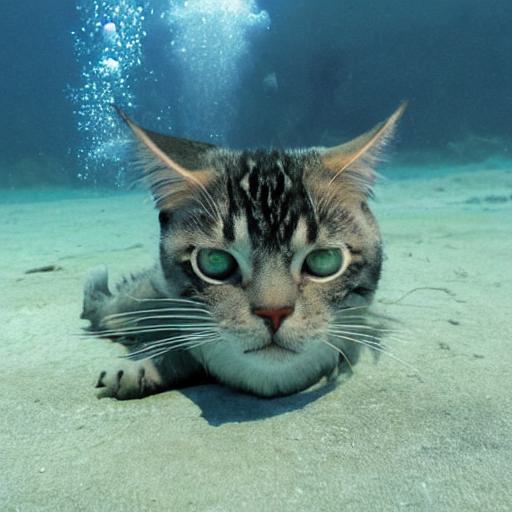} &
\includegraphics[width=0.17\linewidth]{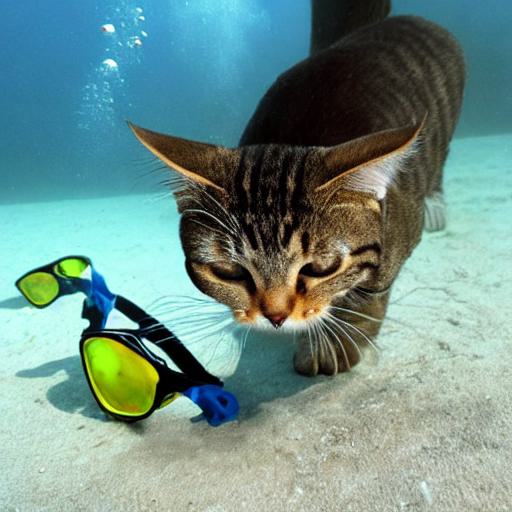} &
\includegraphics[width=0.17\linewidth]{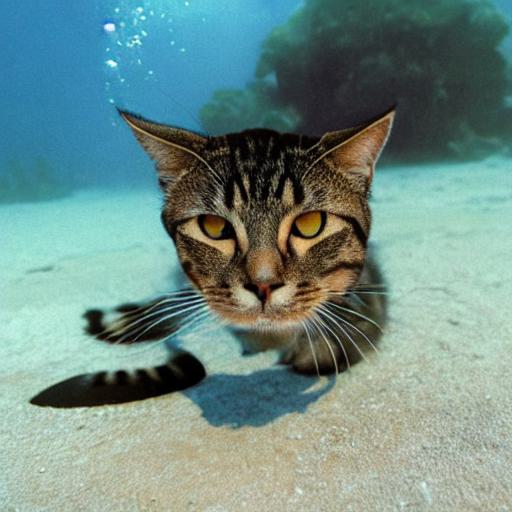} &
\includegraphics[width=0.17\linewidth]{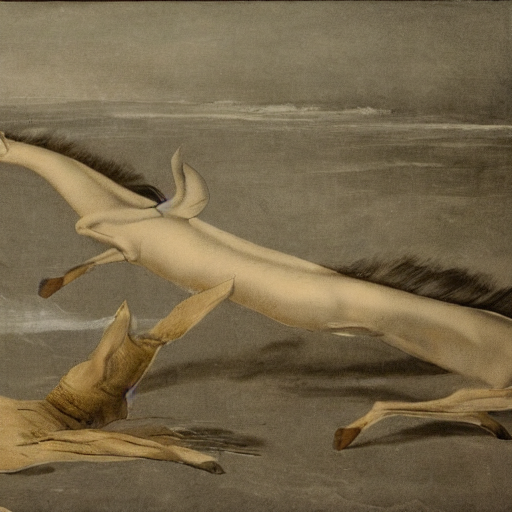} &
\includegraphics[width=0.17\linewidth]{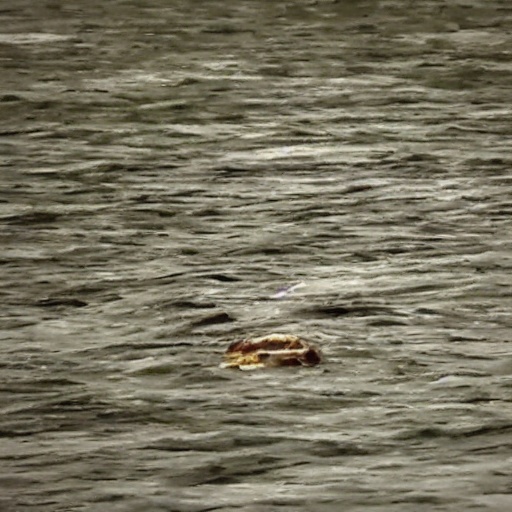} \\\vspace{-0.2em}
\includegraphics[width=0.17\linewidth]{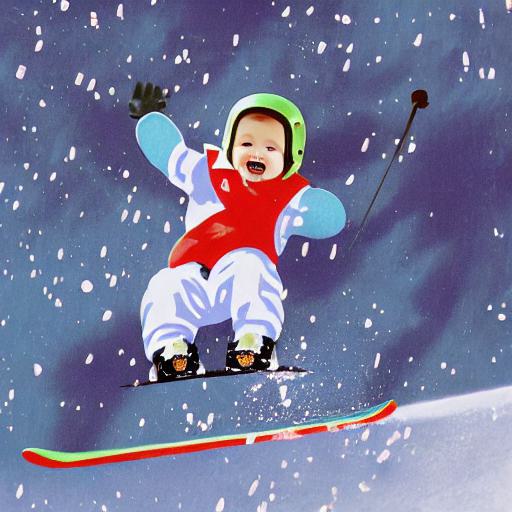} &
\includegraphics[width=0.17\linewidth]{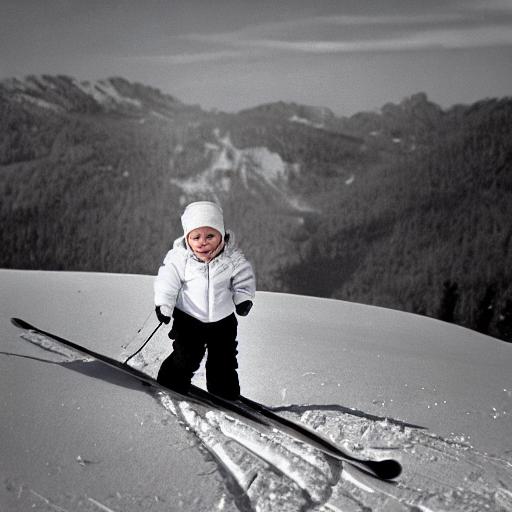} &
\includegraphics[width=0.17\linewidth]{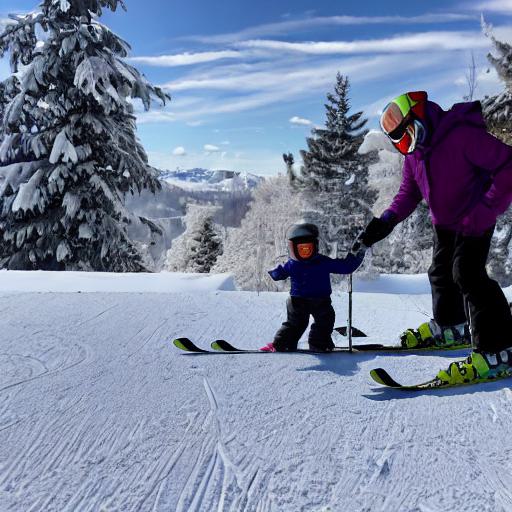} &
\includegraphics[width=0.17\linewidth]{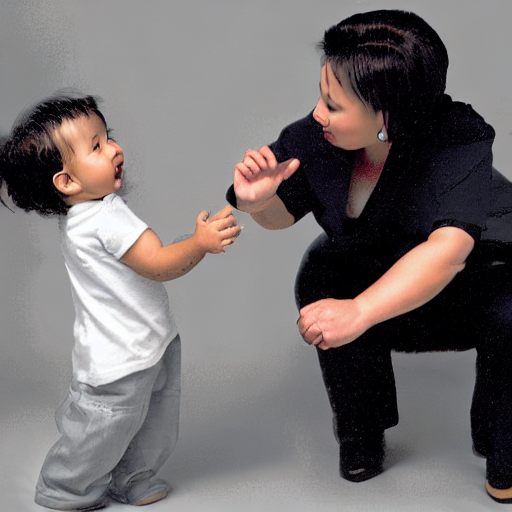} &
\includegraphics[width=0.17\linewidth]{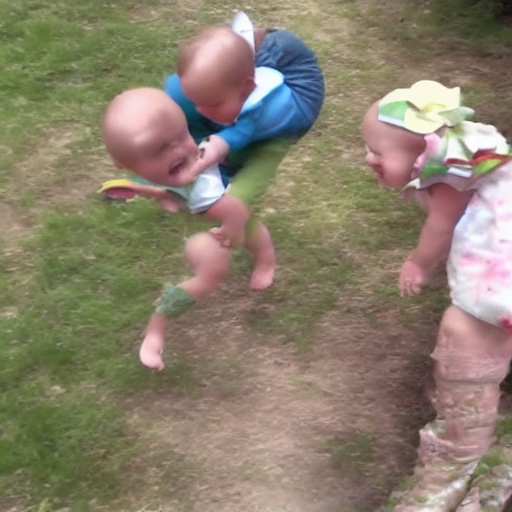} \\
\includegraphics[width=0.17\linewidth]{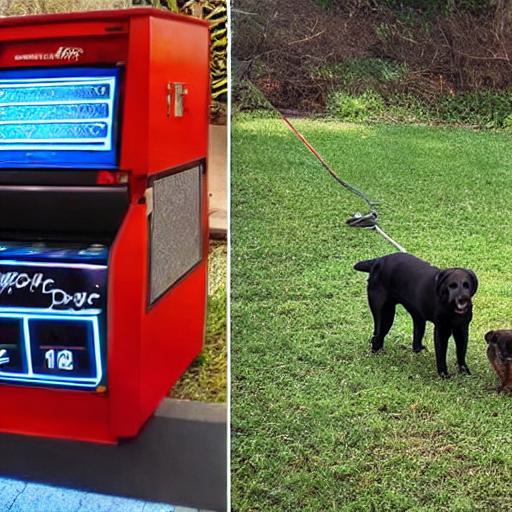} &
\includegraphics[width=0.17\linewidth]{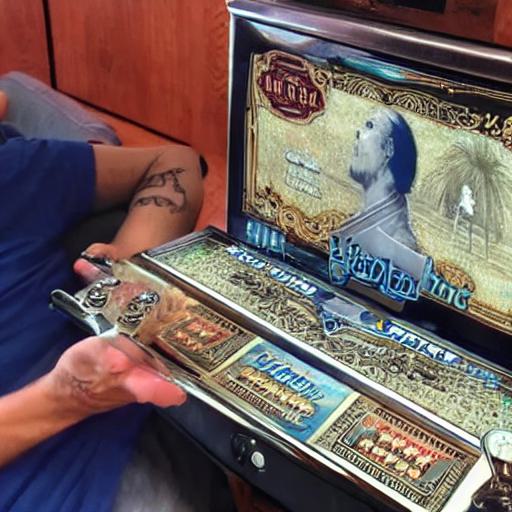} &
\includegraphics[width=0.17\linewidth]{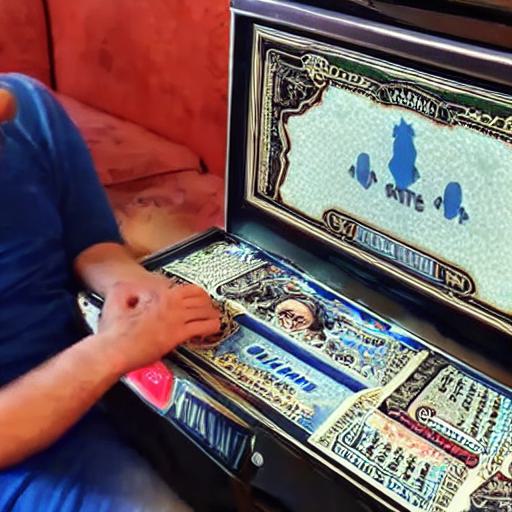} &
\includegraphics[width=0.17\linewidth]{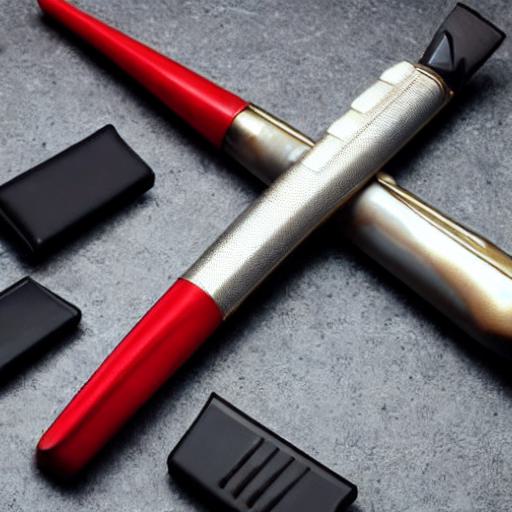} &
\includegraphics[width=0.17\linewidth]{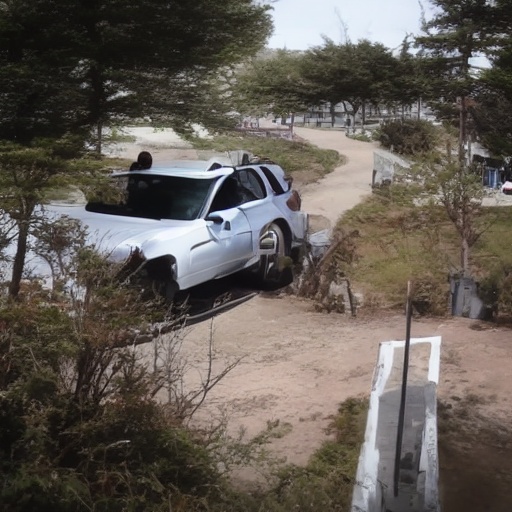} \\
\end{tabular}\vspace{-0.5em}
\caption{\textbf{Qualitative comparison across different compositional settings.} The first row shows $n{=}1$ on VGG with \textit{fire truck siren}; the second and third rows show $n{=}2$ on VGG2 with \textit{cat purring \& scuba diving} and \textit{baby laughter \& skiing}, respectively; and the fourth row shows $n{=}3$ on VGG3 with \textit{dog barking \& slot machine \& hedge trimmer running}. Columns are ordered from our method to the baselines. Our method generates all requested objects in separate, coherent spatial regions, whereas \textit{AT (309)} and \textit{MACS} often miss one source or merge multiple objects. This difference becomes more pronounced at $n{=}3$, where the cross-attention budget is shared among more tokens.}
\label{fig:results_main}
\vspace{-1.3em}
\end{figure}

The central prediction of the geometry hypothesis is that encoder substitution should substantially improve compositional fidelity, beyond the effects of inference-time optimization. The ablation results support this hypothesis. On VGGSound, AT~(Clap) without any inference optimization achieves AIC $= 0.5788$, compared to AT~Mixed with the best inference strategy at AIC $= 0.4477$, a gap of $+0.131$ associated with the improved token geometry. The same ordering holds on VGG2 ($0.4010$ vs.\ $0.3926$). Since SonicDiffusion and Sound2Scene are single-audio methods, only AudioToken and MACS appear on the compositional benchmarks.

\vspace{1mm} \noindent
\textbf{Single-audio (VGG).} AT Mixed (Clap) achieves the best score on $4$ of $6$ metrics, including the most
class-discriminative one (AIC: $0.5904$ vs.\ $0.4477$ for AT Mixed and $0.2616$ for MACS, a $\times 2.3$ margin).The AIC margin is particularly informative in this block, as it measures whether the generated image is recognizable as the correct audio class and gives an AIC approximately $2.3\times$ higher for our method. IIS, CLS, and CCS also favor AT Mixed (Clap), supporting the observation that the improved CLAP-based token representation translates into more visually and semantically faithful generations. MACS retains an advantage on AIS ($0.7940$ vs $0.7484$), which we attribute to its joint training objective 
that directly optimizes audio-visual alignment. Our method trades raw alignment for class-discriminative fidelity, reflected in the substantially higher AIC scores, which we consider the more informative metric for compositional generation.

\vspace{1mm}\noindent
\textbf{Two-audio (VGG2).} The introduction of a second audio token makes attention neglect a real concern, and the value of geometric alignment becomes more visible. AT Mixed (Clap) achieves the best score on IIS, AIS, AIC, and FID, while AT Mixed is second on IIS, AIC, and FID and best on CLS and CCS, indicating that the mixed loss already produces a strong compositional baseline before CLAP is added. Against MACS, the AIC margin is $\times 2.0$ ($0.4332$ vs.\ $0.2126$); we improve IIS by $5.2$ points and FID by $3.7$. Figure~\ref{fig:results_main} (middle rows) shows the corresponding qualitative pattern: our method renders \emph{both} a cat and a scuba diver (row 2) or a laughing baby and a skier (row 3) in distinct regions, while AT(309) drops one source and MACS produces a single ambiguous object.

\vspace{1mm}\noindent
\textbf{IS3 out-of-distribution results.} AT~Mixed~(Clap) scores below
AT~Mixed on AIC ($0.1688$ vs.\ $0.2131$) and on CLS. This does not contradict our geometric analysis, since it follows from how each encoder behaves on unseen classes. CLAP produces a single global embedding aligned to a fixed language vocabulary, so audio outside that vocabulary may be mapped toward familiar concepts, potentially reducing the detail needed to render the correct object. BEATs is instead trained by masked acoustic-token prediction over the broad AudioSet distribution, with no text alignment, and it exposes a frame-level feature sequence rather than a single compressed vector, so it degrades more gracefully on unseen classes. The angular alignment that benefits CLAP in-distribution may therefore be less effective when the test audio falls outside the learned vocabulary, as on IS3, where the broader features of BEATs provide better robustness. Extending the embedder to broader training vocabularies is a direction for future work, as discussed in Appendix~$G$.

\vspace{1mm}\noindent
\textbf{Three-audio and variable-$n$.} The relative ranking established at $n{=}2$ is stable: AT Mixed (Clap) wins IIS, AIS, AIC,
and FID on VGG3 and IIS, AIC, and FID on VGGn. Absolute scores degrade with $n$, as expected, because more objects must share the same canvas and more tokens compete for a fixed cross-attention budget. Crucially, the AIC margin between our method and MACS remains pronounced across all $n$ values ($\times 1.9$ on VGG3, $\times 1.8$ on VGGn), indicating that the benefits of CLAP-based token geometry and the mixed objective remain substantial as compositional complexity increases. The bottom row of Figure~\ref{fig:results_main} shows the three-audio case: our method renders the dog, slot machine, and hedge trimmer as identifiable objects, whereas baselines collapse the scene into one or two of them.

\begin{figure}[!t]
    \centering

    \begin{minipage}{0.75\linewidth}
        \centering
        \includegraphics[width=\linewidth]{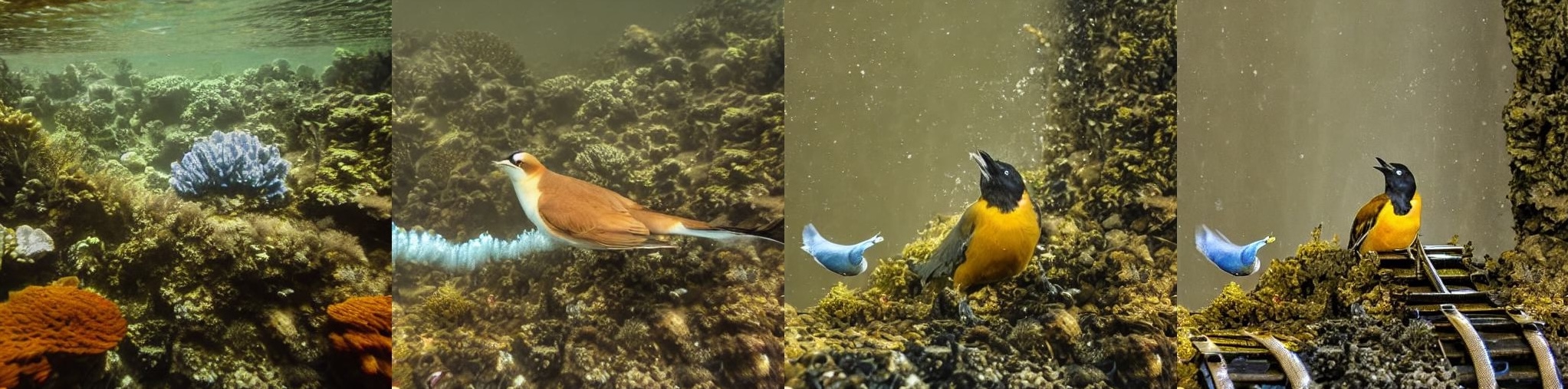}
        
        \vspace{-0.3mm}
        
        {\scriptsize\itshape
        \makebox[0.25\linewidth][c]{scuba diving}%
        \makebox[0.25\linewidth][c]{+ baltimore oriole}%
        \makebox[0.25\linewidth][c]{+ woodpecker}%
        \makebox[0.25\linewidth][c]{+ train whistling}\par}
    \end{minipage}

    \vspace{1pt}

    \begin{minipage}{0.75\linewidth}
        \centering
        \includegraphics[width=\linewidth]{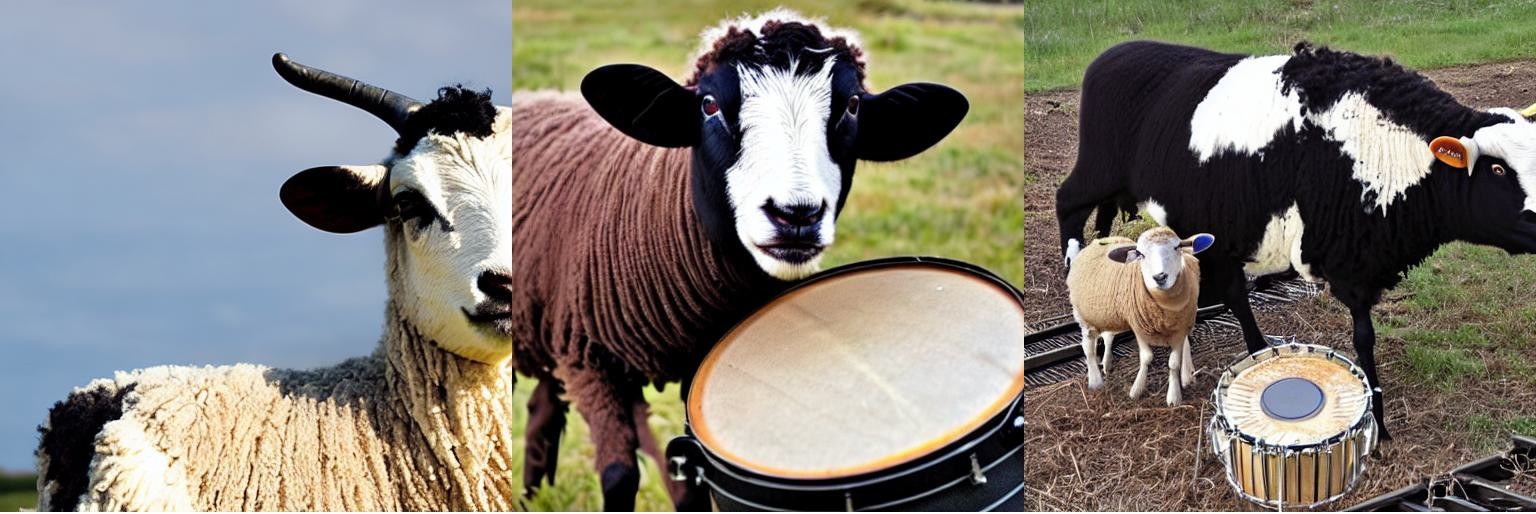}
        
        \vspace{-0.3mm}
        
        {\scriptsize\itshape
        \makebox[0.330\linewidth][c]{sheep bleating}%
        \makebox[0.330\linewidth][c]{+ drum kit}%
        \makebox[0.330\linewidth][c]{+ train wagon}\par}
    \end{minipage}

    \vspace{1pt}

    \begin{minipage}{0.75\linewidth}
        \centering
        \includegraphics[width=\linewidth]{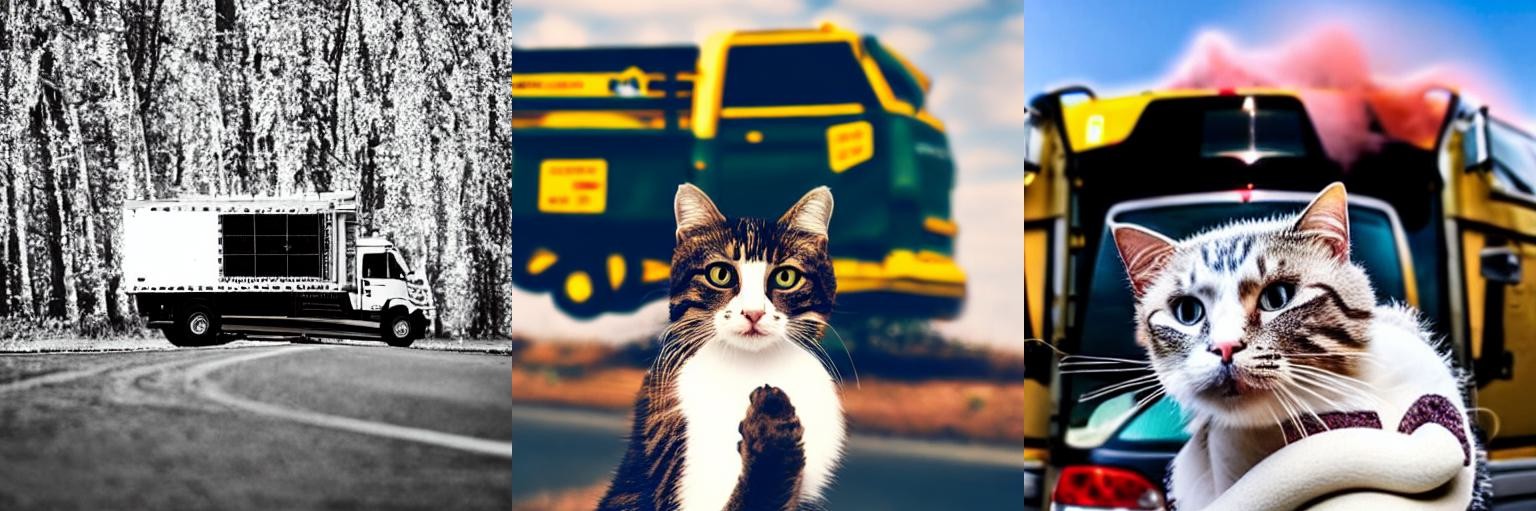}
        
        \vspace{-0.08mm}
        
        {\scriptsize\itshape
        \makebox[0.35\linewidth][c]{ice cream truck}%
        \makebox[0.30\linewidth][c]{+ cat purring}%
        \makebox[0.30\linewidth][c]{+ volcano}\par}
    \end{minipage}\vspace{-0.1em}
    \caption{\textbf{Additional progressive compositions generated by our method.} Each row extends a scene by adding one audio source at a time. The \textit{top row} shows a four-source progression starting from \textit{scuba diving} and successively adding \textit{baltimore oriole}, \textit{woodpecker}, and \textit{train whistling}, with almost all four concepts remaining visible in the final image. The \textit{middle} and \textit{bottom} rows show three-source examples. Across all cases, our method preserves previously generated objects instead of overwriting them as new audio sources are added.}
    
    \label{fig:prog_extras}
    \vspace{-1.3em}
\end{figure}

\vspace{1mm}\noindent
\textbf{Progressive compositions.} Figure~\ref{fig:prog_extras} grows the prompt one audio source at a time, adding one new source per panel, consistent with the quantitative drop-off in Table~\ref{tab:main_eval}. Object retention is preserved through $n{=}3$ and remains qualitatively legible at $n{=}4$ and $n{=}5$, even though new sources increasingly compete with prior ones for the shared attention budget. The teaser (Figure~\ref{fig:teaser}) shows the same pattern on the $n{=}4$ snare-drum-to-church-bell series and the $n{=}3$ church-bell-to-drum-kit series.

\subsection{Ablations}
\label{sec:ablation}

We ablate the two components of the mixed loss, attend-and-excite ($\Lae$) and the spatial A$^{*}$ loss ($\Lastar$), over the BEATs and CLAP variants of the embedder. Table~\ref{tab:ablation_summary} reports the consistent trend across the four compositional benchmarks, and per-dataset breakdowns are in the supplementary Tables~$S1-S5$.

\vspace{1mm} \noindent
\textbf{(i) $\Lae$ primarily improves perceptual quality.} Adding $\Lae$ consistently yields the largest single-component reduction in FID, while ranking metrics change
little. $\Lae$ encourages each token to claim a meaningful spatial region, which improves perceptual coherence without necessarily affecting which class is recognized at that
region.

\vspace{1mm} \noindent
\textbf{(ii) $\Lastar$ preserves caption alignment.} On compositional splits, $\Lastar$ achieves the highest CLS and CCS among single-loss variants. By forcing
distinct tokens onto disjoint spatial supports, it produces cleaner per-object regions, which captioners describe more accurately.

\vspace{1mm} \noindent
\textbf{(iii) The mixed objective is consistently best on AIC.} Across VGG2, VGG3, and VGGn, AT + $\Lmix$ achieves the joint-highest AIC within each encoder family. Activation
alone is insufficient (tokens fuse); separation alone is insufficient (weak tokens stay weak). Both contribute to class-correct compositional generation, which is directly reflected in the AIC metric, and the same pattern is visualized in Figure~\ref{fig:concept}.

\begin{table}[!t]
\centering
\small
\setlength{\tabcolsep}{4.5pt}
\renewcommand{\arraystretch}{1.0}
\begin{tabular}{lcccccc}
\toprule
\textbf{Encoder / Loss} & \textbf{IIS}$\uparrow$ & \textbf{AIS}$\uparrow$ & \textbf{AIC}$\uparrow$ & \textbf{CLS}$\uparrow$ & \textbf{CCS}$\uparrow$ & \textbf{FID}$\downarrow$ \\
\midrule
\multicolumn{7}{l}{\textit{BEATs encoder (AT)}} \\
~~~~no inference loss              & 0.6593 & 0.6221 & 0.3205 & 0.5176 & \textcolor{bestcol}{0.4878} & 71.1378 \\
~~~~+ $\Lae$                       & 0.6604 & 0.6154 & 0.3159 & 0.5036 & 0.4776 & \textcolor{bestcol}{67.4298} \\
~~~~+ $\Lastar$                    & 0.6597 & 0.6199 & 0.3182 & \textcolor{bestcol}{0.5209} & 0.4822 & 71.8018 \\
~~~~+ $\Lmix$ (\textbf{mixed})     & \textcolor{bestcol}{0.6661} & \textcolor{bestcol}{0.6246} & \textcolor{bestcol}{0.3241} & 0.5133 & 0.4850 & 68.1845 \\
\midrule
\multicolumn{7}{l}{\textit{CLAP encoder (AT-Clap)}} \\
~~~~no inference loss              & 0.6670 & 0.6132 & 0.3123 & 0.5104 & 0.4831 & 69.9187 \\
~~~~+ $\Lae$                       & 0.6654 & 0.6148 & 0.3171 & 0.4949 & 0.4729 & 71.9798 \\
~~~~+ $\Lastar$                    & 0.6625 & 0.6127 & 0.3148 & \textcolor{bestcol}{0.5162} & \textcolor{bestcol}{0.4866} & 70.9196 \\
~~~~+ $\Lmix$ (\textbf{mixed})     & \textcolor{bestcol}{0.6749} & \textcolor{bestcol}{0.6266} & \textcolor{bestcol}{0.3276} & 0.5044 & 0.4802 & \textcolor{bestcol}{68.5351} \\
\bottomrule
\end{tabular}
\caption{\textbf{Loss ablation averaged across compositional benchmarks (VGG2/IS3/VGG3/VGGn).} For each encoder-loss combination, we report the average score across all four benchmarks. Per-dataset ablation results are provided in supplementary Tables~$S1-S5$. \textcolor{bestcol}{Best} indicates the best result within each encoder block. Two trends are consistent across experiments: (i) the mixed objective performs best or tied-best on AIC, the metric most directly related to compositional correctness, and (ii) CLAP-based variants consistently outperform their BEATs counterparts on in-distribution benchmarks.
}
\label{tab:ablation_summary}
\vspace{-1.5em}
\end{table}

\vspace{1mm} \noindent
\textbf{(iv) CLAP dominates in-distribution; BEATs is more robust out-of-distribution.} Every CLAP variant achieves higher IIS and AIC than its BEATs counterpart with the same
loss on VGG, VGG2, and VGG3. On IS3, however,
BEATs variants outperform CLAP variants on AIC and CLS, for the reasons discussed in Section~$4.4$.

\vspace{1mm} \noindent
\textbf{(v) The CLAP advantage narrows as $n$ grows.} On VGGn, the gap between CLAP and BEATs variants is much smaller than on VGG2 and VGG3. At high $n$ the attention budget, rather than per-token quality, appears to become the limiting factor, which points to a clear direction for future work.

\vspace{1mm}\noindent
Due to space constraints, we provide an in-depth interpretation and discussion of these results in Supplementary Appendix~$G$.
\section{Conclusion}
\label{sec:conclusion}

We addressed the problem of compositional audio-to-image generation by studying the geometry of audio tokens in the CLIP text-token space. Our analysis identifies three key properties spanning intra-class compactness, angular alignment with the text manifold, and inter-token spatial separation. We show that these properties are closely associated with common failure modes in existing methods, and we design our framework to address each. Although our method builds on existing components such as textual inversion, Attend-and-Excite, and A$^{*}$ separation, it combines them within a unified geometric framework. The proposed approach achieves strong performance on compositional metrics such as AIC, IIS, and FID, with AIC improvements of up to 2.25$\times$ over prior methods. We also introduce three new multi-source benchmarks, VGG2, VGG3, and VGGn, to support future research in compositional audio-to-image generation. While our study focuses on audio, the proposed geometric perspective may extend to other modalities that project learned embeddings into pre-trained text spaces for cross-attention.

\vspace{1mm}\noindent
\textbf{Limitations.} Our method assumes separated audio streams and inherits the limitations of any external separation model. The embedder is trained on a curated $50$-class subset, so broader vocabularies require additional training. Finally, the mixed objective acts during denoising, raising per-image cost by roughly $2.8$-$3.1\times$ (Appendix~$F$).

\section{Acknowledgment}
This work was supported by the Prime Minister Research Fellowship (PMRF 2122-2557) awarded to Prajwal Singh and by the Dr. Vilas Mujumdar Chair held by Shanmuganathan Raman.

\bibliography{egbib}

\clearpage
\appendix

\setcounter{section}{0}
\renewcommand{\thesection}{\Alph{section}}
\renewcommand{\thetable}{S\arabic{table}}
\renewcommand{\thefigure}{S\arabic{figure}}
\setcounter{table}{0}
\setcounter{figure}{0}

\begin{center}
{\Large \textbf{Supplementary Material}}
\end{center}

\vspace{1em}\noindent
This supplementary provides material that complements the main paper:
\textbf{(A) full implementation details and hyperparameters},
\textbf{(B) per-dataset ablation tables} (the main paper reports the benchmark-averaged
summary, Table~$3$),
\textbf{(C) additional qualitative rows} for single-, two-, three-, and large-$n$ inputs (the main paper shows one representative row per regime in Figure~$5$), \textbf{(D) cross-attention map evolution} across denoising timesteps for several audio pairs and triples (the main paper shows the $t{=}50$ snapshot for one pair in Figure~$2$), and
\textbf{(E) the remaining UMAP panels} of the audio-token geometry (one panel is in the main paper as Figure~$4$), 
\textbf{(F) an inference-time cost analysis} of the mixed objective, and 
\textbf{(G) an extended discussion} of the results.

\section{Implementation Details}
\label{supp:impl}

\vspace{1mm} \noindent
\textbf{Diffusion backbone.} We use Stable Diffusion v1.5~\cite{ldm} as the frozen latent diffusion backbone
throughout. All experiments use $50$ DDIM sampling steps with a classifier-free guidance scale of $7.5$, an image resolution of $512{\times}512$, and a fixed random seed per
evaluation group for reproducibility.

\vspace{1mm} \noindent
\textbf{Audio encoders.} The BEATs~\cite{beats} variant uses the \texttt{BEATs\_iter3\_plus\_AS2M} \texttt{\_finetu-} \texttt{ned} checkpoint and consumes raw $16$~kHz waveforms of $10$~s duration, producing a frame-level feature sequence of shape $(L, 768)$ which is then attention-pooled by the original AudioToken~\cite{audiotoken} embedder. The CLAP~\cite{clap} variant uses the official \texttt{laion-clap-htsat-fused} checkpoint, which produces a single
$512$-dimensional global embedding per clip via the audio encoder branch.

\vspace{1mm} \noindent
\textbf{Embedder architecture (CLAP variant).} The redesigned embedder (Section~$3.4$, Eq.~$1$) operates on the global CLAP embedding $\phi_{\text{CLAP}}(a)\in\mathbb{R}^{512}$ and produces a single token $e_{\text{audio}}\in\mathbb{R}^{768}$:
\begin{enumerate}[leftmargin=*,itemsep=2pt,topsep=2pt]
  \item Linear projection $W_{\text{in}}\in\mathbb{R}^{2304\times 512}$ to a hidden space.
  \item A residual block: LayerNorm $\to$ Linear $\to$ GELU $\to$ Dropout (rate $0.05$).
  \item Linear projection $W_{\text{out}}\in\mathbb{R}^{768\times 2304}$ to the CLIP token
        space.
  \item $\ell_{2}$-normalisation.
  \item A lightweight attention refinement module (FGA), which performs a single
        multi-head attention pass over the normalised token using a small set of
        learnable CLIP-space queries.
\end{enumerate}
\noindent
The embedder has $\sim$8.3M trainable parameters; both CLAP and the diffusion backbone are
frozen. The BEATs embedder is the original AudioToken architecture (attention pooling +
MLP) and is left unchanged in its mechanics but retrained on our $50$-class subset.

\vspace{1mm} \noindent
\textbf{Training.} We train the embedder with the standard diffusion denoising loss on paired (audio, image)
data. Optimiser: AdamW, learning rate $1{\times}10^{-5}$, weight decay $1{\times}10^{-2}$,
batch size $4$. We train for $30,000$ steps on the $50$-class VGGSound subset ($\sim$25K pairs). Training takes $\sim$$36$ hours on a single NVIDIA A100 (40~GB).

\vspace{1mm} \noindent
\textbf{Multi-token registration (MTE).} Register five placeholder tokens $\mathcal{T} = \{\langle\!*\!\rangle, \langle\hat{\,}\rangle, \langle!\rangle,\langle@\rangle, \langle\$\rangle\}$ in the CLIP tokenizer (handling $n\leq 5$). The text-encoder embedding table is resized accordingly, and the new rows are zero-initialized and overwritten with audio-derived embeddings at inference time. No fine-tuning of the text encoder is performed.

\vspace{1mm} \noindent
\textbf{Inference-time optimization hyperparameters.}
Following~\cite{attendexcite}, we apply optimisation in the first $T_{\max}=25$ of
$50$ denoising steps. At steps $\{0,10,20\}$ we additionally run an iterative
refinement of up to $20$ inner iterations, with per-token activation targets of
$\{0.05,\,0.5,\,0.8\}$ respectively. The latent step size has a base value of
$20.0$ and is scaled by $\sqrt{s_i}$, where $s_i$ decreases linearly from $1.0$ to
$0.5$ across the schedule. The mixed-loss weighting is
$(\lambda_{\text{AE}},\lambda_{\text{seg}},\lambda_{\text{ret}})=(1.0,\,2.0,\,0.5)$.
Cross-attention maps are collected from the $16\times16$ resolution U-Net layers
and averaged across heads.

\begin{figure*}[!t]
     \centering
     \begin{minipage}{0.45\linewidth}
         \centering
         \includegraphics[width=\linewidth]{cluster_plots/highlight_elk_bugling.png}\vspace{-0.2em}
         {\scriptsize (a) Elk Bugling.}
     \end{minipage}
     \begin{minipage}{0.45\linewidth}
         \centering
         \includegraphics[width=\linewidth]{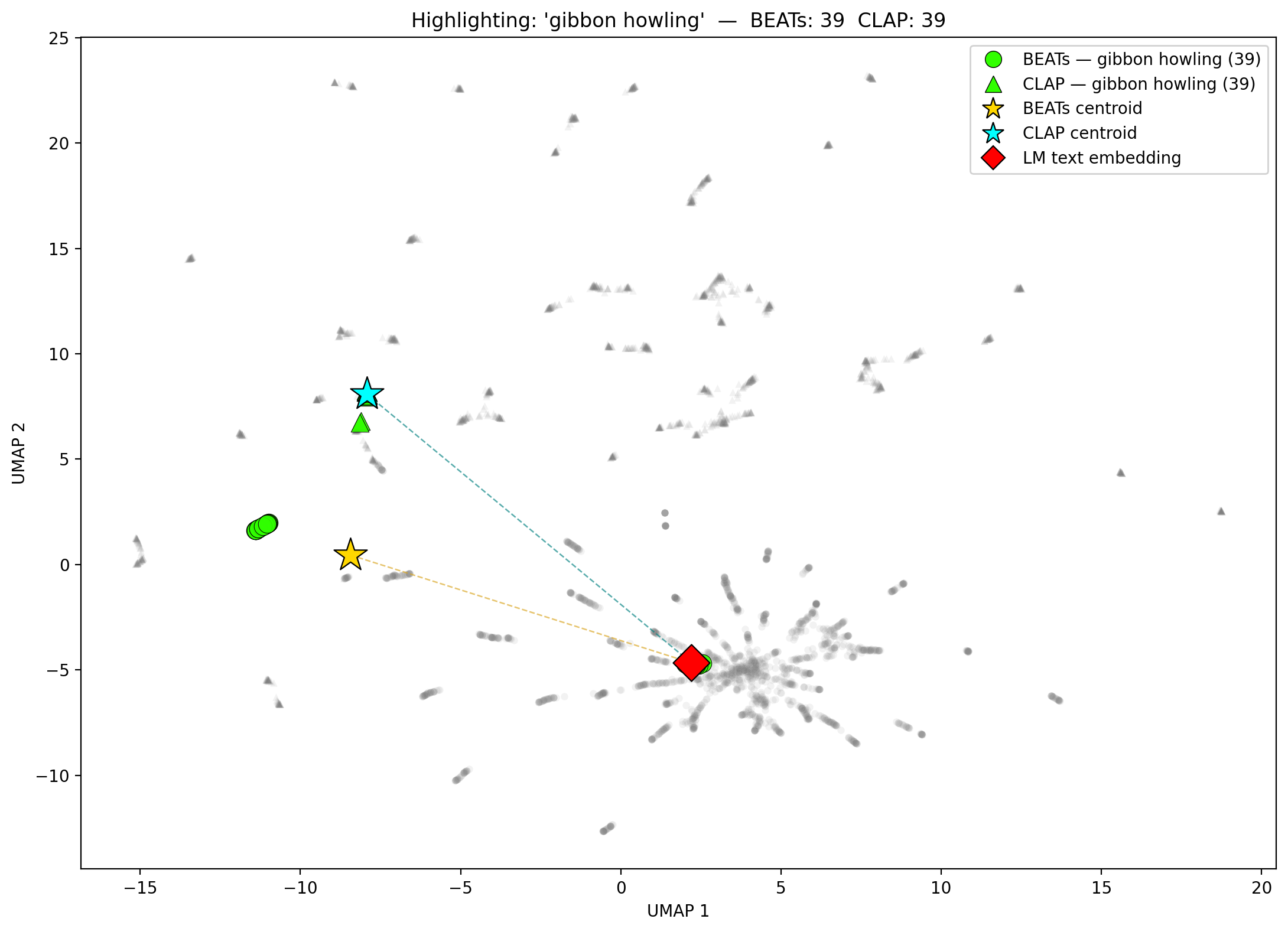}\vspace{-0.2em}
         {\scriptsize (b) Gibbon Howling.}
     \end{minipage}
     \begin{minipage}{0.45\linewidth}
         \centering
         \includegraphics[width=\linewidth]{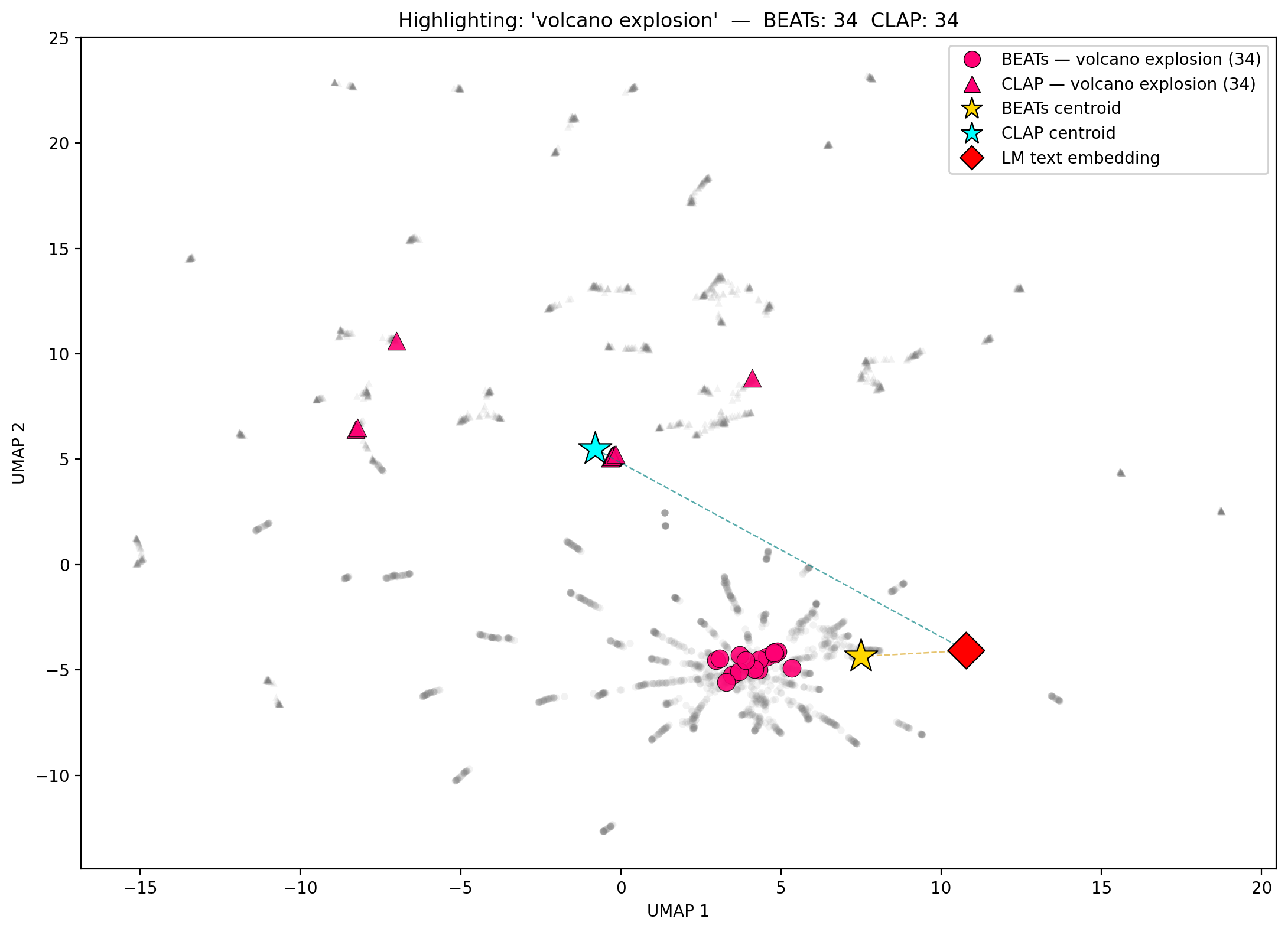}\vspace{-0.2em}
         {\scriptsize (c) Volcano Explosion.}
     \end{minipage}
     \caption{\textbf{UMAP projections of BEATs and CLAP tokens in CLIP space} on the $50$-class VGGSound subset. Circles represent BEATs tokens and triangles represent CLAP tokens. Colored points correspond to the highlighted class, while grey points represent the remaining classes. The gold star denotes the BEATs centroid, the cyan star denotes the CLAP centroid, and the red diamond represents the CLIP text embedding of the class label. CLAP tokens form tighter, better-separated clusters while maintaining angular alignment with the corresponding text embeddings.}
     \label{fig:umap_tokens}
\end{figure*}

\vspace{1mm} \noindent
\textbf{Clustering metrics.} All cluster-geometry metrics (Table~$1$ in the main paper) are computed
on $\ell_{2}$-normalised embeddings using the implementations from \texttt{scikit-learn}
(\texttt{silhouette\_score}, \texttt{davies\_bouldin\_score}). Per-class WCV is computed
as the mean squared distance to the class centroid in $\ell_{2}$-normalized space.
Intra-class cosine similarity is the mean pairwise cosine similarity within each class.
UMAP visualisations (Figures~$4$ and~\ref{fig:umap_tokens}) use
$\texttt{n\_neighbors}=15$, $\texttt{min\_dist}=0.1$, and the cosine metric.

\vspace{1mm} \noindent
\textbf{Benchmark construction.} VGG2 consists of $2500$ groups of two audio--image pairs, each pair drawn from a distinct VGGSound class. VGG3 consists of $1000$ groups of three audio--image pairs, constructed under the same constraints. VGGn consists of $500$ groups of variable size $n\in\{2,3,4,5\}$, where the group size is drawn uniformly before the classes are selected. Within each group, classes are sampled without replacement so that the sources are distinct, and within each class, samples are drawn sequentially so that all clips of that class are exhausted before reuse. All three benchmarks are generated with random seed $42$.

\vspace{1mm} \noindent
\textbf{Metric computation.}
Let $I$ be the generated image for a group of $n$ sources. We denote by
$E^{L}_{\mathrm{img}}$, $E^{B}_{\mathrm{img}}$ the CLIP ViT-L/14 and ViT-B/32
image encoders, $E^{L}_{\mathrm{txt}}$ the CLIP ViT-L/14 text encoder,
$E_{\mathrm{wav}}$ the wav2CLIP~\cite{wav2clip} audio encoder, and
$\mathrm{cap}(\cdot)$ the BLIP~\cite{blip} captioner (one caption per image). All
embeddings are $\ell_2$-normalised and $\langle\cdot,\cdot\rangle$ is cosine
similarity. For slot $o\in\{1,\dots,n\}$, $I^{+}_{o}$, $a^{+}_{o}$ and
$\ell^{+}_{o}$ are the matching reference image, audio and class label, and the
negative banks $\mathcal{B}_{\mathrm{aud}}$ and $\mathcal{B}_{\mathrm{img}}$ (of
sizes $K_A,K_I$) are pooled from the corresponding slots of the \emph{other}
groups in the evaluation batch, excluding the correct item and capped at $256$
entries per slot.

\emph{AIS and IIS} are retrieval win rates. AIS compares the generated image
(ViT-B/32, to match wav2CLIP's joint space) against each input audio relative to
random audios, while IIS compares the generated image (ViT-L/14) against each
reference image relative to random references. We define them as
\begin{equation}
\mathrm{AIS}=\frac{1}{n K_A}\sum_{o=1}^{n}\sum_{a^{-}\in\mathcal{B}_{\mathrm{aud}}}
\mathbb{1}\!\big[\langle E^{B}_{\mathrm{img}}(I),E_{\mathrm{wav}}(a^{+}_{o})\rangle
> \langle E^{B}_{\mathrm{img}}(I),E_{\mathrm{wav}}(a^{-})\rangle\big],
\end{equation}
\begin{equation}
\mathrm{IIS}=\frac{1}{n K_I}\sum_{o=1}^{n}\sum_{I^{-}\in\mathcal{B}_{\mathrm{img}}}
\mathbb{1}\!\big[\langle E^{L}_{\mathrm{img}}(I),E^{L}_{\mathrm{img}}(I^{+}_{o})\rangle
> \langle E^{L}_{\mathrm{img}}(I),E^{L}_{\mathrm{img}}(I^{-})\rangle\big].
\end{equation}

\emph{AIC} is a zero-shot CLIP classifier over the $50$ curated classes
$\mathcal{C}$ with the prompt ``a photo of a $c$''. Writing $\mathrm{Top}_{n}$ for
the $n$ highest-scoring classes and $\mathcal{L}=\{\ell^{+}_{1},\dots,\ell^{+}_{n}\}$
for the ground-truth labels, AIC is the recall
\begin{equation}
\mathrm{AIC}=\frac{1}{n}\,\big|\,\mathcal{L}\cap\mathrm{Top}_{n}\,\big|,
\qquad
\mathrm{Top}_{n}=\operatorname*{arg\,top\text{-}n}_{c\in\mathcal{C}}\;
\langle E^{L}_{\mathrm{img}}(I),E^{L}_{\mathrm{txt}}(c)\rangle,
\end{equation}
which reduces to top-1 accuracy when $n=1$.

\emph{CLS and CCS} aggregate caption similarities with a geometric mean over the
$n$ slots, using $\epsilon=10^{-6}$. CLS compares the generated-image caption(s)
$\mathcal{K}=\mathrm{cap}(I)$ to each class label, while CCS compares them instead
to the BLIP caption $r_{o}=\mathrm{cap}(I^{+}_{o})$ of each reference image, giving
\begin{equation}
\mathrm{CLS}=\exp\!\Bigg(\frac{1}{n}\sum_{o=1}^{n}\log\max\!\Big(\epsilon,\,
\max_{k\in\mathcal{K}}\langle E^{L}_{\mathrm{txt}}(k),E^{L}_{\mathrm{txt}}(\ell^{+}_{o})\rangle\Big)\!\Bigg),
\end{equation}
\begin{equation}
\mathrm{CCS}=\exp\!\Bigg(\frac{1}{n}\sum_{o=1}^{n}\log\max\!\Big(\epsilon,\,
\max_{k\in\mathcal{K}}\langle E^{L}_{\mathrm{txt}}(k),E^{L}_{\mathrm{txt}}(r_{o})\rangle\Big)\!\Bigg).
\end{equation}

\emph{FID}~\cite{fid} is accumulated across the whole benchmark rather than
computed per group. Each group contributes its $n$ reference images to the real
set and, paired with each reference, its single generated image to the generated
set, so the generated image enters with multiplicity $n$ and the two sets stay
balanced (real$_1$, gen, real$_2$, gen, $\dots$). The Fréchet distance is then
computed once from the InceptionV3 statistics accumulated over all groups, with
lower values indicating closer distributions.

\vspace{1mm} \noindent
\textbf{Baseline reproduction.} AudioToken~\cite{audiotoken}, SonicDiffusion~\cite{sonicdiffusion},
Sound2Scene~\cite{sound2scene}, and MACS~\cite{macs} are run from their official public implementations with default hyperparameters. For multi-audio evaluation, \textit{AT (309)} uses the released $309$-class AudioToken weights together with our multi-token extension, so that a group of $n$ sources is registered to $n$ placeholder tokens in a single prompt and produces one combined image, exactly as for our own variants. Its embedder architecture is identical to that of our \textbf{AT} variant, so the two differ only in the training vocabulary of the embedder. SonicDiffusion and Sound2Scene accept a single audio input and are therefore reported on VGG only. Since our benchmarks already provide separated sources, we bypass the weakly supervised separation stage of MACS and pass the separated sources directly to its adapter, matching the separated-input assumption shared by all methods and isolating the comparison to the generation stage.

\section{Per-Dataset Ablation Tables}
\label{supp:ablations}

The main paper reports the benchmark-averaged ablation in
Table~$3$. Here we provide the per-dataset breakdown supporting that summary. Best results per column within each table are highlighted in \textcolor{bestcol}{violet}, second-best in \textbf{bold}. ``AT'' refers to our
multi-token AudioToken backbone with the BEATs encoder retrained on the $50$-class subset; ``AT (Clap)'' is the same backbone with the CLAP encoder and redesigned
embedder. ``AnE'' is $\Lae$; ``A$^{*}$'' is $\Lastar$; ``AnE + A$^{*}$'' is the mixed
loss.

\begin{table}[!t]
\centering
\small
\setlength{\tabcolsep}{4pt}
\begin{tabular}{lcccccc}
\toprule
\textbf{Method} & \textbf{IIS}$\uparrow$ & \textbf{AIS}$\uparrow$ & \textbf{AIC}$\uparrow$ & \textbf{CLS}$\uparrow$ & \textbf{CCS}$\uparrow$ & \textbf{FID}$\downarrow$ \\
\midrule
AT                  & 0.7330 & 0.7200 & 0.4337 & 0.5837 & 0.5343 & 70.0351 \\
AT + AnE            & 0.7271 & 0.7226 & 0.4333 & 0.5744 & 0.5275 & 61.5910 \\
AT + A$^{*}$        & 0.7327 & 0.7228 & 0.4347 & 0.5833 & 0.5359 & 70.5935 \\
AT + AnE + A$^{*}$  & 0.7372 & 0.7215 & 0.4477 & 0.5846 & 0.5330 & 62.3452 \\
\midrule
AT (Clap)                 & \textbf{0.8052} & \textcolor{bestcol}{0.7624} & \textbf{0.5788} & \textbf{0.5994} & \textbf{0.5600} & \textcolor{bestcol}{55.9076} \\
AT + AnE (Clap)           & \textcolor{bestcol}{0.8064} & 0.7588 & 0.5737 & 0.5961 & 0.5574 & 60.8501 \\
AT + A$^{*}$ (Clap)       & 0.8044 & \textbf{0.7593} & 0.5723 & \textcolor{bestcol}{0.5995} & \textcolor{bestcol}{0.5606} & 61.6084 \\
AT + AnE + A$^{*}$ (Clap) & 0.8047 & 0.7484 & \textcolor{bestcol}{0.5904} & 0.5878 & 0.5496 & \textbf{60.0736} \\
\bottomrule
\end{tabular}
\vspace{0.4em}
\caption{Per-dataset ablation on VGGSound (VGG), single audio.}
\label{tab:ablation_vgg}
\end{table}

\begin{table}[!t]
\centering
\small
\setlength{\tabcolsep}{4pt}
\begin{tabular}{lcccccc}
\toprule
\textbf{Method} & \textbf{IIS}$\uparrow$ & \textbf{AIS}$\uparrow$ & \textbf{AIC}$\uparrow$ & \textbf{CLS}$\uparrow$ & \textbf{CCS}$\uparrow$ & \textbf{FID}$\downarrow$ \\
\midrule
AT                  & 0.6553 & 0.6550 & 0.3822 & 0.5312 & 0.4929 & 60.4968 \\
AT + AnE            & 0.6638 & 0.6537 & 0.3758 & 0.5153 & 0.4826 & \textcolor{bestcol}{53.4250} \\
AT + A$^{*}$        & 0.6563 & 0.6531 & 0.3770 & \textcolor{bestcol}{0.5349} & 0.4936 & 60.9600 \\
AT + AnE + A$^{*}$  & 0.6701 & \textbf{0.6611} & 0.3926 & 0.5315 & \textbf{0.4938} & 55.5968 \\
\midrule
AT (Clap)                 & 0.6901 & 0.6463 & 0.4010 & 0.5212 & 0.4872 & 57.0914 \\
AT + AnE (Clap)           & \textbf{0.6907} & 0.6552 & \textbf{0.4112} & 0.5125 & 0.4826 & 55.3890 \\
AT + A$^{*}$ (Clap)       & 0.6793 & 0.6492 & 0.4086 & \textbf{0.5320} & \textcolor{bestcol}{0.4976} & 58.1220 \\
AT + AnE + A$^{*}$ (Clap) & \textcolor{bestcol}{0.6993} & \textcolor{bestcol}{0.6677} & \textcolor{bestcol}{0.4332} & 0.5237 & 0.4902 & \textbf{54.1177} \\
\bottomrule
\end{tabular}
\vspace{0.4em}
\caption{Per-dataset ablation on VGG2, two audios.}
\label{tab:ablation_vgg2}
\end{table}

\begin{table}[!t]
\centering
\small
\setlength{\tabcolsep}{4pt}
\begin{tabular}{lcccccc}
\toprule
\textbf{Method} & \textbf{IIS}$\uparrow$ & \textbf{AIS}$\uparrow$ & \textbf{AIC}$\uparrow$ & \textbf{CLS}$\uparrow$ & \textbf{CCS}$\uparrow$ & \textbf{FID}$\downarrow$ \\
\midrule
AT                  & 0.7490 & 0.6174 & \textbf{0.2110} & \textbf{0.5205} & \textcolor{bestcol}{0.5101} & 69.6100 \\
AT + AnE            & 0.7462 & 0.6042 & 0.2006 & 0.5032 & 0.4966 & \textbf{66.4122} \\
AT + A$^{*}$        & \textcolor{bestcol}{0.7531} & 0.6156 & 0.2108 & \textcolor{bestcol}{0.5237} & \textcolor{bestcol}{0.5101} & 68.7482 \\
AT + AnE + A$^{*}$  & \textbf{0.7520} & \textbf{0.6188} & \textcolor{bestcol}{0.2131} & 0.5079 & \textbf{0.5035} & \textcolor{bestcol}{66.4075} \\
\midrule
AT (Clap)                 & 0.7206 & 0.5980 & 0.1642 & 0.5096 & 0.4969 & 69.5848 \\
AT + AnE (Clap)           & 0.7150 & 0.6050 & 0.1574 & 0.4907 & 0.4840 & 71.5160 \\
AT + A$^{*}$ (Clap)       & 0.7189 & 0.6010 & 0.1674 & 0.5120 & 0.4961 & 69.5840 \\
AT + AnE + A$^{*}$ (Clap) & 0.7291 & \textcolor{bestcol}{0.6221} & 0.1688 & 0.4983 & 0.4914 & 69.0440 \\
\bottomrule
\end{tabular}
\vspace{0.4em}
\caption{Per-dataset ablation on IS3 (out-of-distribution two-audio benchmark).}
\label{tab:ablation_is3}
\end{table}

\begin{table}[!t]
\centering
\small
\setlength{\tabcolsep}{4pt}
\begin{tabular}{lcccccc}
\toprule
\textbf{Method} & \textbf{IIS}$\uparrow$ & \textbf{AIS}$\uparrow$ & \textbf{AIC}$\uparrow$ & \textbf{CLS}$\uparrow$ & \textbf{CCS}$\uparrow$ & \textbf{FID}$\downarrow$ \\
\midrule
AT                  & 0.6237 & \textbf{0.6215} & 0.3523 & 0.5147 & 0.4783 & 76.3441 \\
AT + AnE            & 0.6234 & 0.6100 & 0.3527 & 0.5005 & 0.4676 & 75.0173 \\
AT + A$^{*}$        & 0.6238 & 0.6125 & 0.3537 & 0.5160 & 0.4551 & 77.1600 \\
AT + AnE + A$^{*}$  & 0.6298 & 0.6210 & 0.3527 & 0.5111 & 0.4742 & 74.8436 \\
\midrule
AT (Clap)                 & 0.6387 & 0.6196 & 0.3533 & \textbf{0.5164} & \textbf{0.4830} & \textbf{74.5290} \\
AT + AnE (Clap)           & \textbf{0.6425} & 0.6162 & \textbf{0.3630} & 0.4924 & 0.4659 & 79.9670 \\
AT + A$^{*}$ (Clap)       & 0.6395 & 0.6155 & 0.3533 & \textcolor{bestcol}{0.5206} & \textcolor{bestcol}{0.4841} & 75.8300 \\
AT + AnE + A$^{*}$ (Clap) & \textcolor{bestcol}{0.6528} & \textcolor{bestcol}{0.6266} & \textcolor{bestcol}{0.3703} & 0.5026 & 0.4730 & \textcolor{bestcol}{74.0027} \\
\bottomrule
\end{tabular}
\vspace{0.4em}
\caption{Per-dataset ablation on VGG3, three audios.}
\label{tab:ablation_vgg3}
\vspace{-0.9em}
\end{table}

\begin{table}[!t]
\centering
\small
\setlength{\tabcolsep}{4pt}
\begin{tabular}{lcccccc}
\toprule
\textbf{Method} & \textbf{IIS}$\uparrow$ & \textbf{AIS}$\uparrow$ & \textbf{AIC}$\uparrow$ & \textbf{CLS}$\uparrow$ & \textbf{CCS}$\uparrow$ & \textbf{FID}$\downarrow$ \\
\midrule
AT                  & 0.6094 & 0.5946 & 0.3366 & \textbf{0.5040} & \textbf{0.4697} & 78.1004 \\
AT + AnE            & 0.6084 & 0.5936 & 0.3343 & 0.4953 & 0.4637 & \textcolor{bestcol}{74.8646} \\
AT + A$^{*}$        & 0.6055 & \textcolor{bestcol}{0.5982} & 0.3313 & \textcolor{bestcol}{0.5089} & \textcolor{bestcol}{0.4700} & 80.3390 \\
AT + AnE + A$^{*}$  & 0.6124 & \textbf{0.5973} & \textcolor{bestcol}{0.3381} & 0.5026 & 0.4684 & \textbf{75.8900} \\
\midrule
AT (Clap)                 & \textcolor{bestcol}{0.6186} & 0.5888 & 0.3308 & 0.4945 & 0.4654 & 78.4695 \\
AT + AnE (Clap)           & 0.6135 & 0.5827 & \textbf{0.3368} & 0.4840 & 0.4590 & 81.0473 \\
AT + A$^{*}$ (Clap)       & 0.6122 & 0.5850 & 0.3298 & 0.5003 & 0.4685 & 80.1423 \\
AT + AnE + A$^{*}$ (Clap) & \textbf{0.6183} & 0.5898 & \textcolor{bestcol}{0.3381} & 0.4931 & 0.4661 & 76.9760 \\
\bottomrule
\end{tabular}
\vspace{0.4em}
\caption{Per-dataset ablation on VGGn, variable $n\in\{2,3,4,5\}$.}
\label{tab:ablation_vggn}
\end{table}

\section{Additional Qualitative Results}
\label{supp:qualitative}

We provide qualitative comparisons across the four input regimes considered in the main
paper. Each figure uses identical layout: column headers list method variants
(ours~$\to$~baselines, left to right), and a small line below each image indicates the
audio prompt for that row. The compositional fidelity gap is most pronounced in the two-
and three-audio cases.

\begin{figure*}[!t]
\centering
\setlength{\tabcolsep}{1.5pt}
\renewcommand{\arraystretch}{1.05}
\begin{tabular}{ccccc}
\textbf{\small Ours} & \textbf{\small AT Mixed} & \textbf{\small AT} & \textbf{\small AT (309)} & \textbf{\small MACS} \\
\includegraphics[width=0.19\textwidth]{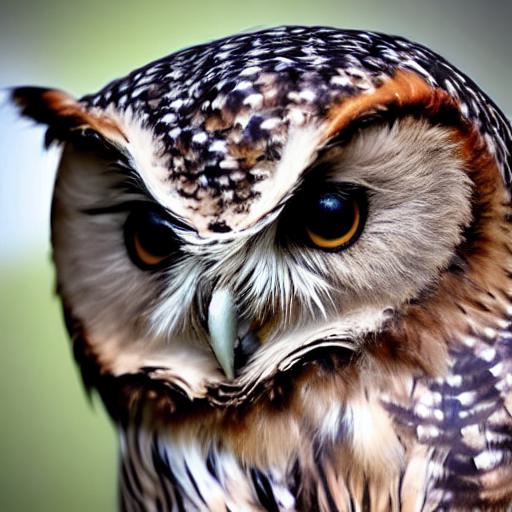} &
\includegraphics[width=0.19\textwidth]{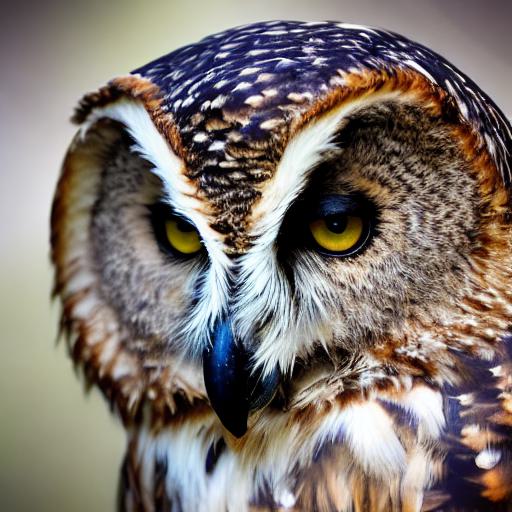} &
\includegraphics[width=0.19\textwidth]{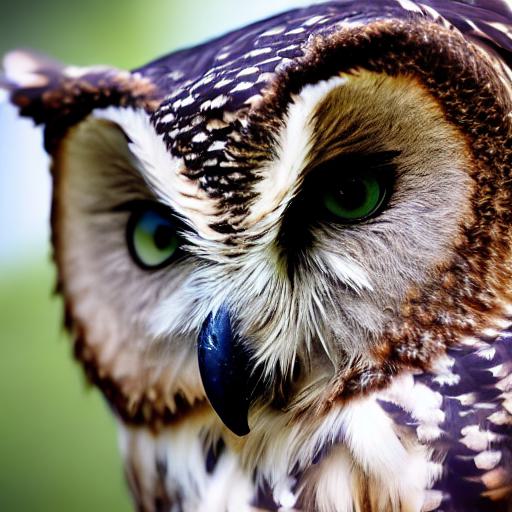} &
\includegraphics[width=0.19\textwidth]{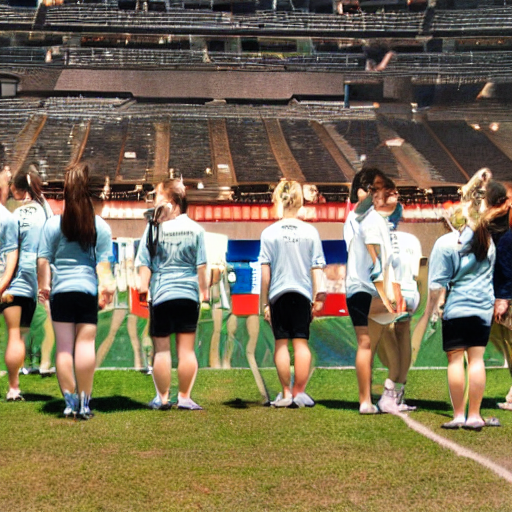} &
\includegraphics[width=0.19\textwidth]{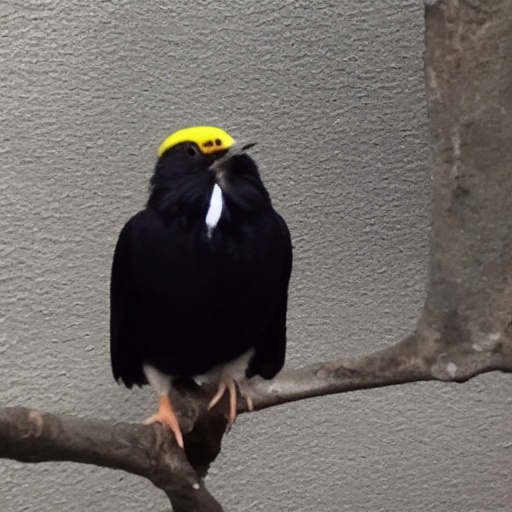} \\
\multicolumn{5}{c}{\scriptsize\textit{Owl Hooting}}\\[3pt]
\includegraphics[width=0.19\textwidth]{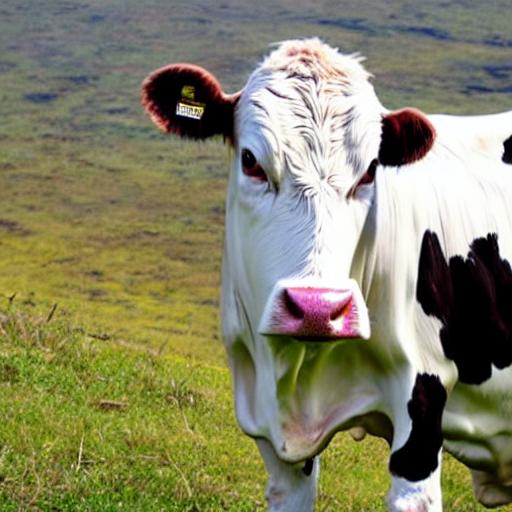} &
\includegraphics[width=0.19\textwidth]{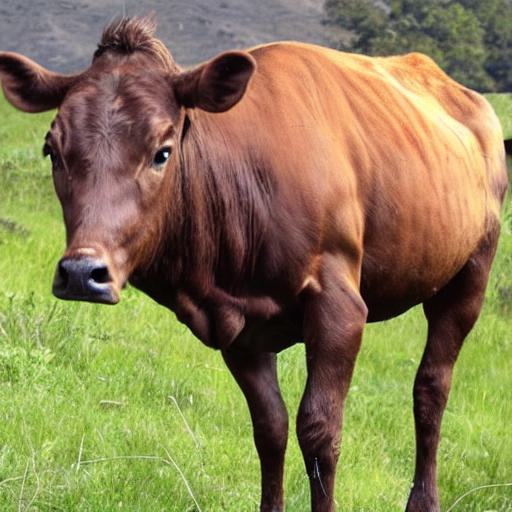} &
\includegraphics[width=0.19\textwidth]{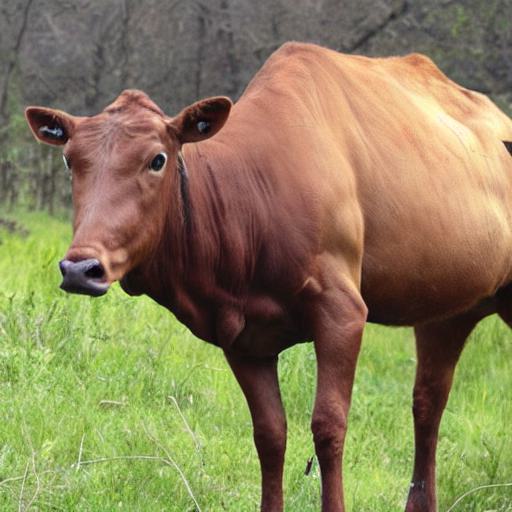} &
\includegraphics[width=0.19\textwidth]{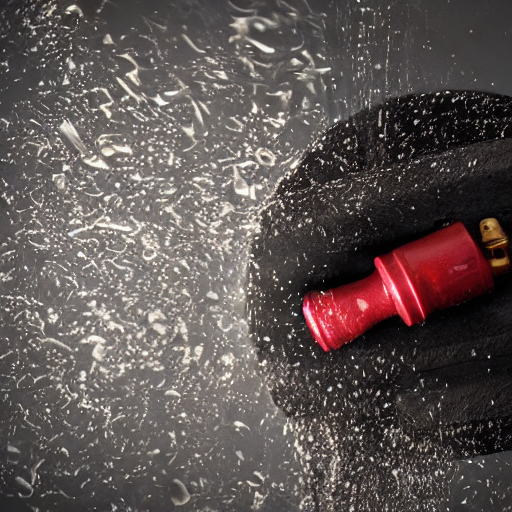} &
\includegraphics[width=0.19\textwidth]{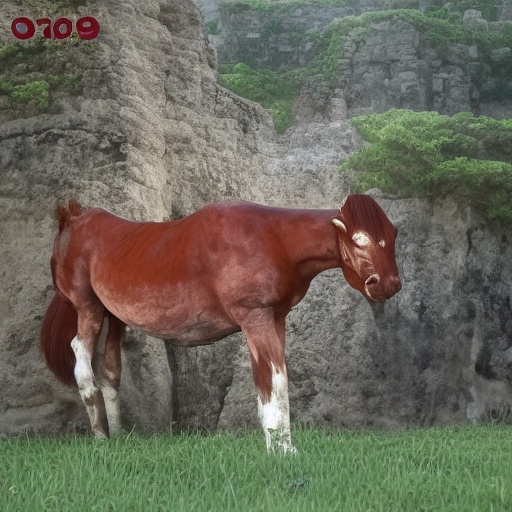} \\
\multicolumn{5}{c}{\scriptsize\textit{Cow Lowing}}\\[3pt]
\includegraphics[width=0.19\textwidth]{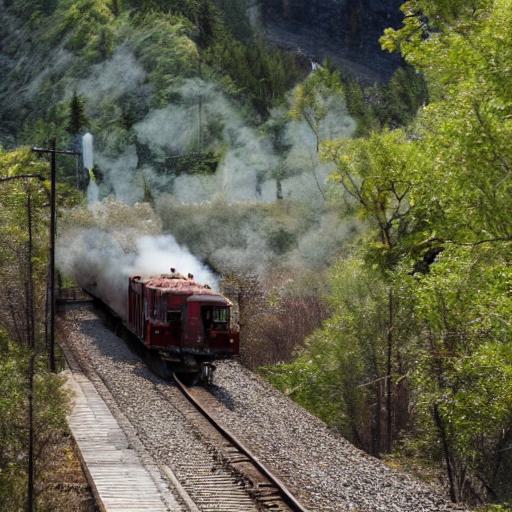} &
\includegraphics[width=0.19\textwidth]{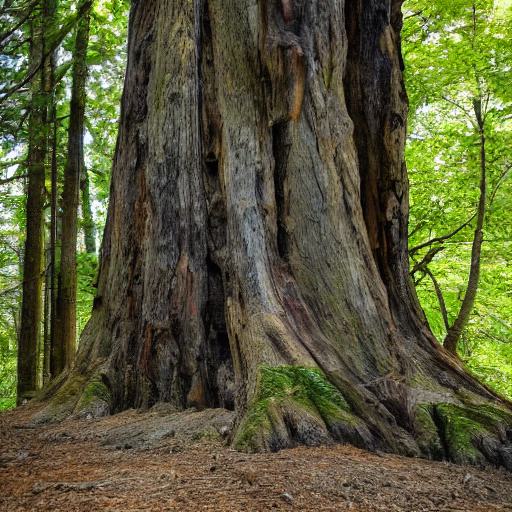} &
\includegraphics[width=0.19\textwidth]{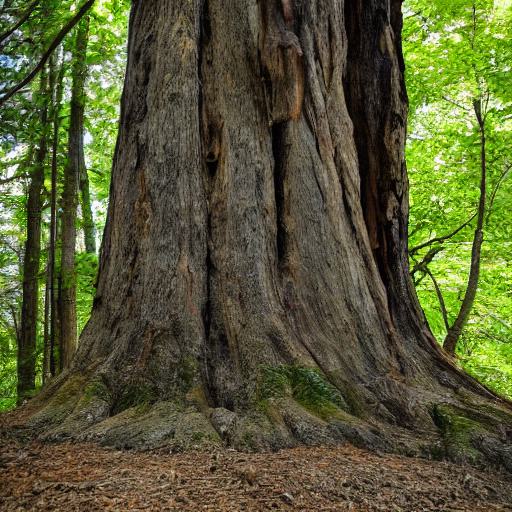} &
\includegraphics[width=0.19\textwidth]{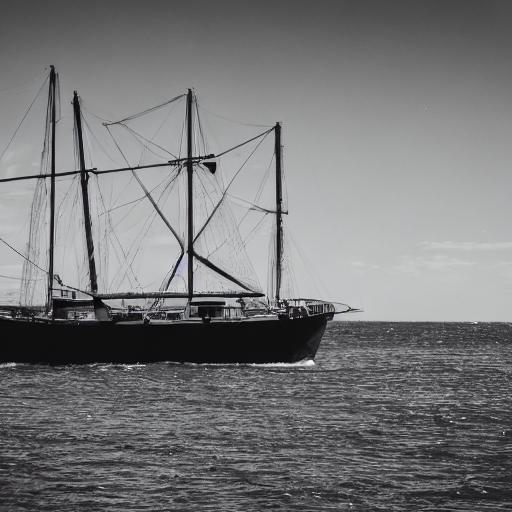} &
\includegraphics[width=0.19\textwidth]{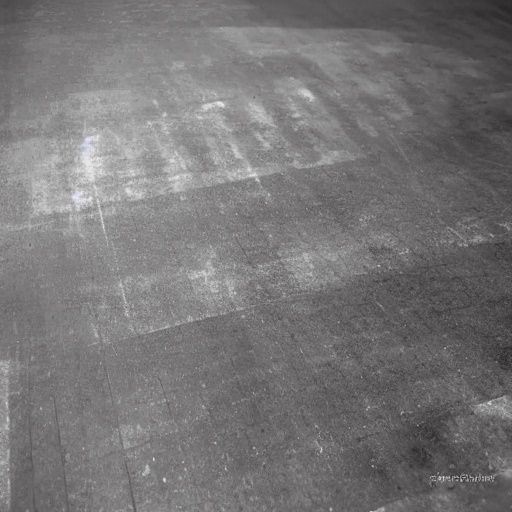} \\
\multicolumn{5}{c}{\scriptsize\textit{Train Whistling}} \\
\end{tabular}
\caption{Qualitative comparison for \textbf{single-audio} inputs (VGG).}
\label{fig:results_single}
\end{figure*}

\begin{figure*}[!t]
\centering
\setlength{\tabcolsep}{1.5pt}
\renewcommand{\arraystretch}{1.05}
\begin{tabular}{ccccc}
\textbf{\small Ours} & \textbf{\small AT Mixed} & \textbf{\small AT} & \textbf{\small AT (309)} & \textbf{\small MACS} \\
\includegraphics[width=0.19\textwidth]{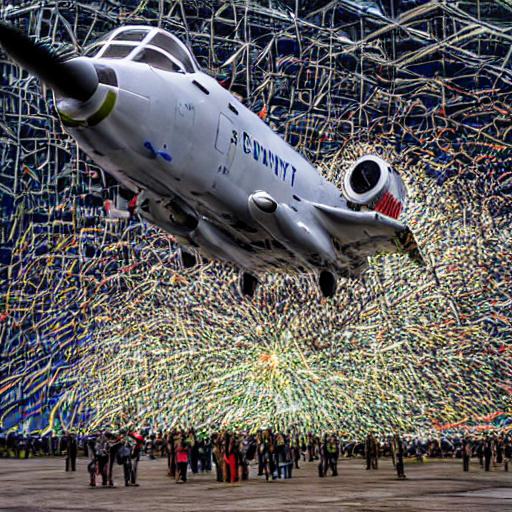} &
\includegraphics[width=0.19\textwidth]{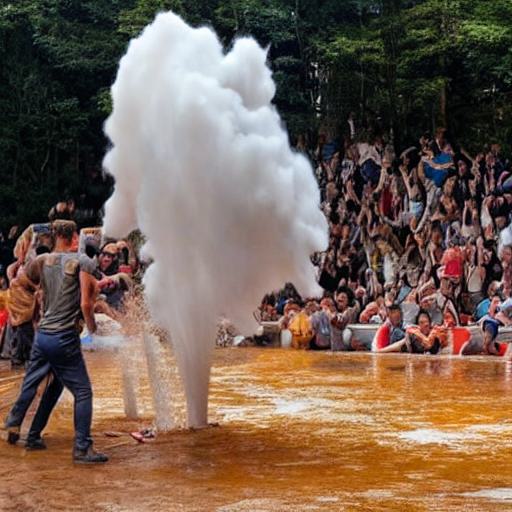} &
\includegraphics[width=0.19\textwidth]{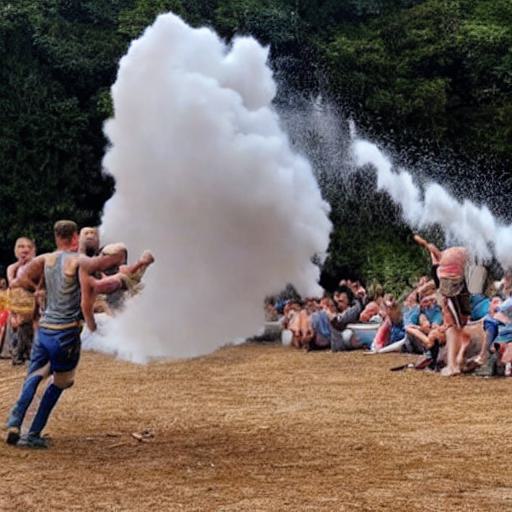} &
\includegraphics[width=0.19\textwidth]{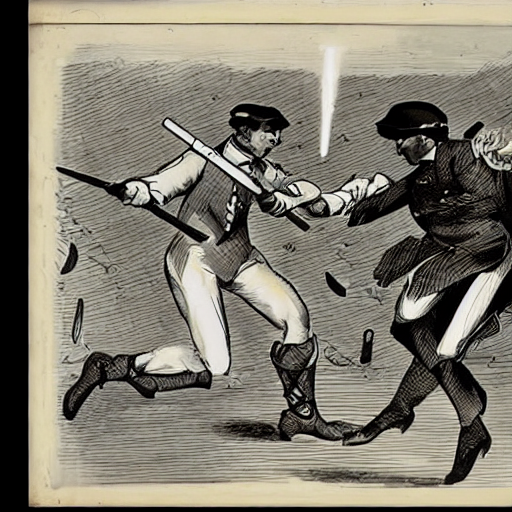} &
\includegraphics[width=0.19\textwidth]{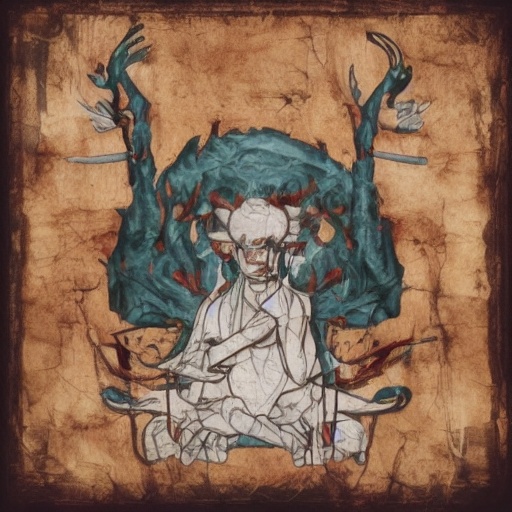} \\
\multicolumn{5}{c}{\scriptsize\textit{Airplane Flyby \& People Cheering}}\\[3pt]
\includegraphics[width=0.19\textwidth]{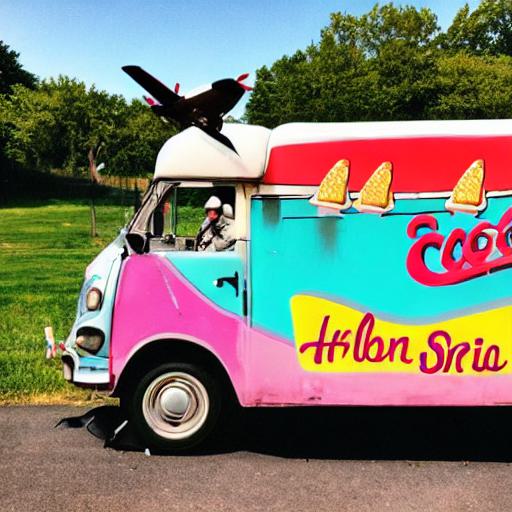} &
\includegraphics[width=0.19\textwidth]{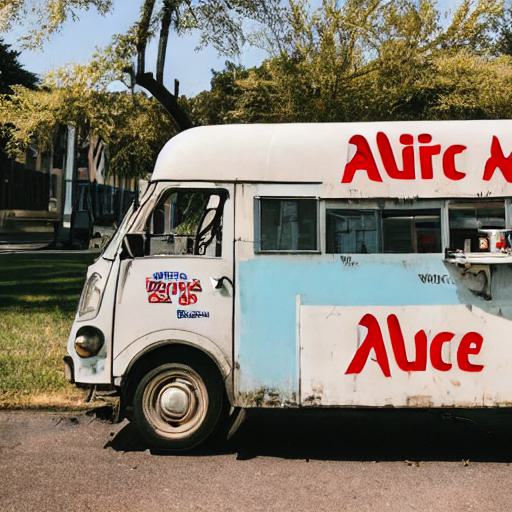} &
\includegraphics[width=0.19\textwidth]{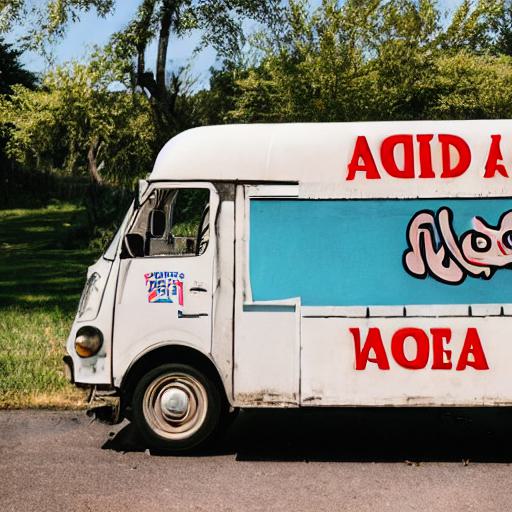} &
\includegraphics[width=0.19\textwidth]{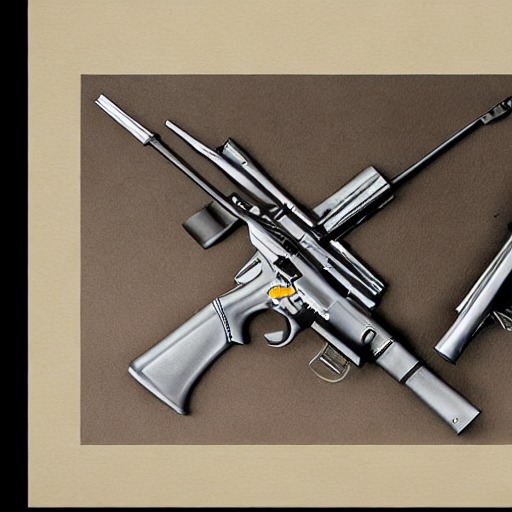} &
\includegraphics[width=0.19\textwidth]{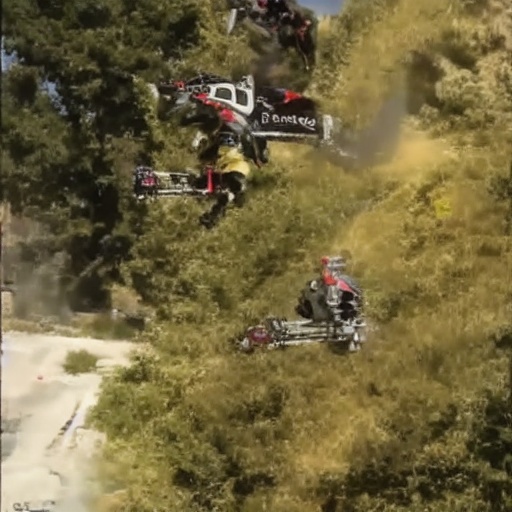} \\
\multicolumn{5}{c}{\scriptsize\textit{Airplane Flyby \& Ice Cream Truck}}\\
\end{tabular}
\caption{Qualitative comparison for two-audio inputs (VGG2).}
\label{fig:results_double}
\end{figure*}

\begin{figure*}[!t]
\centering
\setlength{\tabcolsep}{1.5pt}
\renewcommand{\arraystretch}{1.05}
\begin{tabular}{ccccc}
\textbf{\small Ours} & \textbf{\small AT Mixed} & \textbf{\small AT} & \textbf{\small AT (309)} & \textbf{\small MACS} \\
\includegraphics[width=0.19\textwidth]{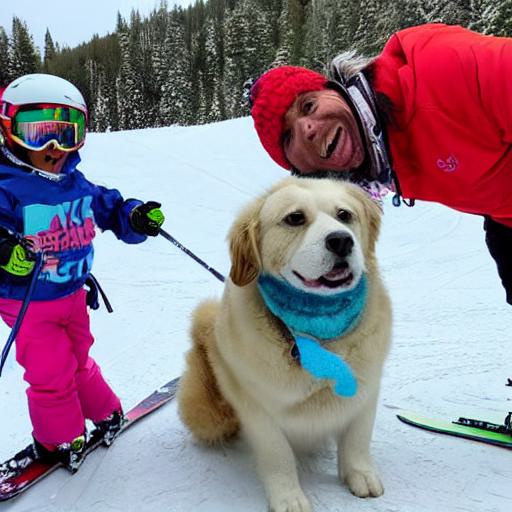} &
\includegraphics[width=0.19\textwidth]{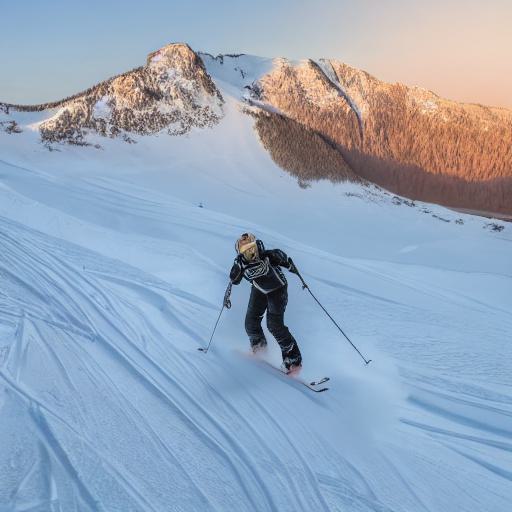} &
\includegraphics[width=0.19\textwidth]{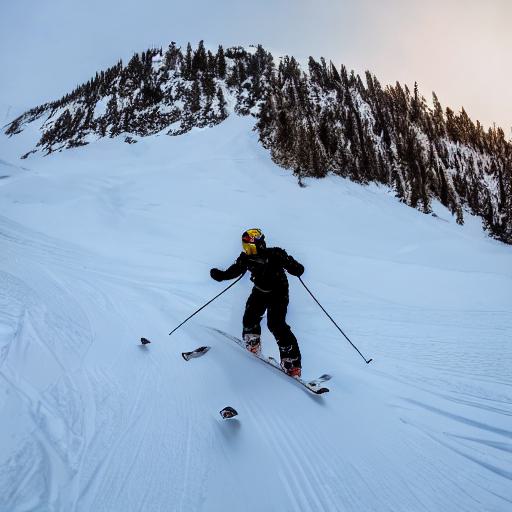} &
\includegraphics[width=0.19\textwidth]{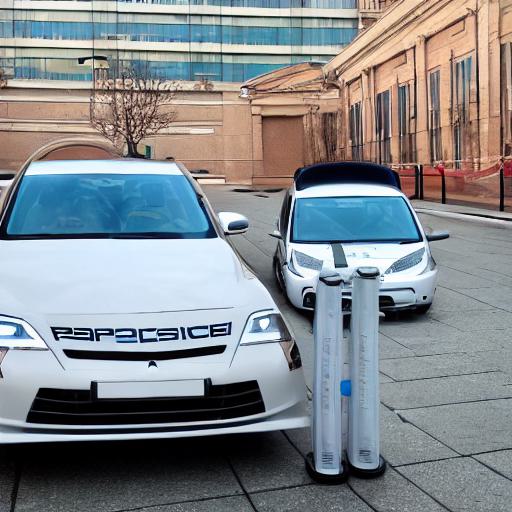} &
\includegraphics[width=0.19\textwidth]{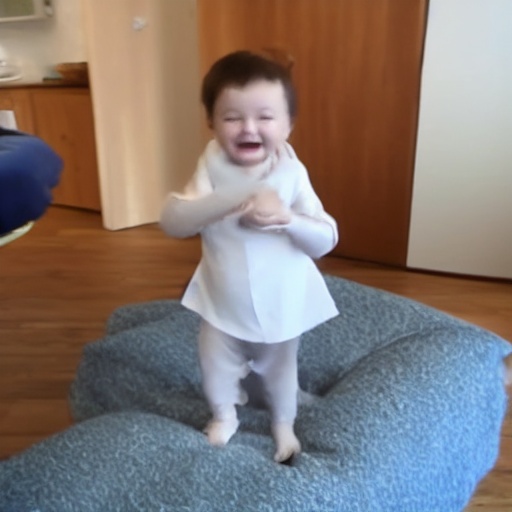} \\
\multicolumn{5}{c}{\scriptsize\textit{Skiing \& Baby Crying \& Dog Barking}}\\[3pt]
\includegraphics[width=0.19\textwidth]{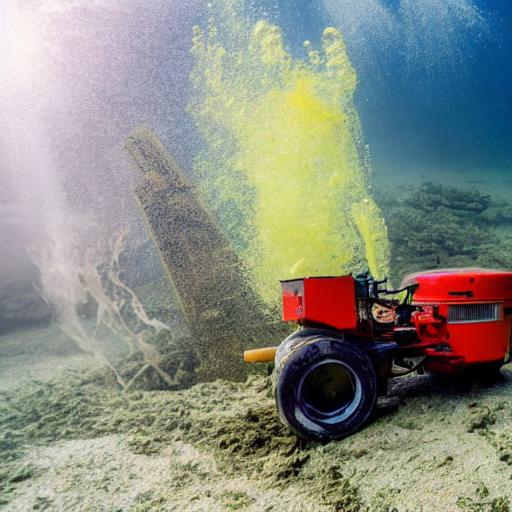} &
\includegraphics[width=0.19\textwidth]{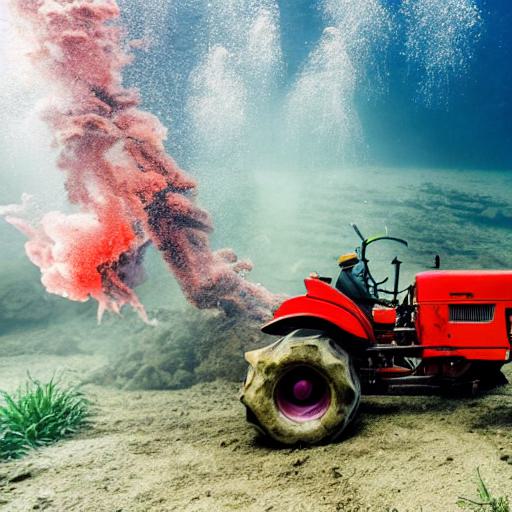} &
\includegraphics[width=0.19\textwidth]{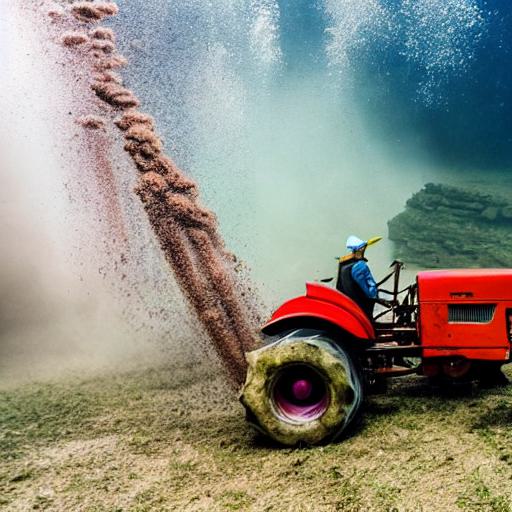} &
\includegraphics[width=0.19\textwidth]{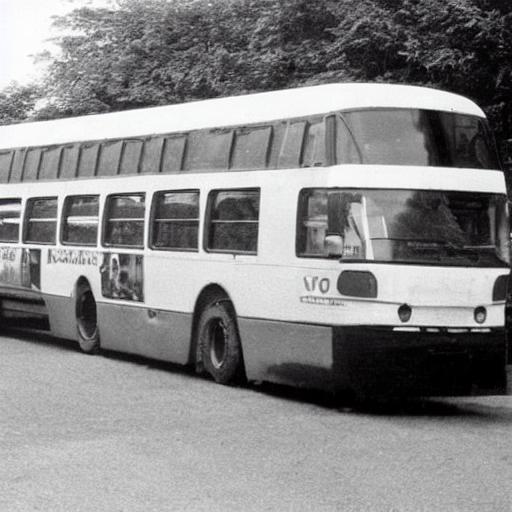} &
\includegraphics[width=0.19\textwidth]{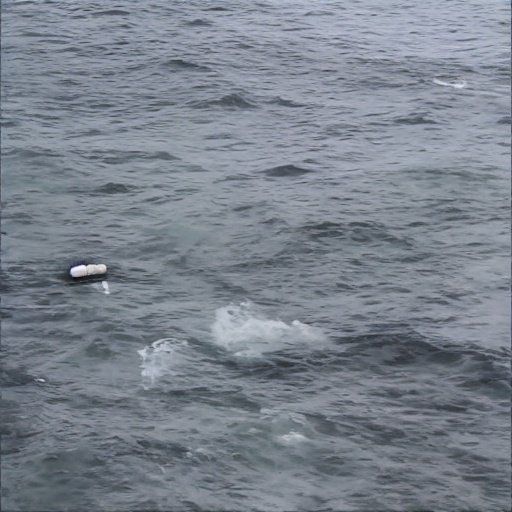} \\
\multicolumn{5}{c}{\scriptsize\textit{Tractor Digging \& Scuba Diving \& Volcano Explosion}}\\
\includegraphics[width=0.19\textwidth]{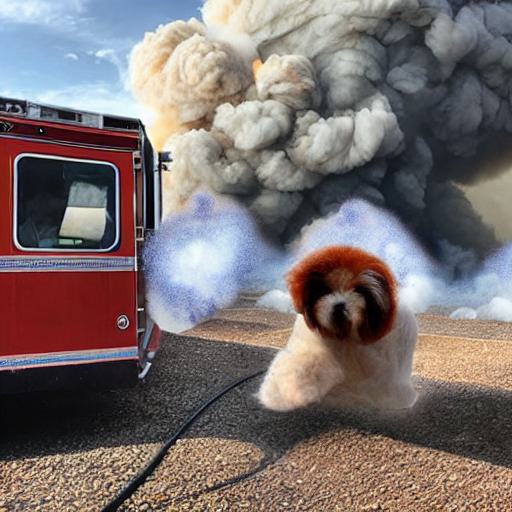} &
\includegraphics[width=0.19\textwidth]{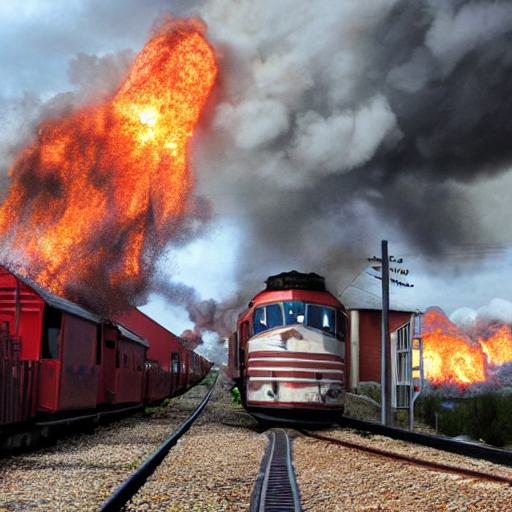} &
\includegraphics[width=0.19\textwidth]{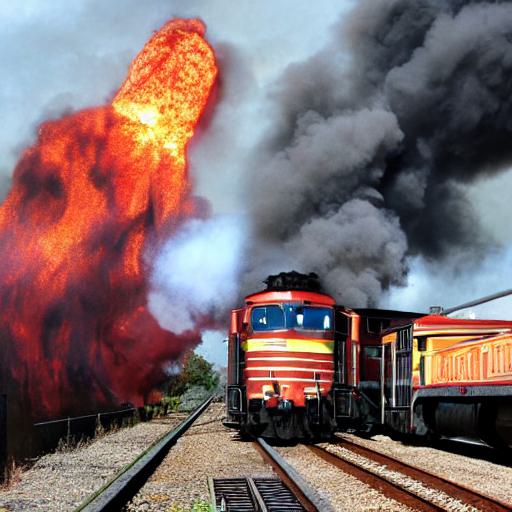} &
\includegraphics[width=0.19\textwidth]{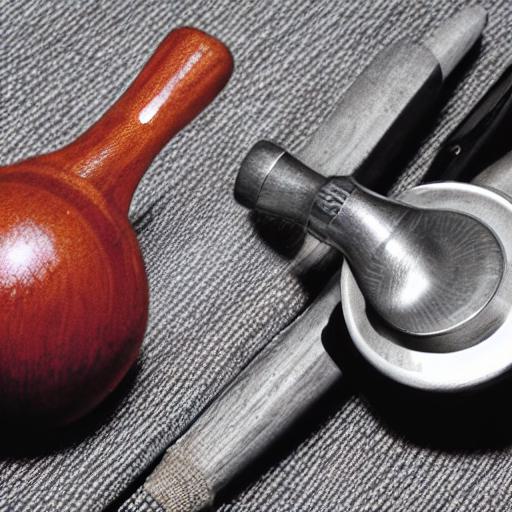} &
\includegraphics[width=0.19\textwidth]{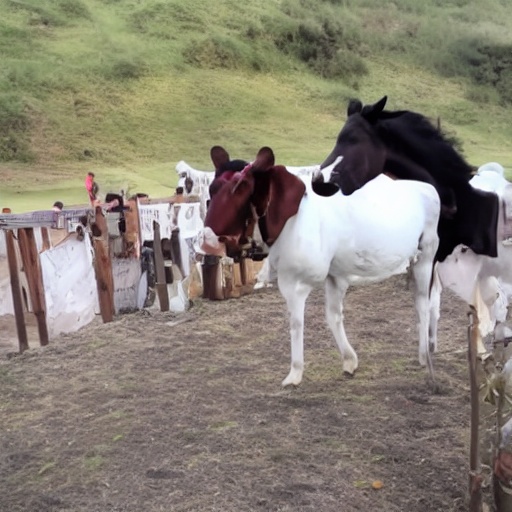} \\
\multicolumn{5}{c}{\scriptsize\textit{Train Horning \& Dog Barking \& Volcano Explosion}}\\
\end{tabular}
\caption{Qualitative comparison for \textbf{three-audio} inputs (VGG3).}
\label{fig:results_triple}
\end{figure*}

\begin{figure*}[!t]
\centering
\setlength{\tabcolsep}{1.5pt}
\renewcommand{\arraystretch}{1.05}
\begin{tabular}{ccccc}
\textbf{\small Ours} & \textbf{\small AT Mixed} & \textbf{\small AT} & \textbf{\small AT (309)} & \textbf{\small MACS} \\
\includegraphics[width=0.19\textwidth]{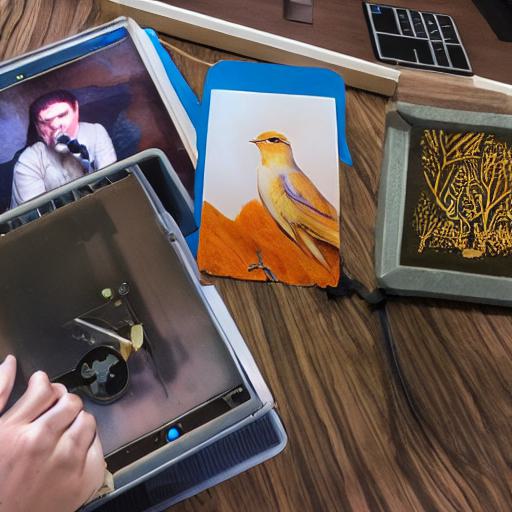} &
\includegraphics[width=0.19\textwidth]{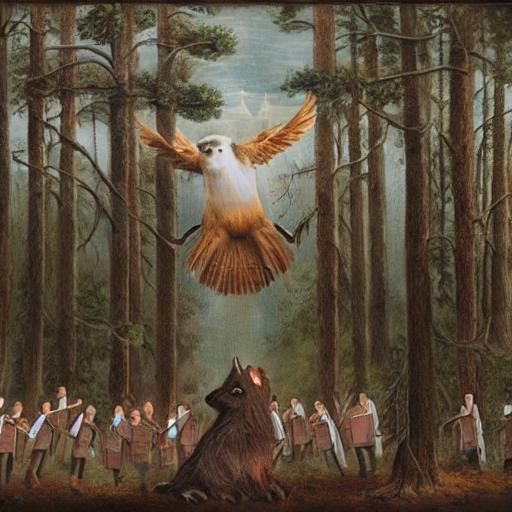} &
\includegraphics[width=0.19\textwidth]{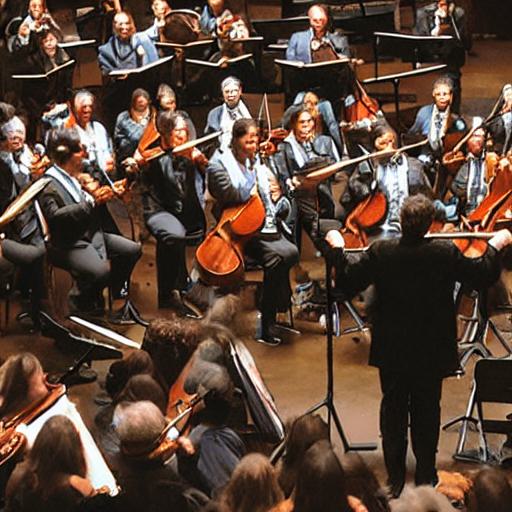} &
\includegraphics[width=0.19\textwidth]{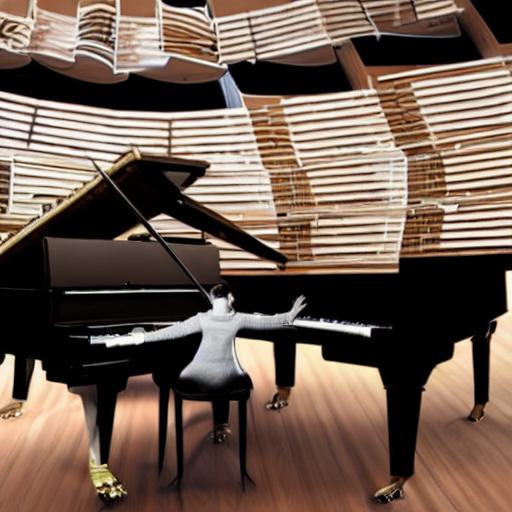} &
\includegraphics[width=0.19\textwidth]{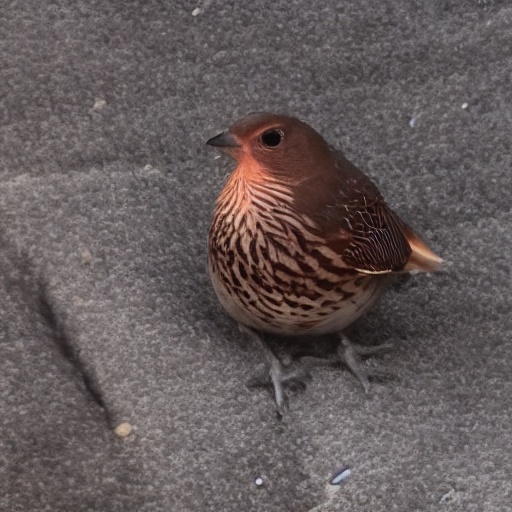} \\
\multicolumn{5}{c}{\scriptsize $n{=}4$: \textit{Elk Bugling \& Baltimore Oriole \& Orchestra \& Baby Laughter}}\\[3pt]
\includegraphics[width=0.19\textwidth]{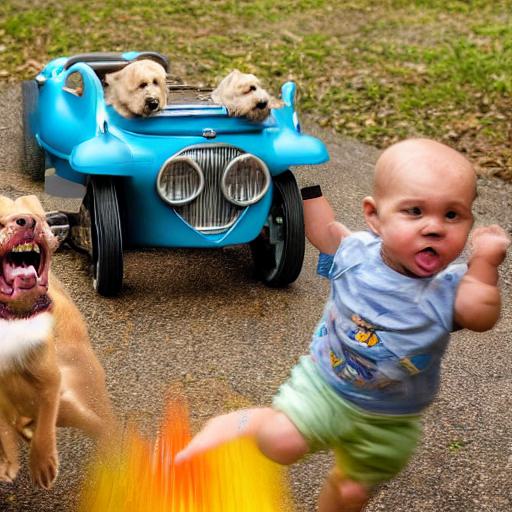} &
\includegraphics[width=0.19\textwidth]{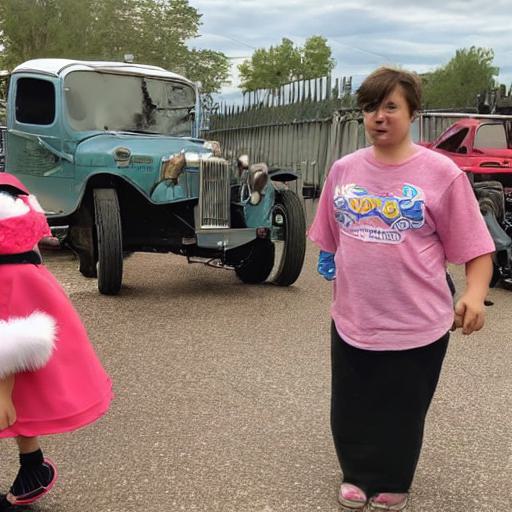} &
\includegraphics[width=0.19\textwidth]{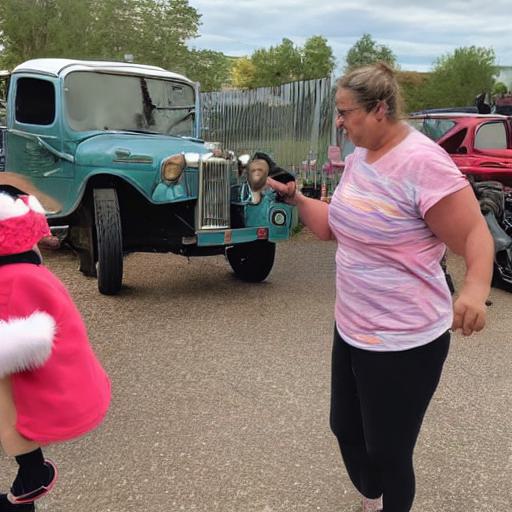} &
\includegraphics[width=0.19\textwidth]{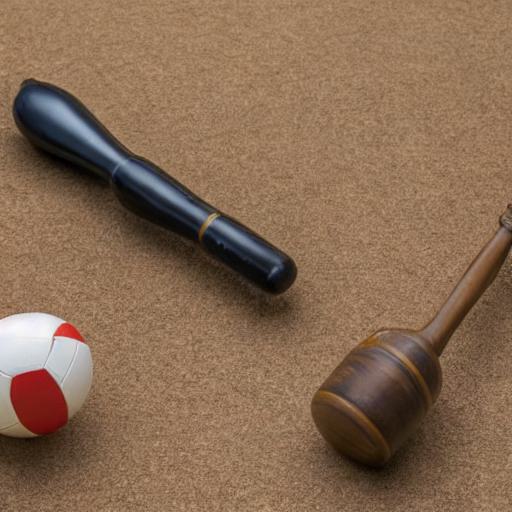} &
\includegraphics[width=0.19\textwidth]{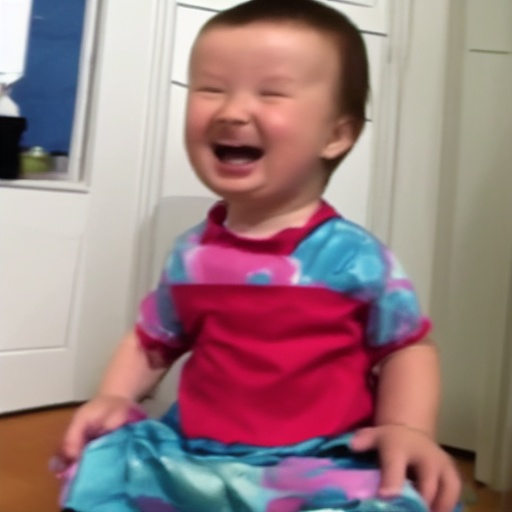} \\
\multicolumn{5}{c}{\scriptsize $n{=}4$: \textit{Dog Barking \& Baby Laughter \& Car Engine Knocking \& Train Horning}}\\
\end{tabular}
\caption{Qualitative comparison for \textbf{large-$n$} inputs (VGGn).}
\label{fig:results_n}
\vspace{-1em}
\end{figure*}

\section{Cross-Attention Visualisations}
\label{supp:attention}

\begin{figure*}[!t]
\centering
\setlength{\tabcolsep}{1.5pt}
\renewcommand{\arraystretch}{1.05}
\begin{tabular}{c@{\hspace{2pt}}cccc}
 & \textbf{\small AT} & \textbf{\small AT + A$^{*}$} & \textbf{\small AT + AnE} & \textbf{\small AT + AnE + A$^{*}$} \\
\rotatebox{90}{\scriptsize $t{=}10$} &
\includegraphics[width=0.235\textwidth]{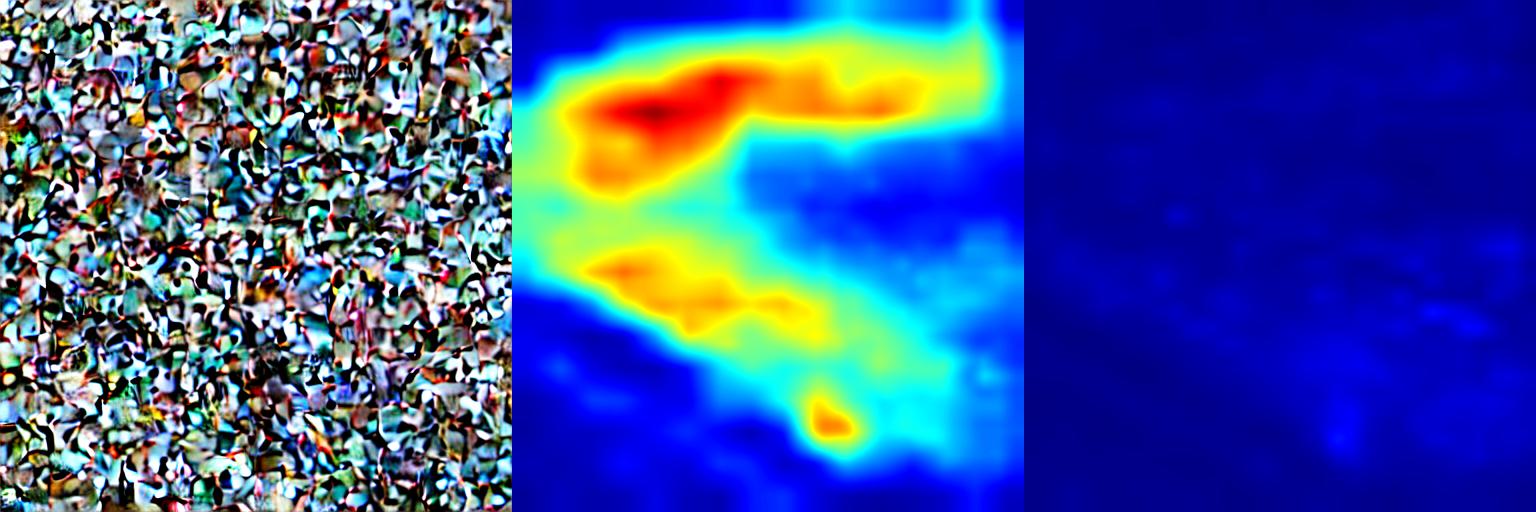} &
\includegraphics[width=0.235\textwidth]{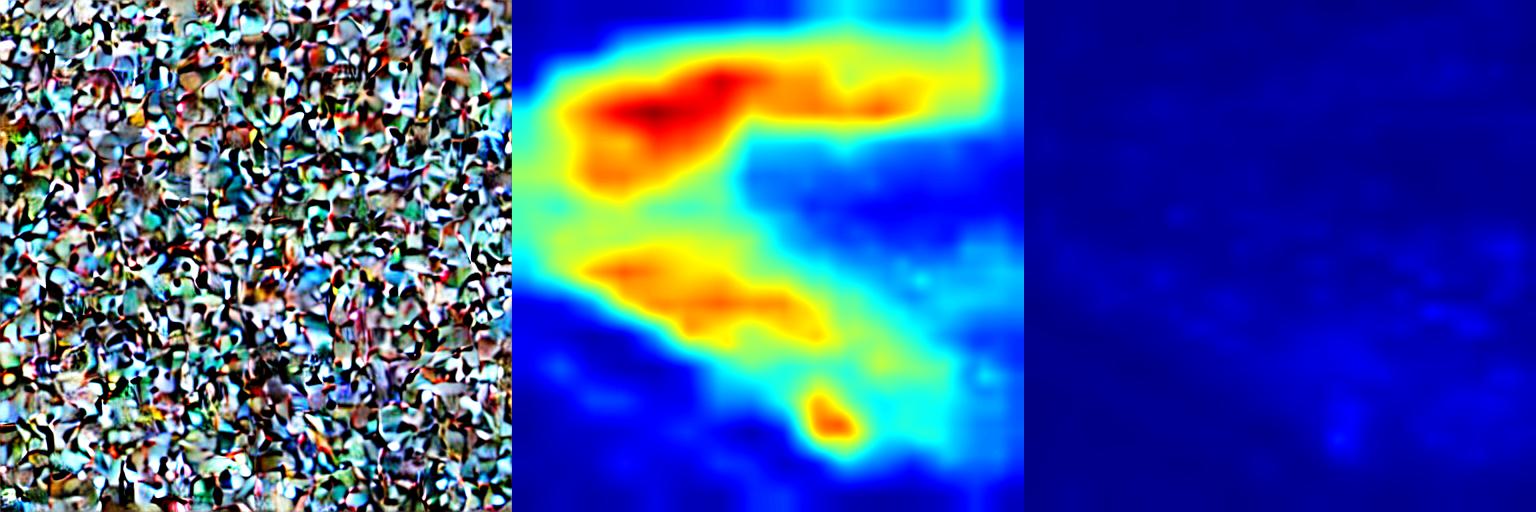} &
\includegraphics[width=0.235\textwidth]{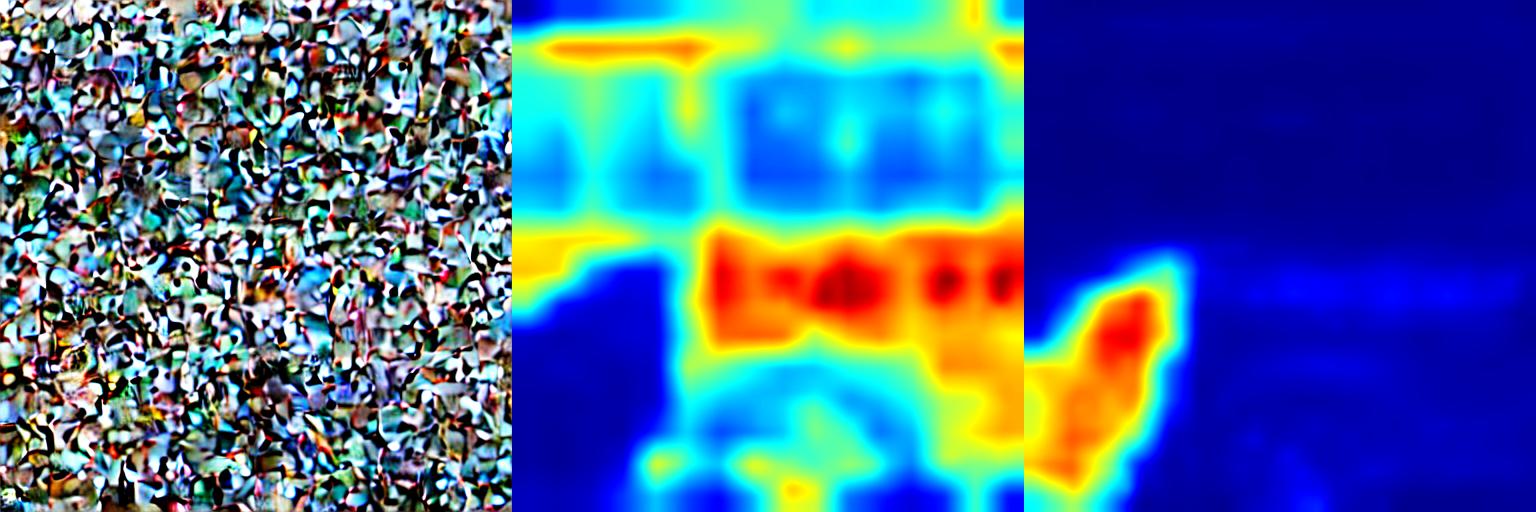} &
\includegraphics[width=0.235\textwidth]{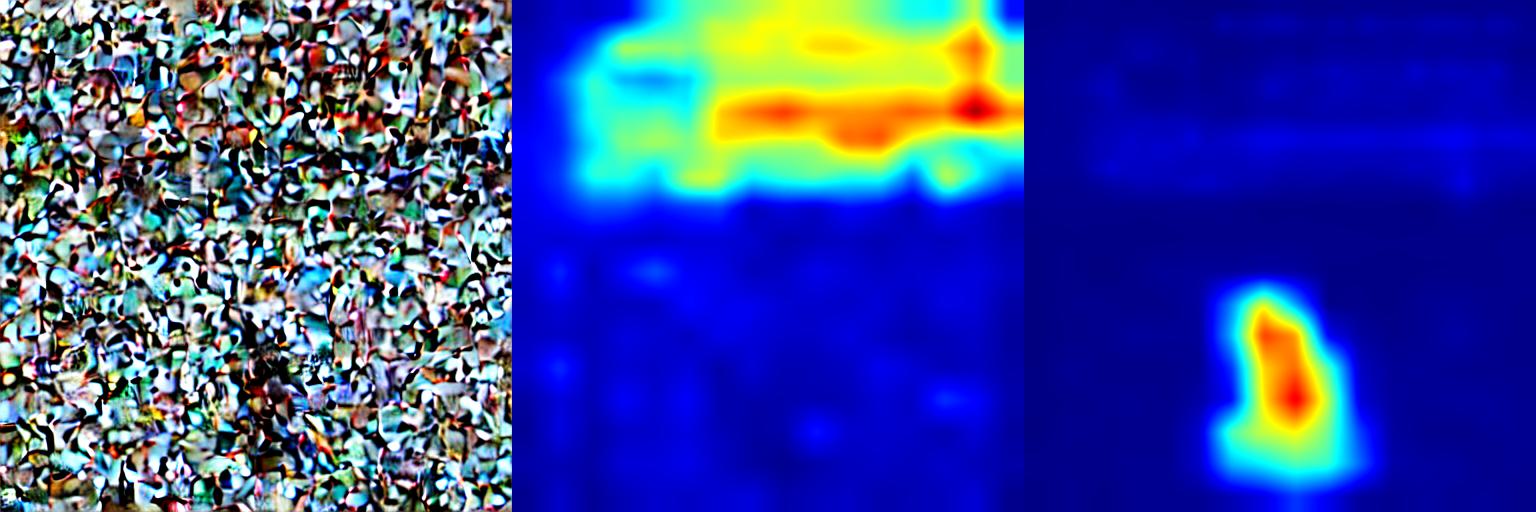} \\
\rotatebox{90}{\scriptsize $t{=}30$} &
\includegraphics[width=0.235\textwidth]{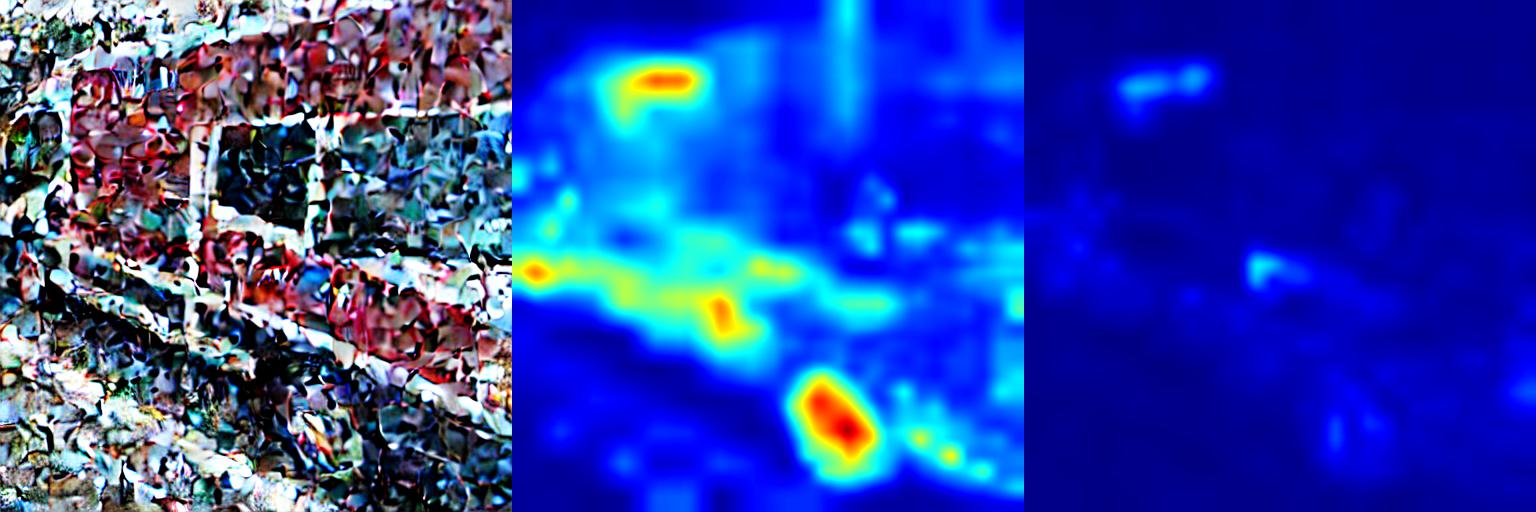} &
\includegraphics[width=0.235\textwidth]{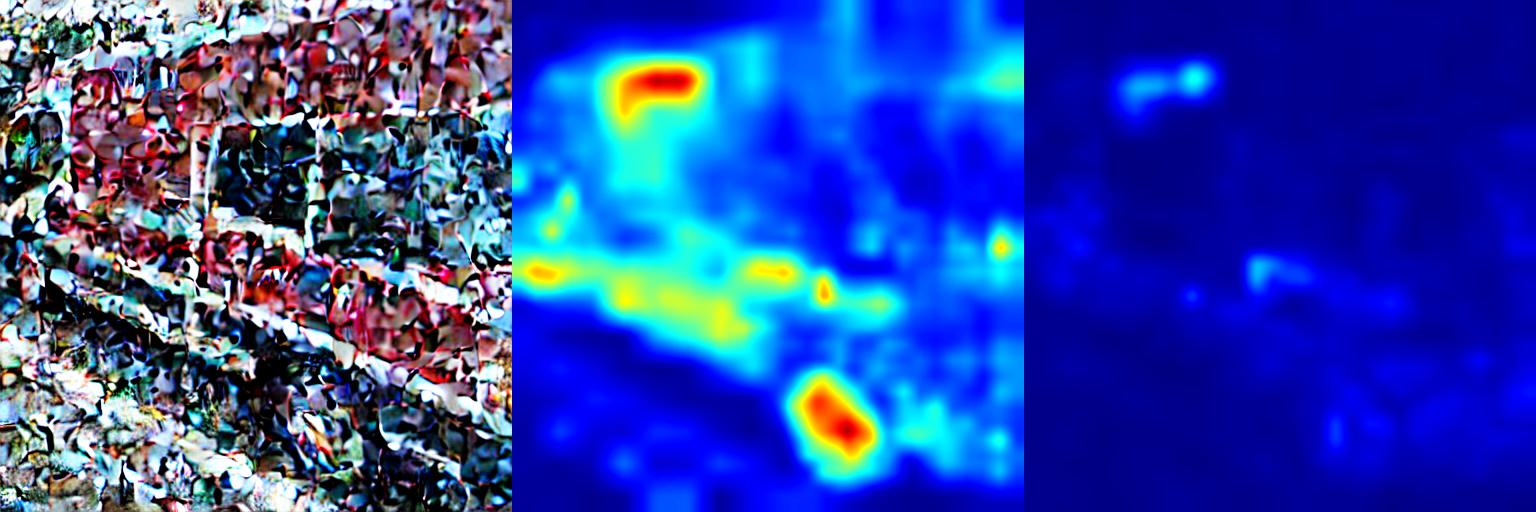} &
\includegraphics[width=0.235\textwidth]{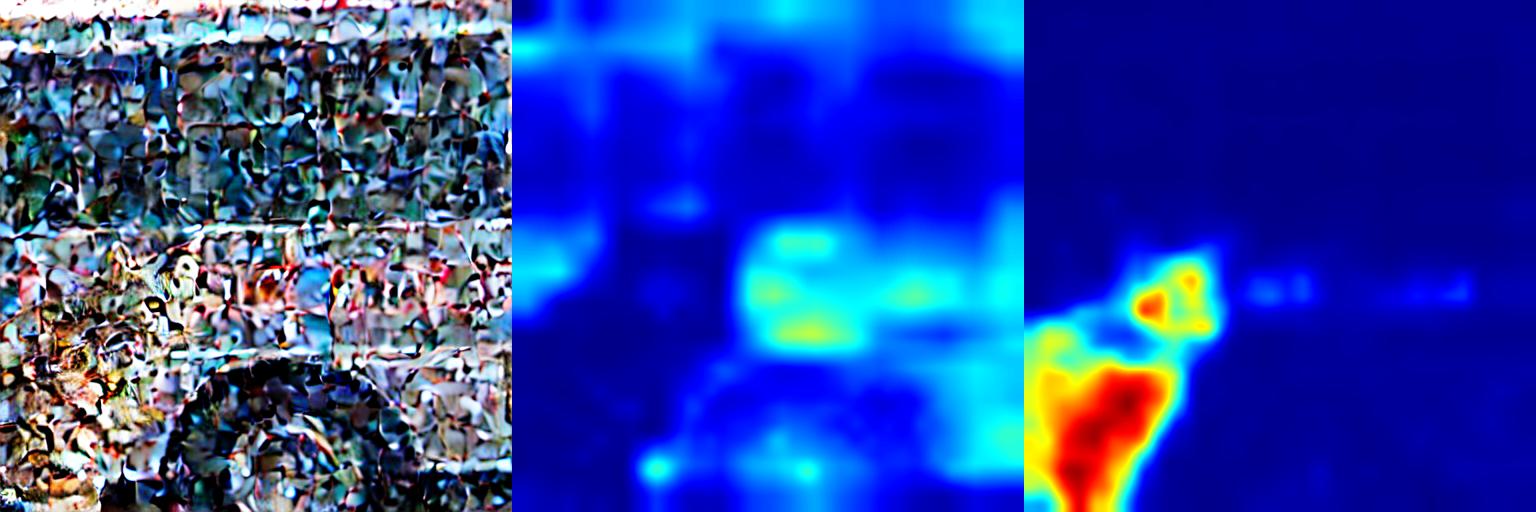} &
\includegraphics[width=0.235\textwidth]{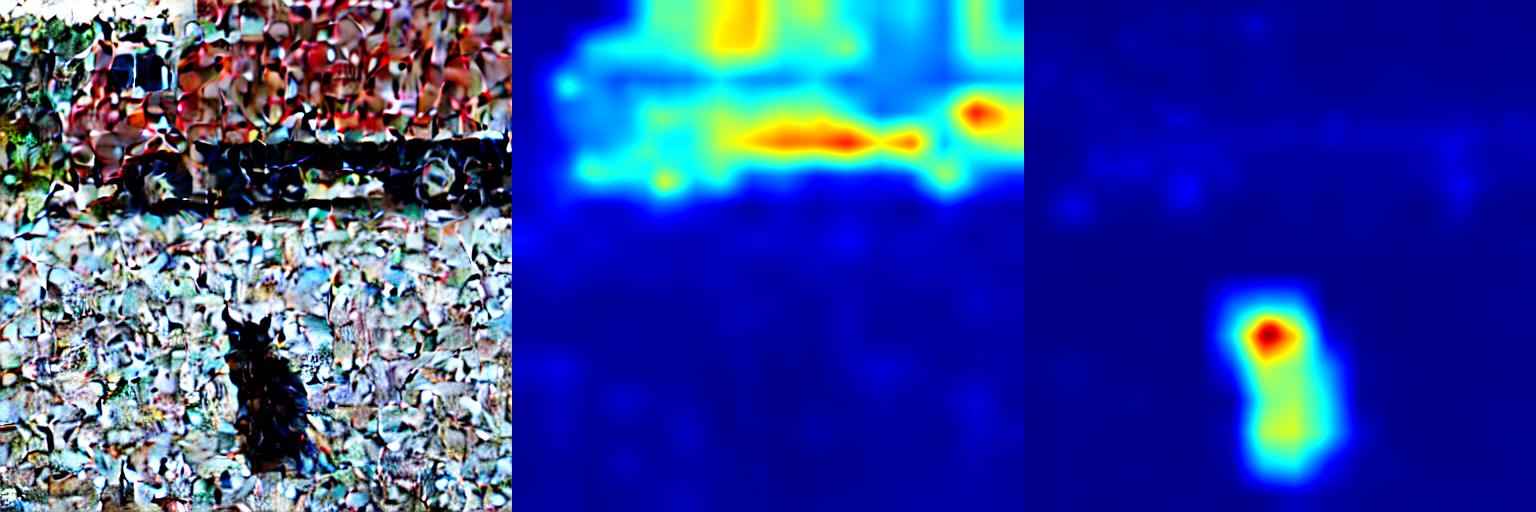} \\
\rotatebox{90}{\scriptsize $t{=}50$} &
\includegraphics[width=0.235\textwidth]{good_attention_maps/AT/2498_fire_truck_siren_cat_purring/frame_50.jpg} &
\includegraphics[width=0.235\textwidth]{good_attention_maps/Astar/2498_fire_truck_siren_cat_purring/frame_50.jpg} &
\includegraphics[width=0.235\textwidth]{good_attention_maps/AnE/2498_fire_truck_siren_cat_purring/frame_50.jpg} &
\includegraphics[width=0.235\textwidth]{good_attention_maps/mixed/2498_fire_truck_siren_cat_purring/frame_50.jpg} \\
\end{tabular}
\caption{Cross-attention maps for the two-audio input \textit{Fire Truck Siren} +
\textit{Cat Purring}. The $t{=}50$ row is reproduced as Figure~$2$ in the main paper.}
\label{fig:attn_firetruck_cat}
\end{figure*}

We visualize how the inference-time losses shape the cross-attention geometry across denoising timesteps. Each figure shows attention maps for both audio tokens at
$t\in\{10,30,50\}$ across four loss configurations: no optimization (AT), $\Lastar$ only
(AT + A$^{*}$), $\Lae$ only (AT + AnE), and the mixed loss (AT + AnE + A$^{*}$). The progression visualizes the two-stage mechanism described in Section~$3.6$, early
timesteps determine global layout, and the mixed loss yields the clearest spatial separation of audio tokens by $t{=}30$. The main paper shows only the $t{=}50$ snapshot
of Figure~\ref{fig:attn_firetruck_cat} (as Figure~$2$), and the full time-evolution is provided below. In every figure, rows are timesteps and columns are loss configurations.

\begin{figure*}[!t]
\centering
\setlength{\tabcolsep}{1.5pt}
\renewcommand{\arraystretch}{1.05}
\begin{tabular}{c@{\hspace{2pt}}cccc}
 & \textbf{\small AT} & \textbf{\small AT + A$^{*}$} & \textbf{\small AT + AnE} & \textbf{\small AT + AnE + A$^{*}$} \\
\rotatebox{90}{\scriptsize $t{=}10$} &
\includegraphics[width=0.235\textwidth]{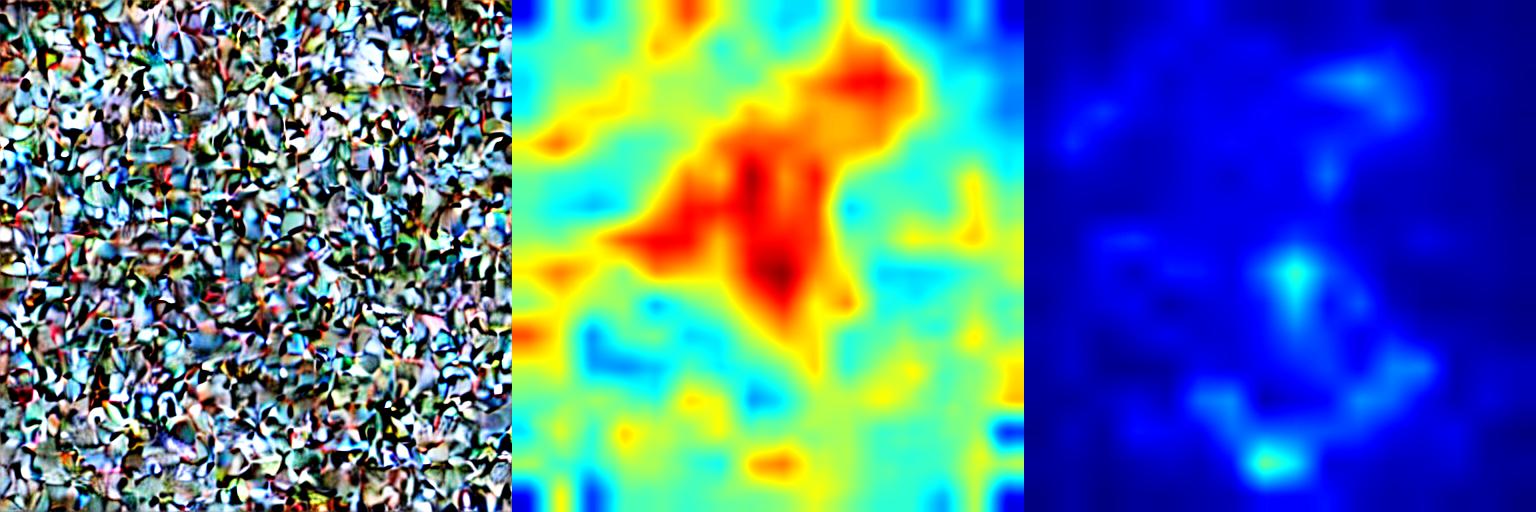} &
\includegraphics[width=0.235\textwidth]{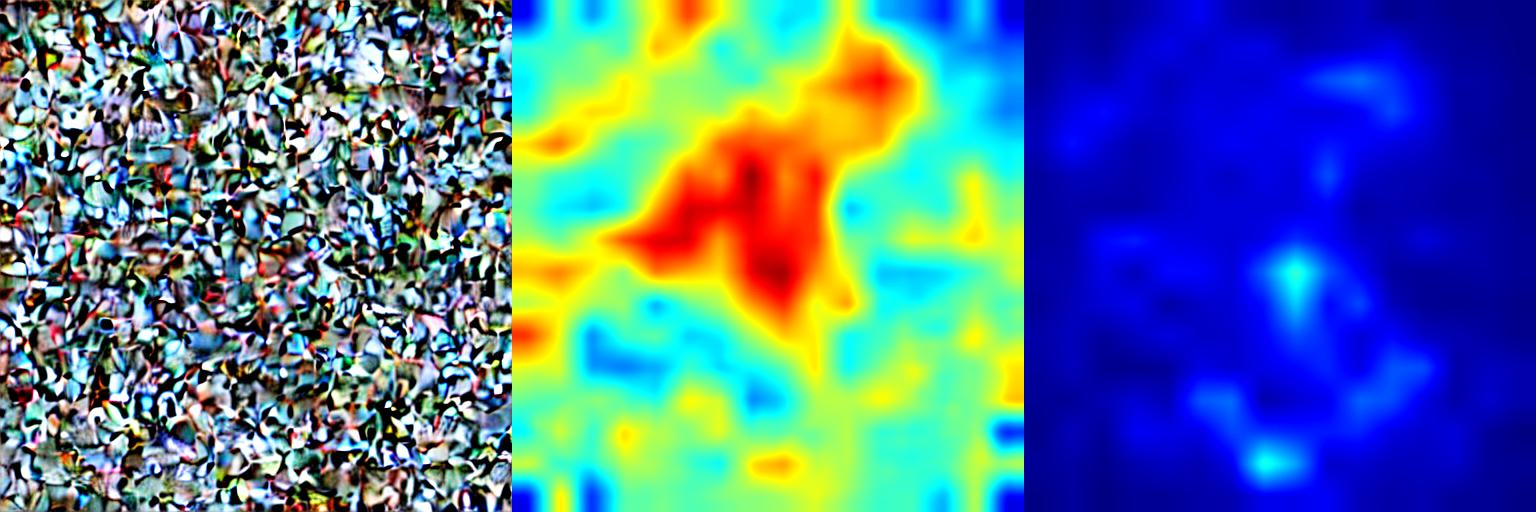} &
\includegraphics[width=0.235\textwidth]{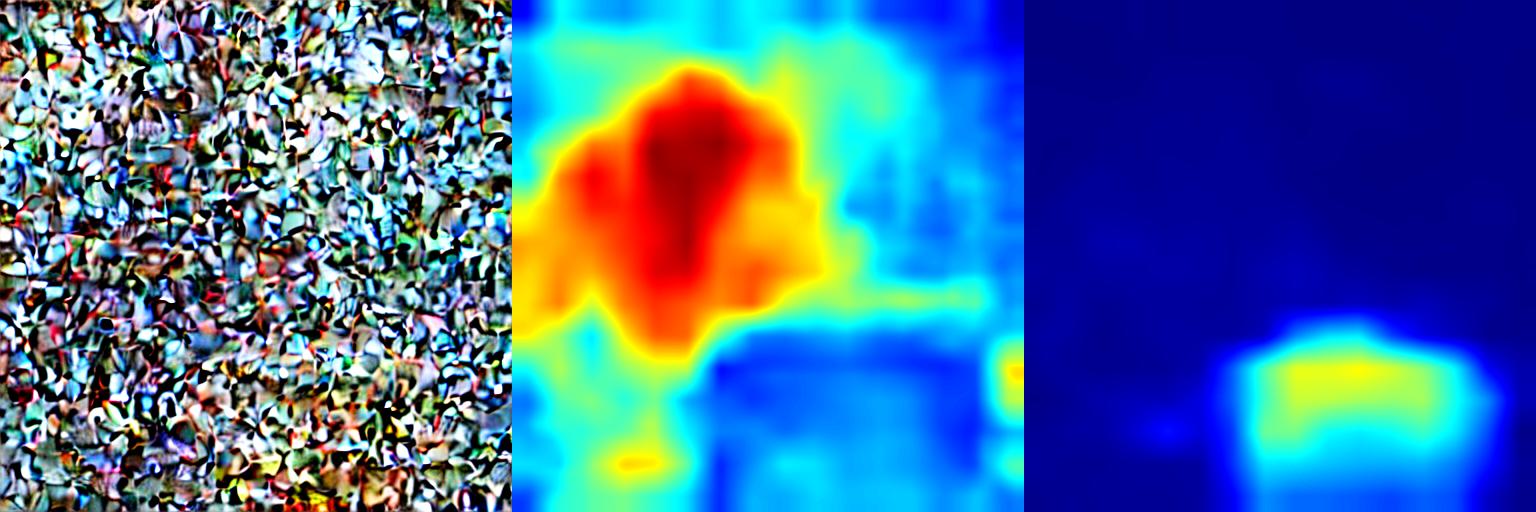} &
\includegraphics[width=0.235\textwidth]{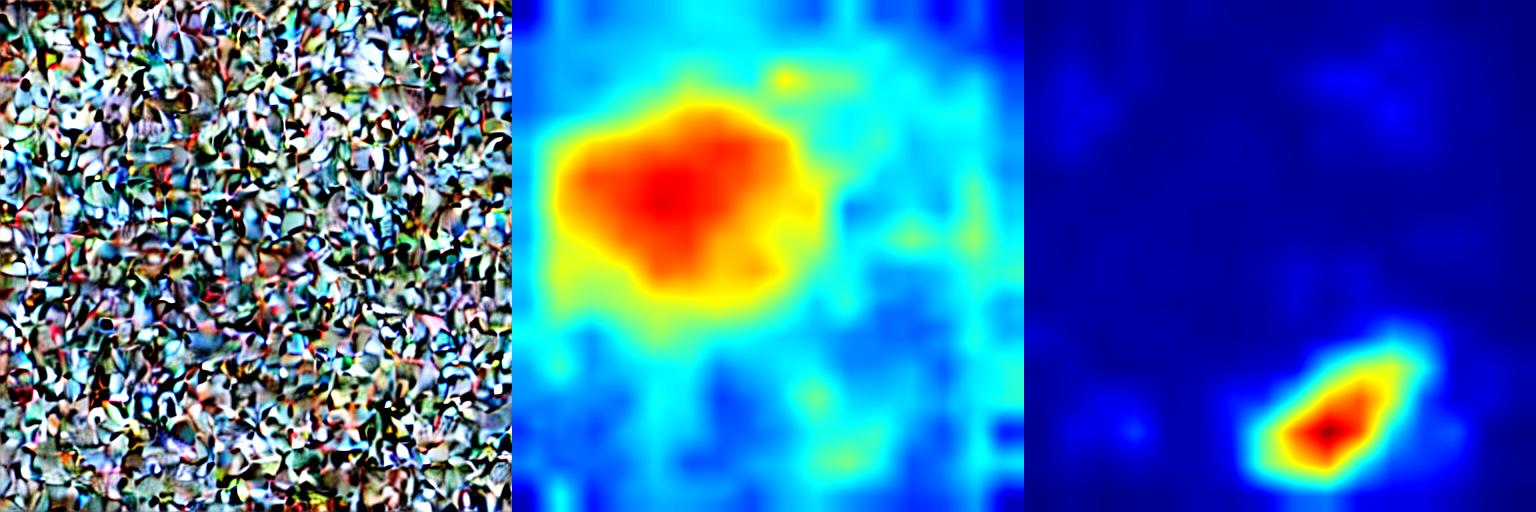} \\
\rotatebox{90}{\scriptsize $t{=}30$} &
\includegraphics[width=0.235\textwidth]{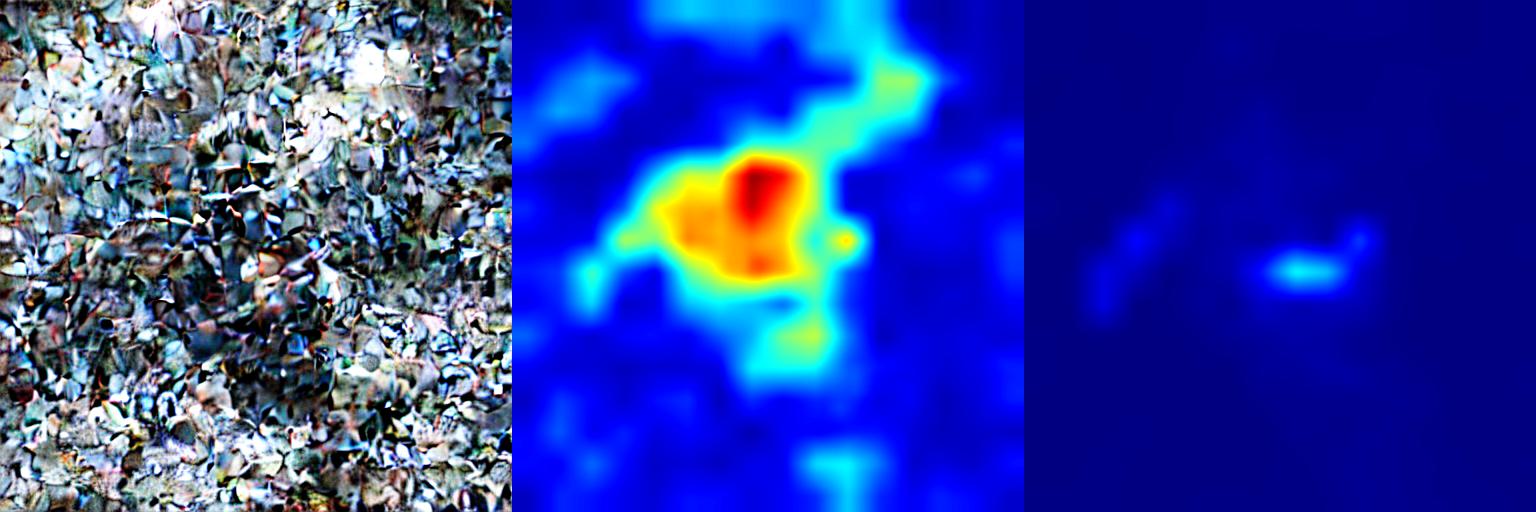} &
\includegraphics[width=0.235\textwidth]{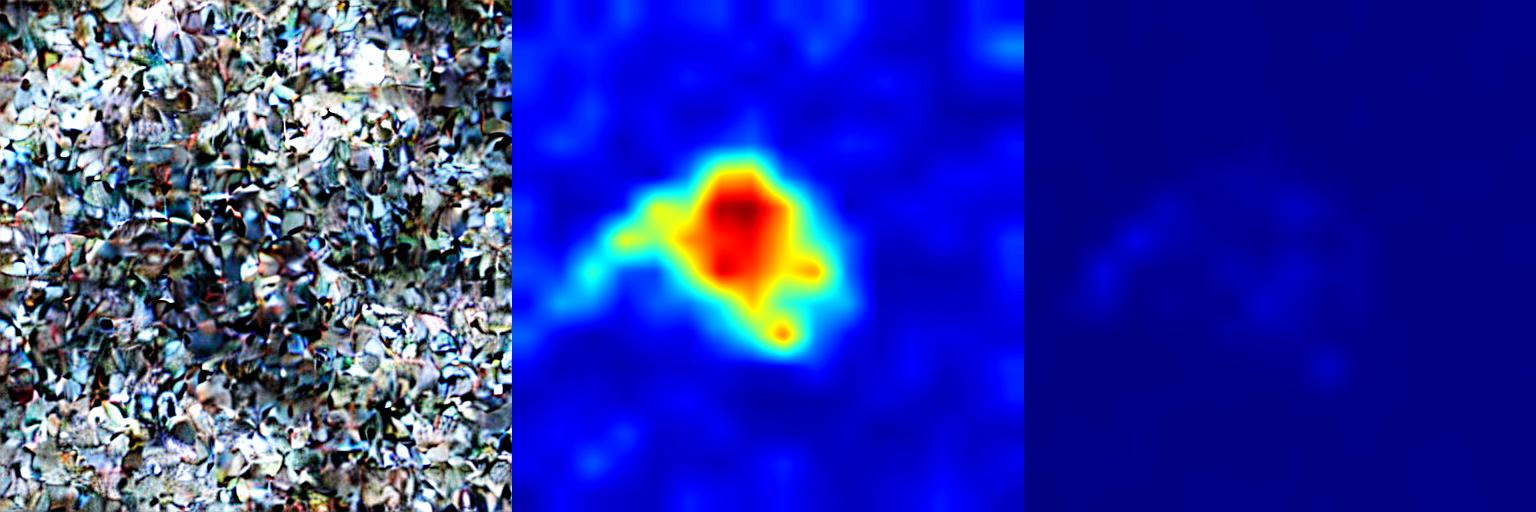} &
\includegraphics[width=0.235\textwidth]{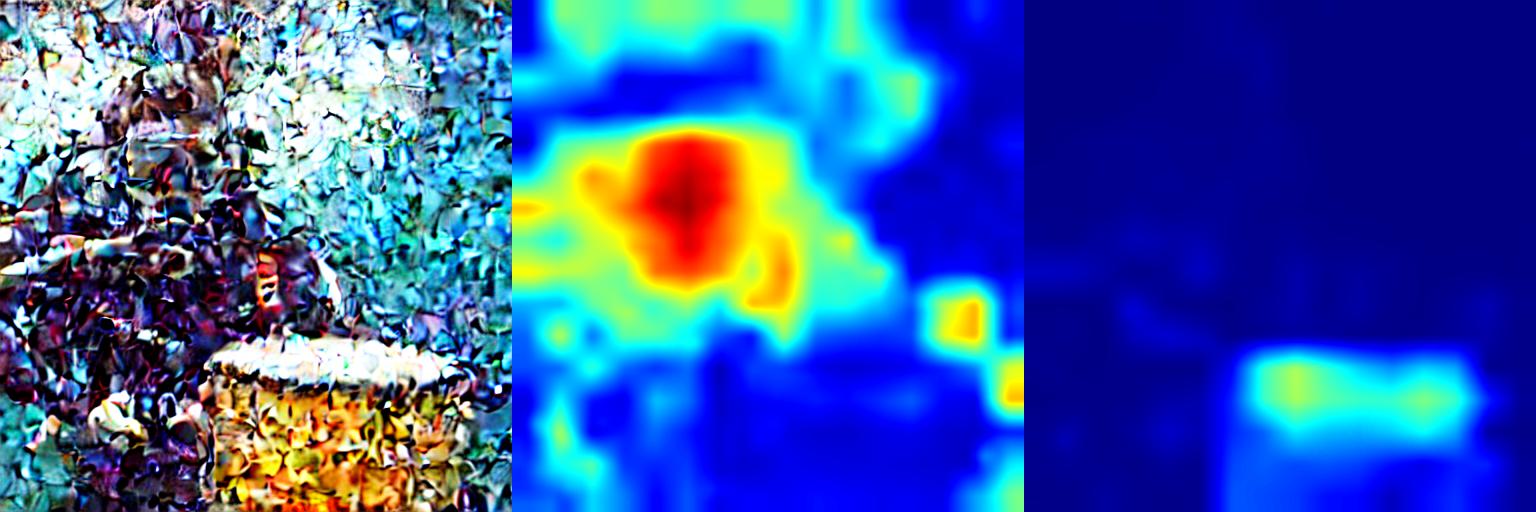} &
\includegraphics[width=0.235\textwidth]{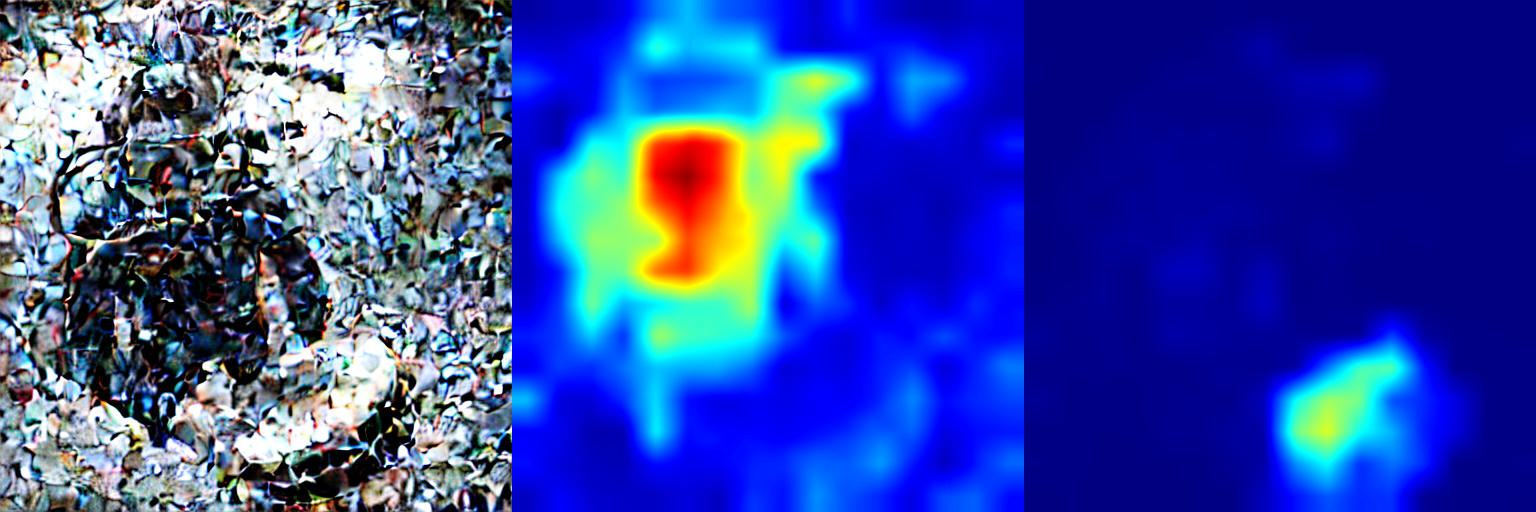} \\
\rotatebox{90}{\scriptsize $t{=}50$} &
\includegraphics[width=0.235\textwidth]{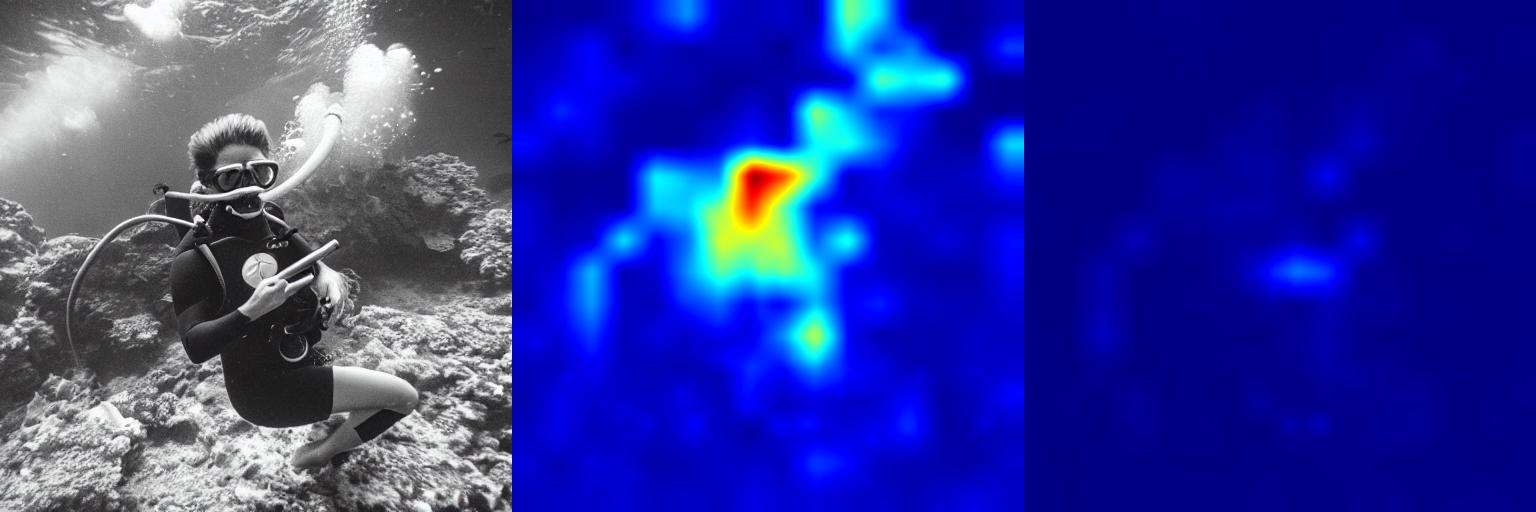} &
\includegraphics[width=0.235\textwidth]{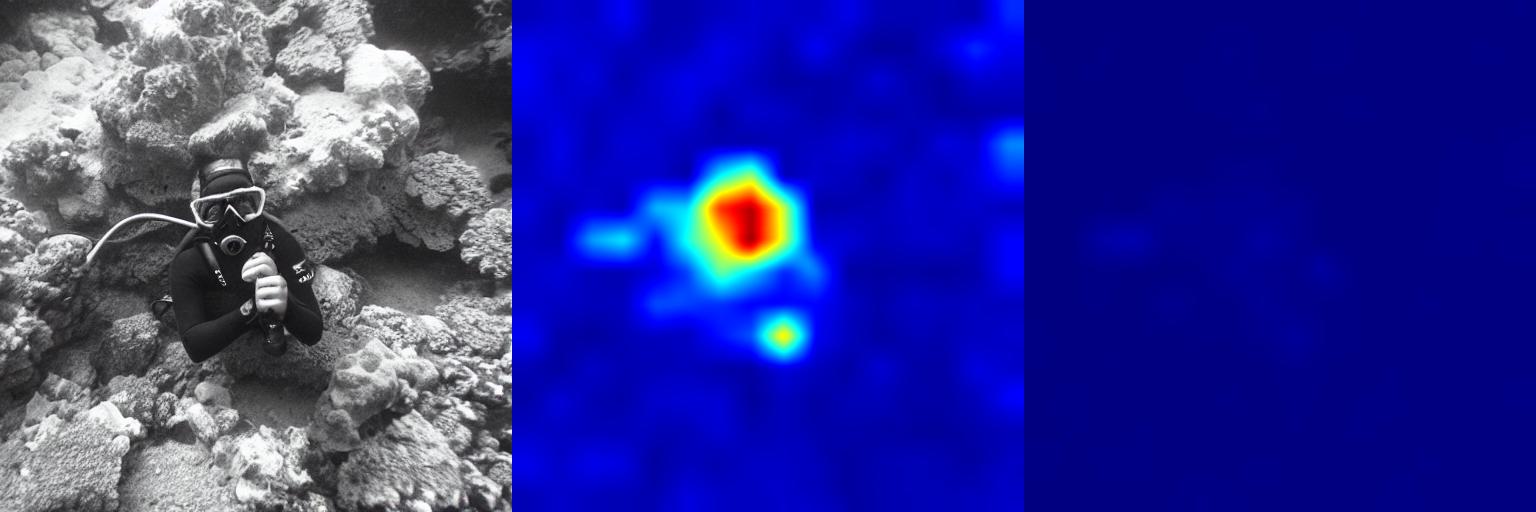} &
\includegraphics[width=0.235\textwidth]{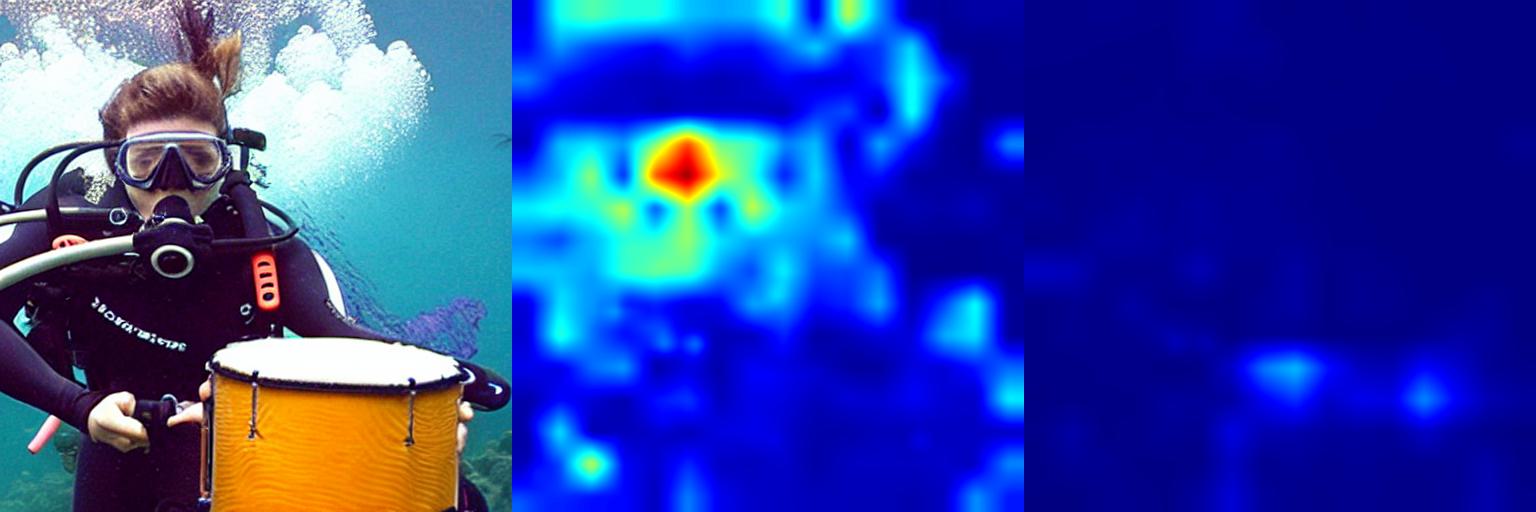} &
\includegraphics[width=0.235\textwidth]{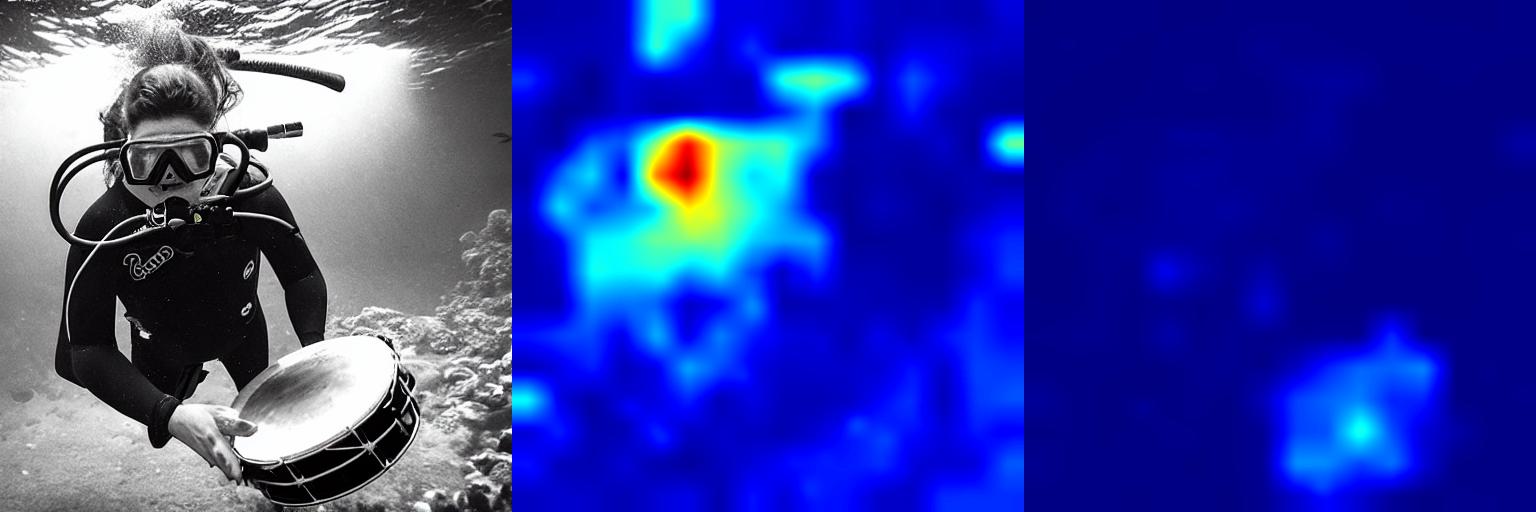} \\
\end{tabular}
\caption{Cross-attention maps for the two-audio input \textit{Scuba Diving} +
\textit{Playing Drum Kit}.}
\label{fig:attn_scuba_drum}
\end{figure*}

\begin{figure*}[!t]
\centering
\setlength{\tabcolsep}{1.5pt}
\renewcommand{\arraystretch}{1.05}
\begin{tabular}{c@{\hspace{2pt}}cccc}
 & \textbf{\small AT} & \textbf{\small AT + A$^{*}$} & \textbf{\small AT + AnE} & \textbf{\small AT + AnE + A$^{*}$} \\
\rotatebox{90}{\scriptsize $t{=}10$} &
\includegraphics[width=0.235\textwidth]{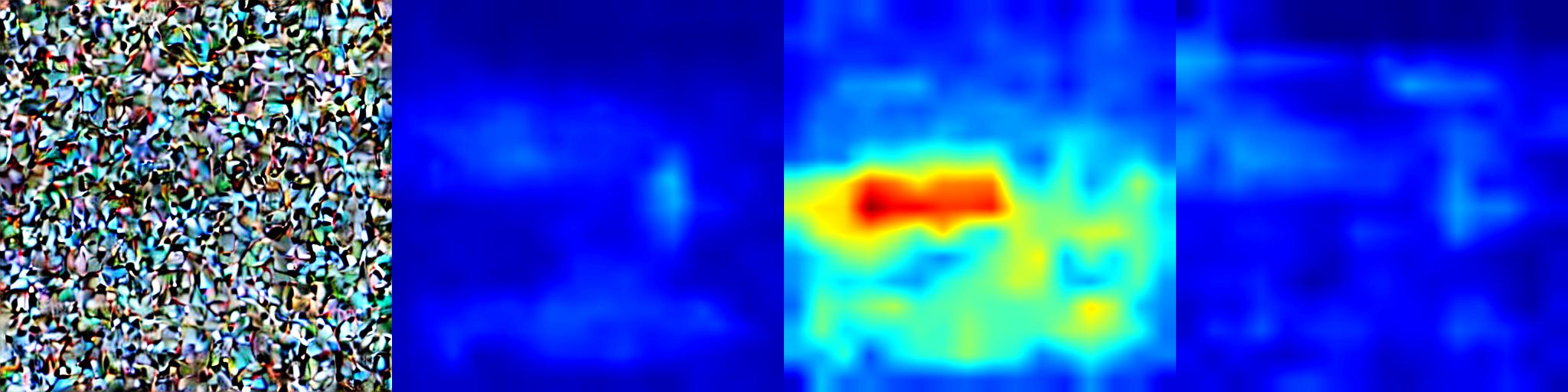} &
\includegraphics[width=0.235\textwidth]{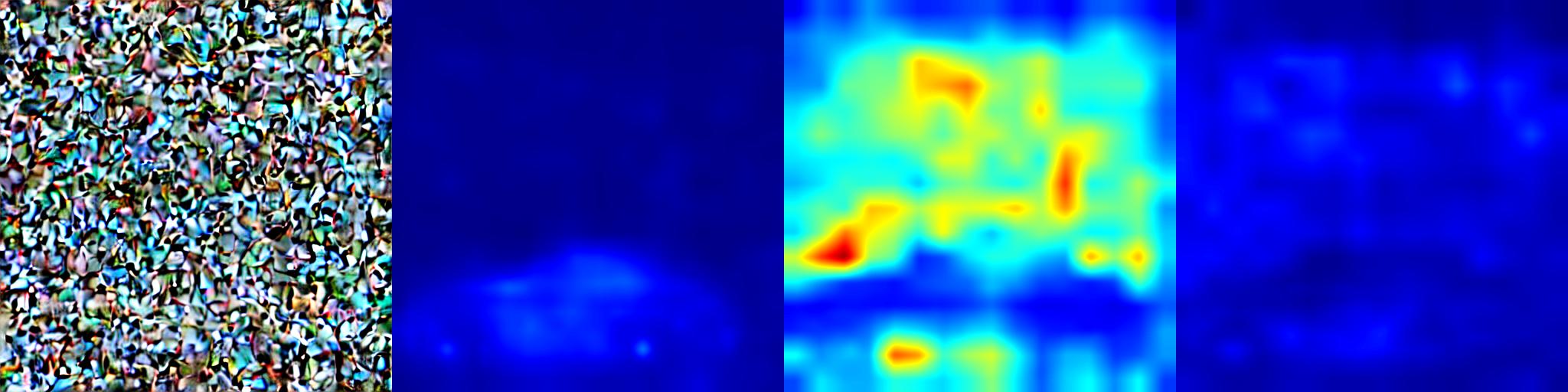} &
\includegraphics[width=0.235\textwidth]{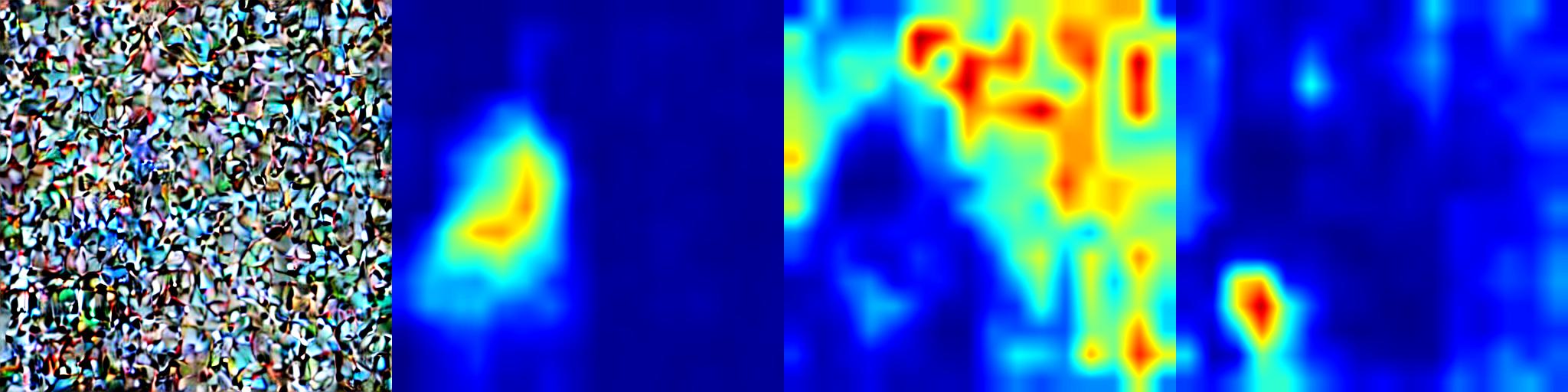} &
\includegraphics[width=0.235\textwidth]{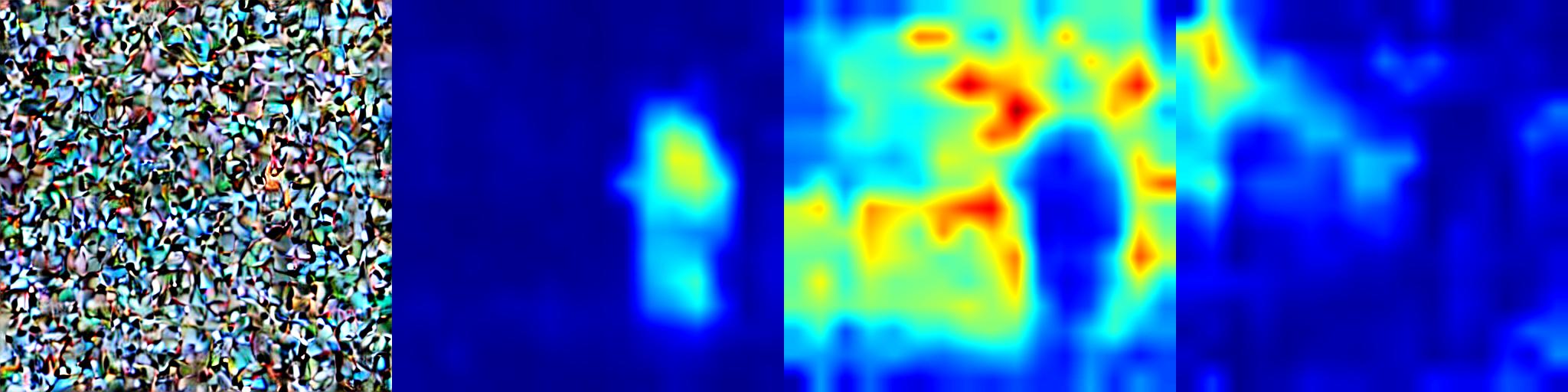} \\
\rotatebox{90}{\scriptsize $t{=}30$} &
\includegraphics[width=0.235\textwidth]{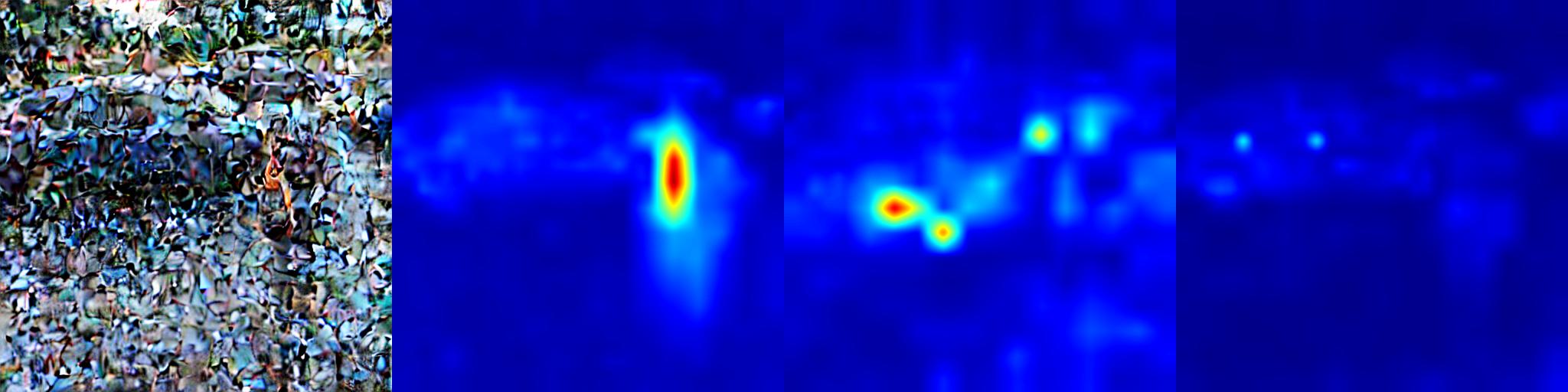} &
\includegraphics[width=0.235\textwidth]{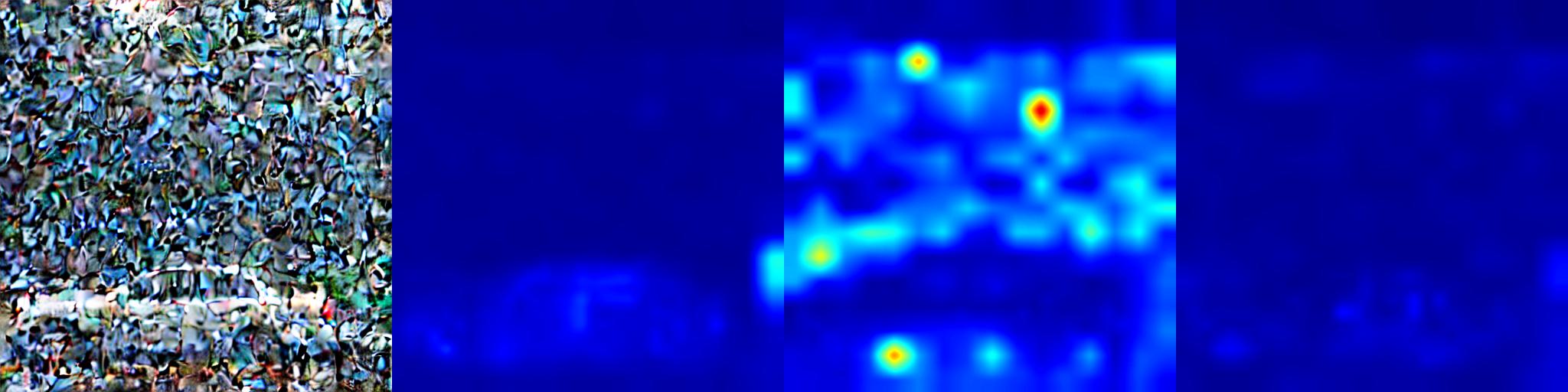} &
\includegraphics[width=0.235\textwidth]{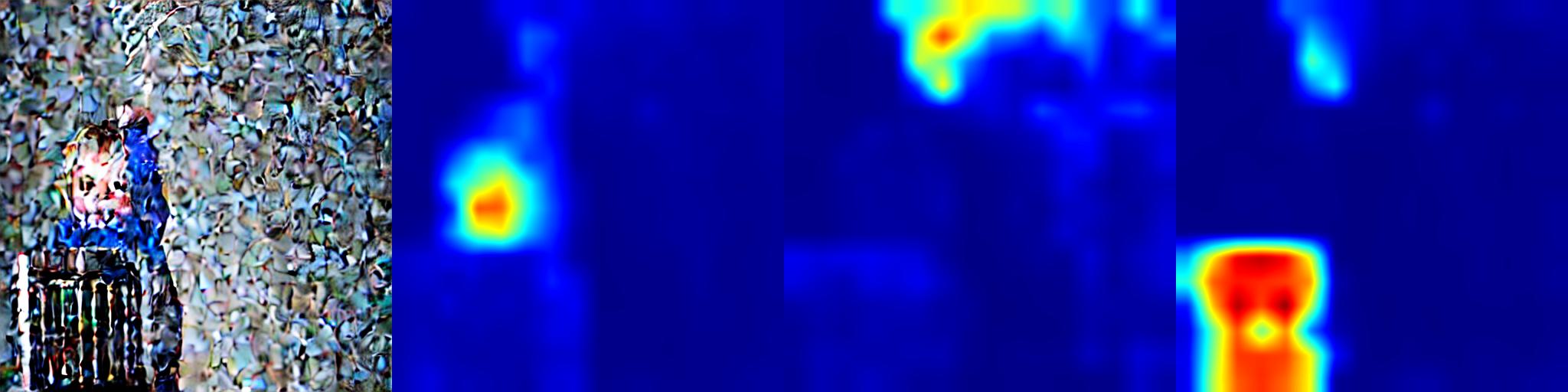} &
\includegraphics[width=0.235\textwidth]{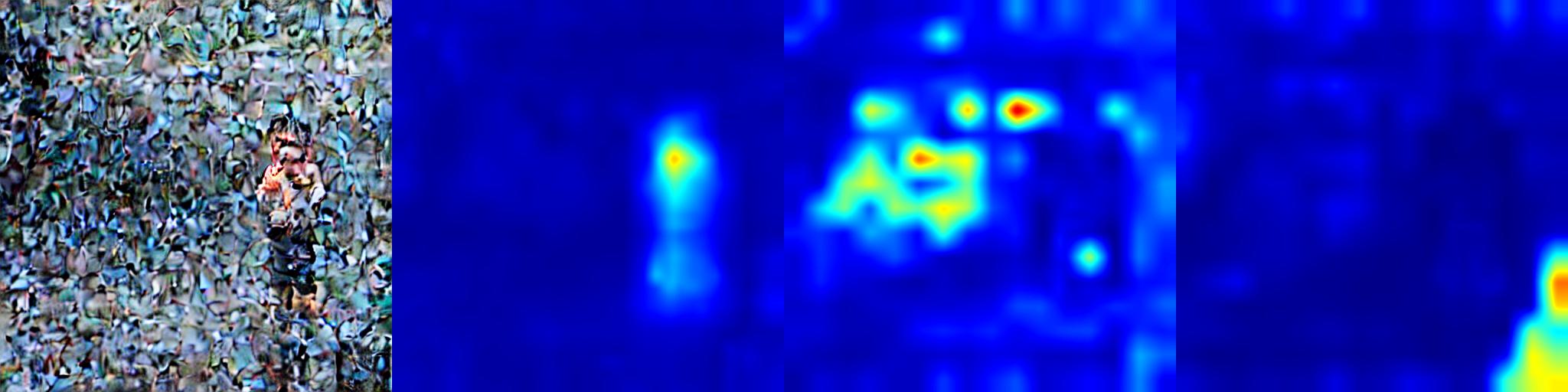} \\
\rotatebox{90}{\scriptsize $t{=}50$} &
\includegraphics[width=0.235\textwidth]{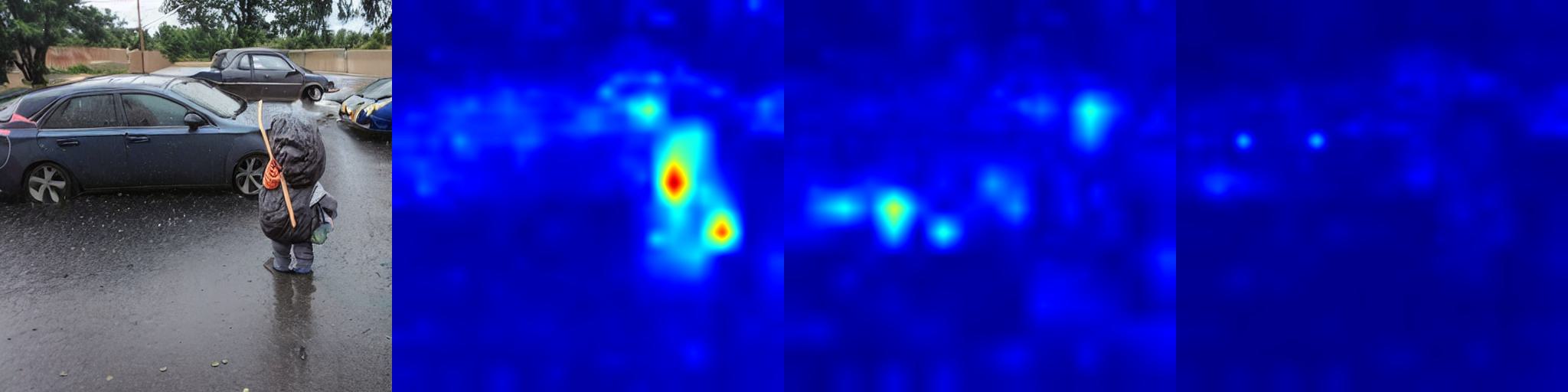} &
\includegraphics[width=0.235\textwidth]{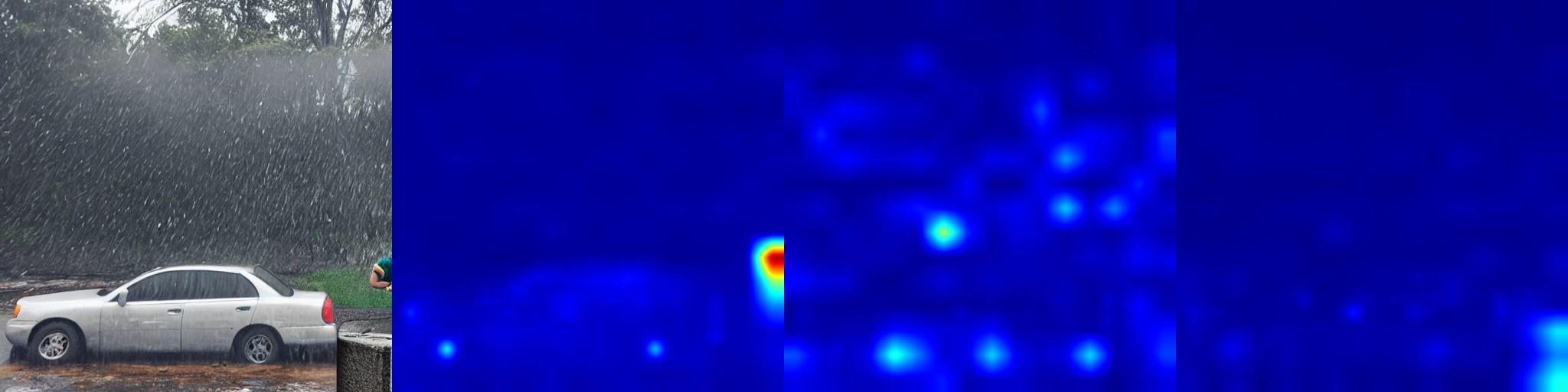} &
\includegraphics[width=0.235\textwidth]{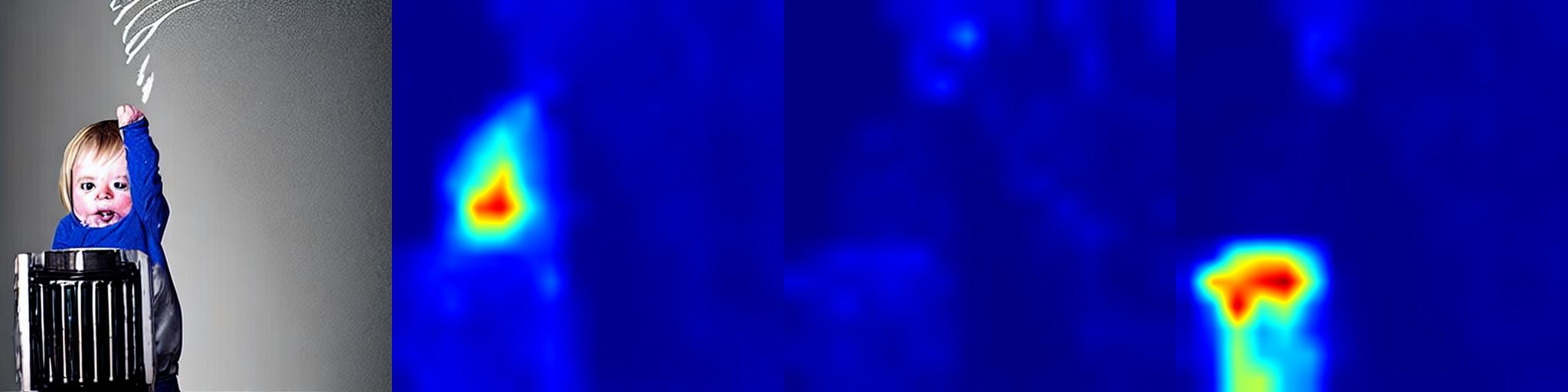} &
\includegraphics[width=0.235\textwidth]{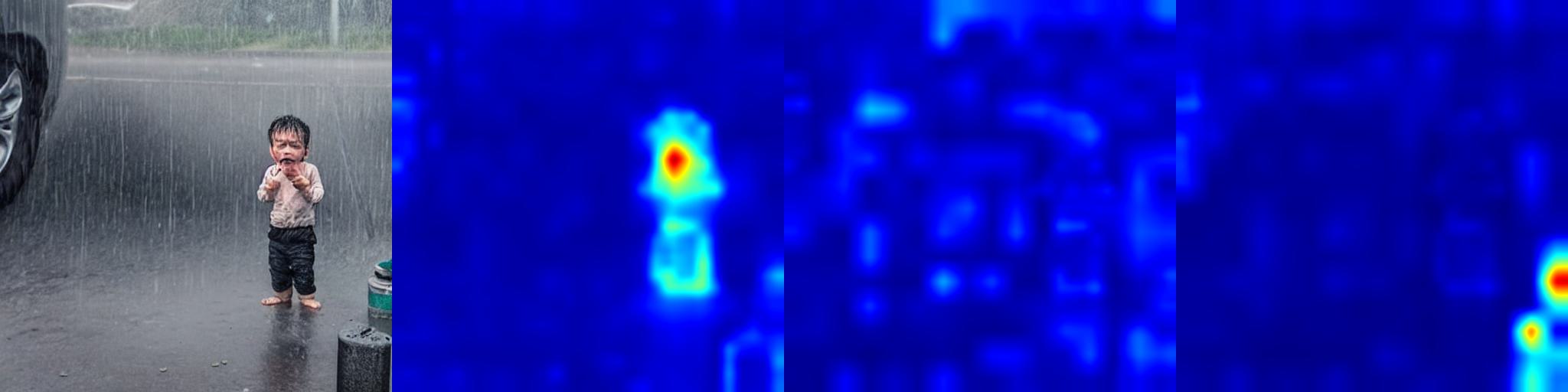} \\
\end{tabular}
\caption{Cross-attention maps for the three-audio input \textit{Baby Crying} +
\textit{Hail} + \textit{Car Engine Knocking}.}
\label{fig:attn_baby_hail_car}
\end{figure*}

\begin{figure*}[!t]
\centering
\setlength{\tabcolsep}{1.5pt}
\renewcommand{\arraystretch}{1.05}
\begin{tabular}{c@{\hspace{2pt}}cccc}
 & \textbf{\small AT} & \textbf{\small AT + A$^{*}$} & \textbf{\small AT + AnE} & \textbf{\small AT + AnE + A$^{*}$} \\
\rotatebox{90}{\scriptsize $t{=}10$} &
\includegraphics[width=0.235\textwidth]{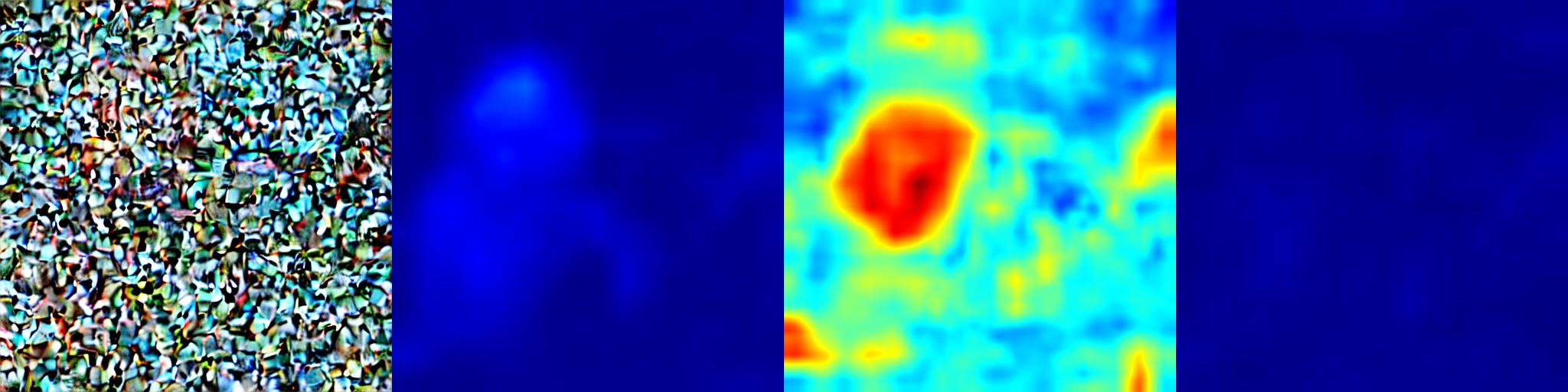} &
\includegraphics[width=0.235\textwidth]{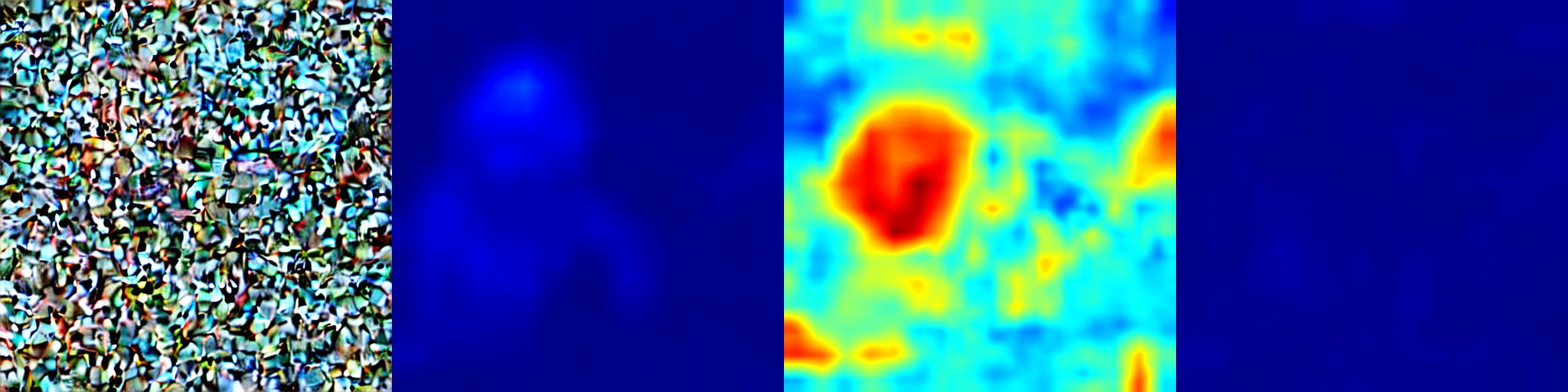} &
\includegraphics[width=0.235\textwidth]{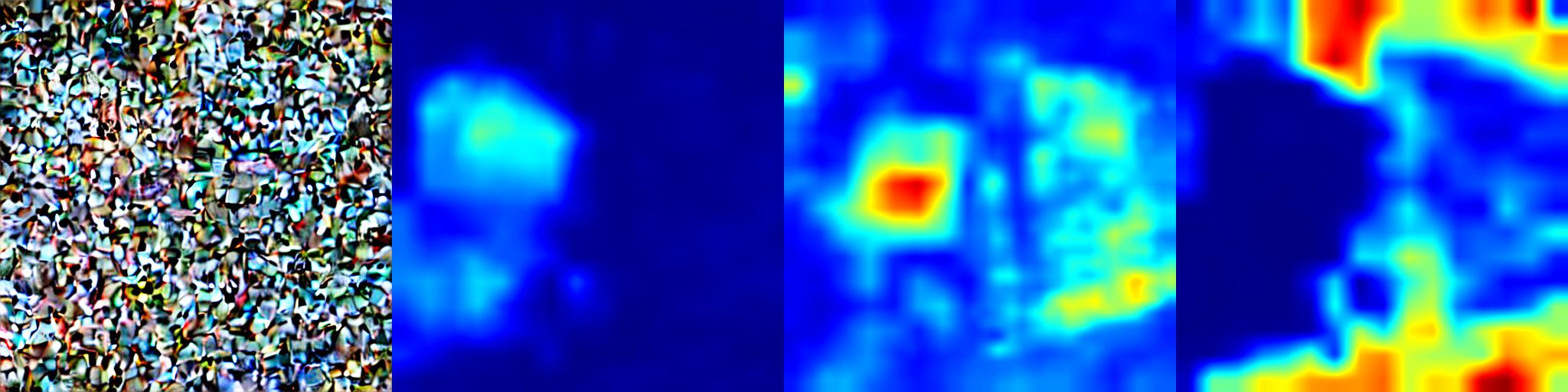} &
\includegraphics[width=0.235\textwidth]{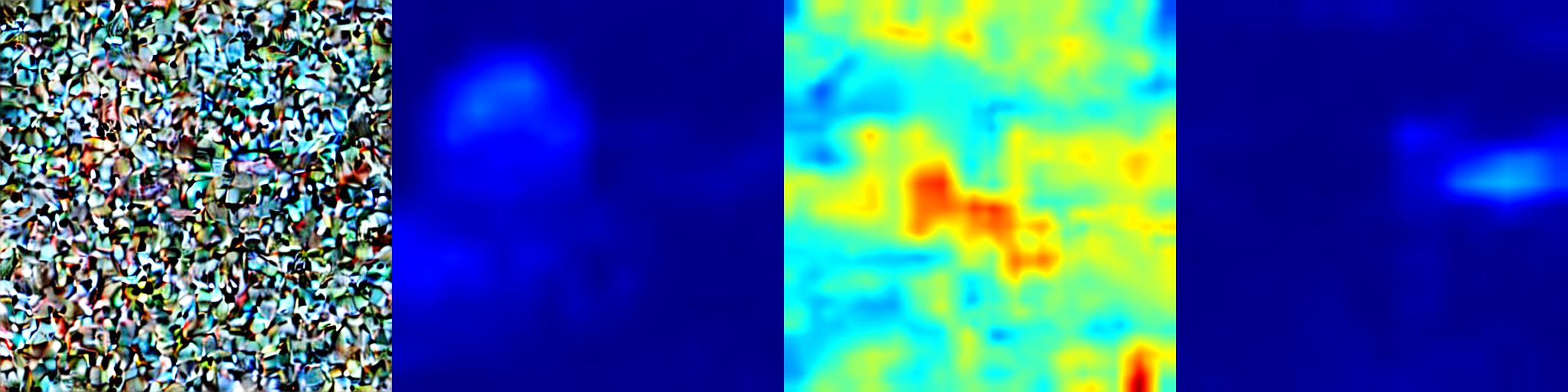} \\
\rotatebox{90}{\scriptsize $t{=}30$} &
\includegraphics[width=0.235\textwidth]{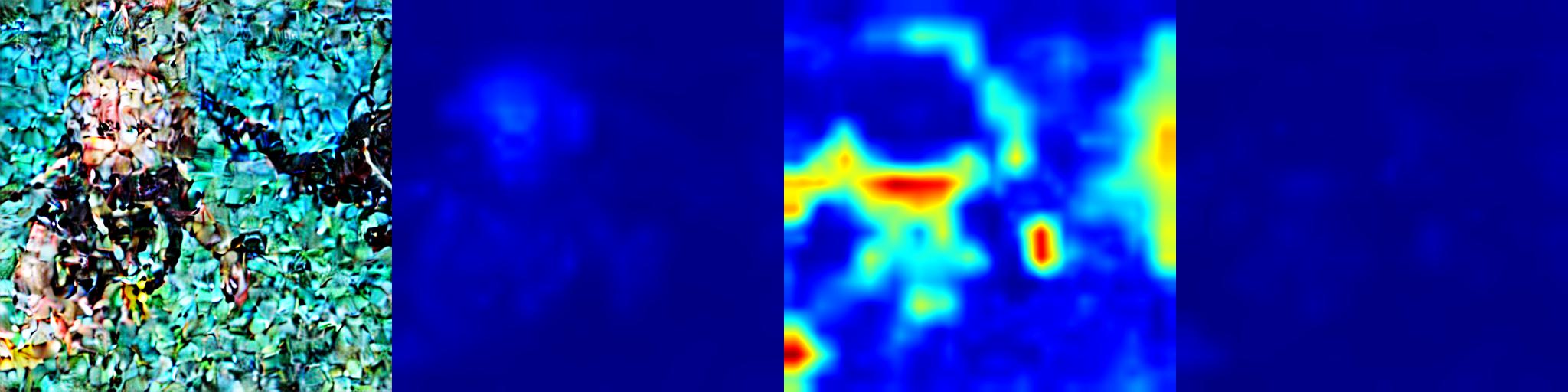} &
\includegraphics[width=0.235\textwidth]{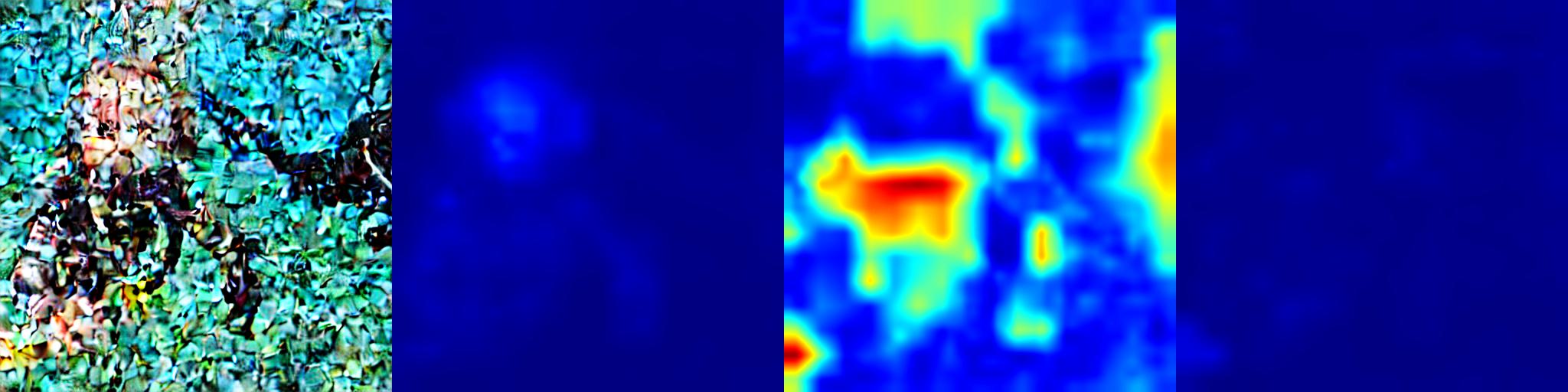} &
\includegraphics[width=0.235\textwidth]{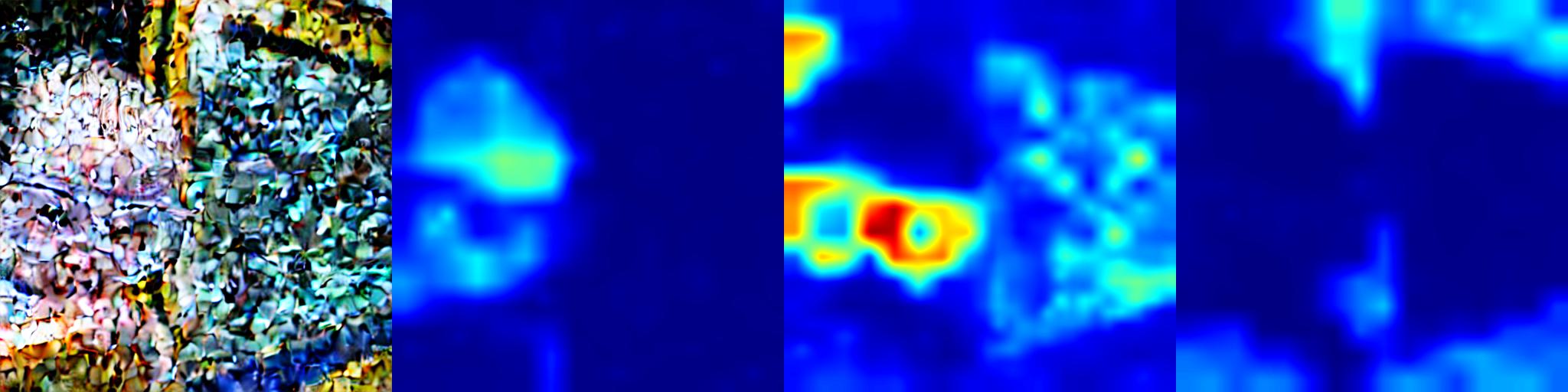} &
\includegraphics[width=0.235\textwidth]{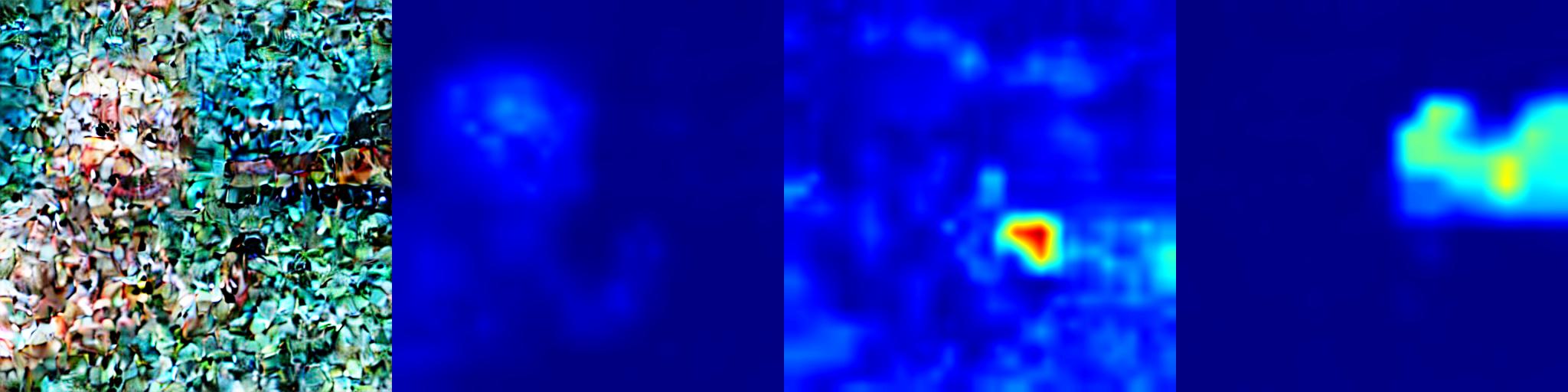} \\
\rotatebox{90}{\scriptsize $t{=}50$} &
\includegraphics[width=0.235\textwidth]{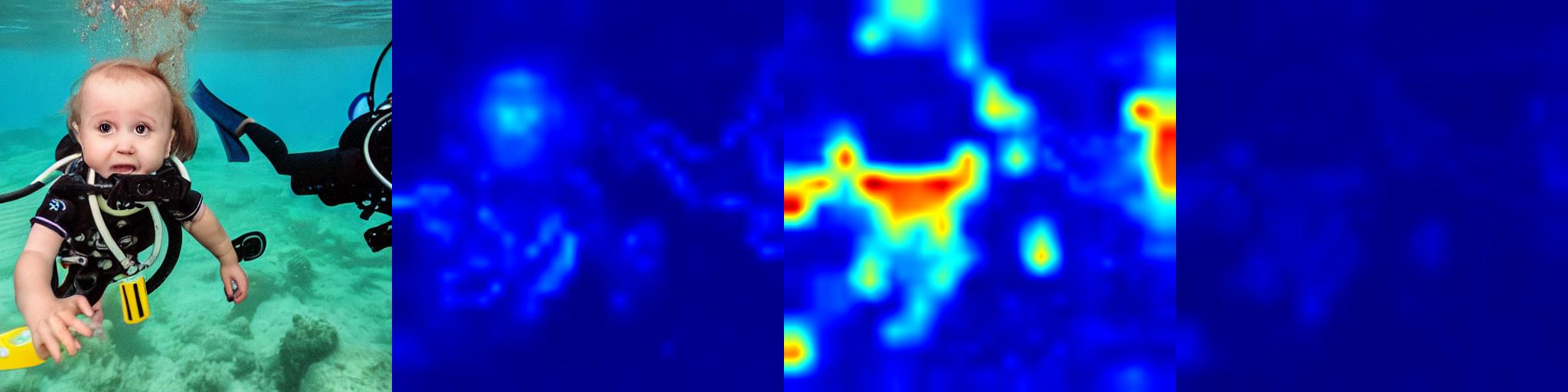} &
\includegraphics[width=0.235\textwidth]{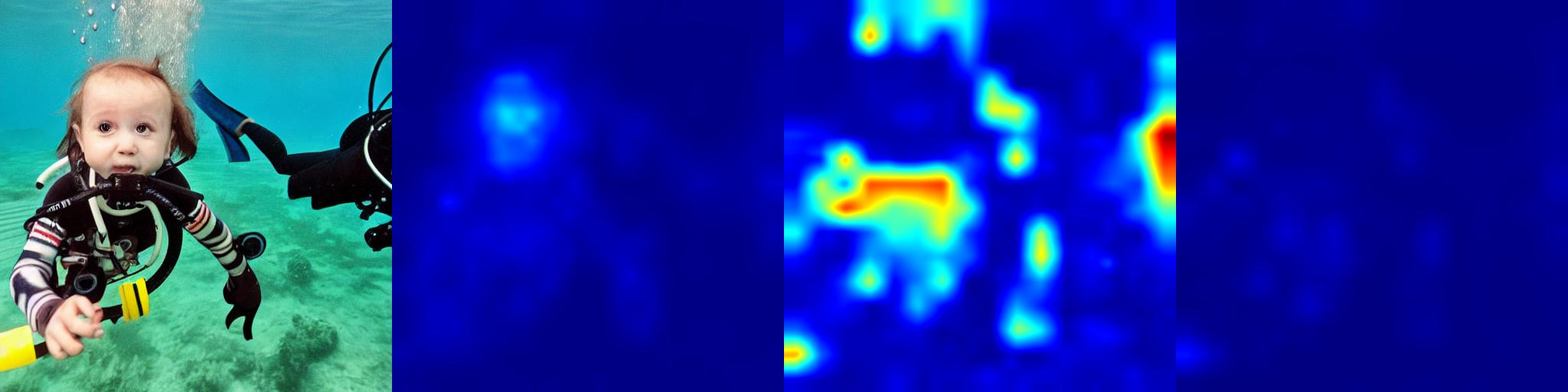} &
\includegraphics[width=0.235\textwidth]{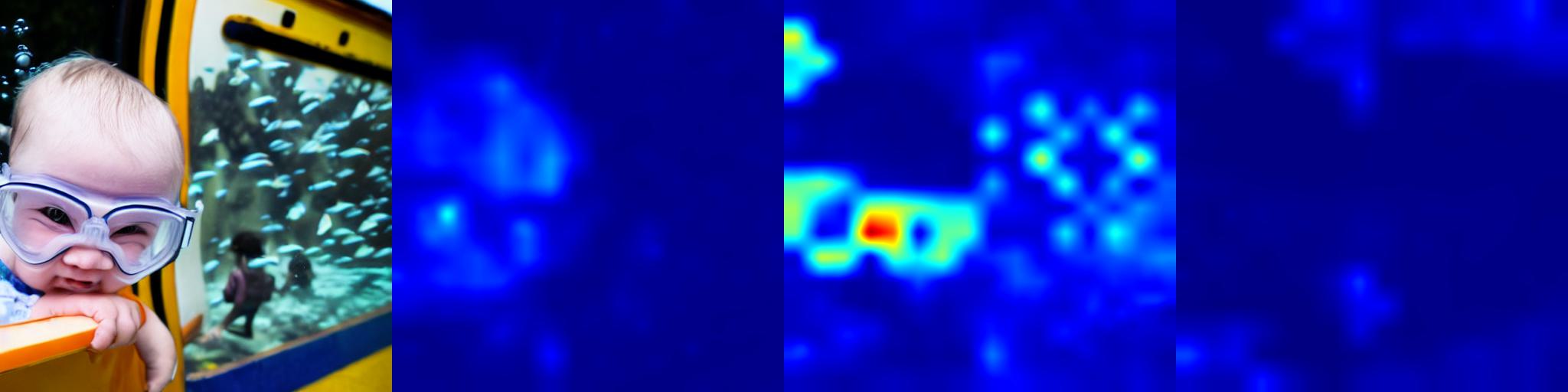} &
\includegraphics[width=0.235\textwidth]{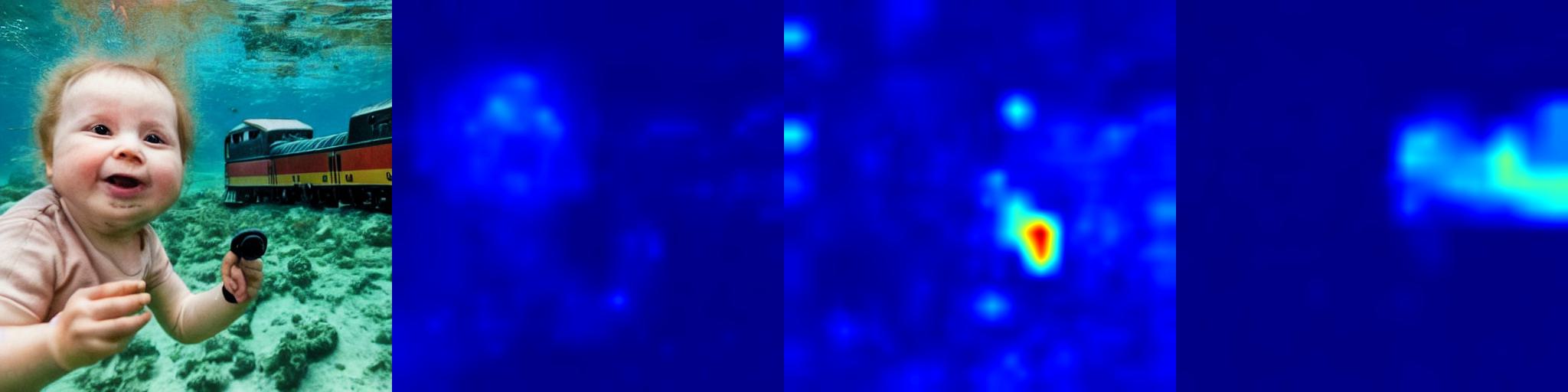} \\
\end{tabular}
\caption{Cross-attention maps for the three-audio input \textit{Baby Laughter} +
\textit{Scuba Diving} + \textit{Train Horning}.}
\label{fig:attn_baby_scuba_train}
\vspace{-1.0em}
\end{figure*}

\section{UMAP Visualizations of the Audio-Token Geometry}
\label{supp:umap}

We provide the remaining UMAP projections of audio tokens in CLIP space (after passing through the trained embedder) that support the geometric claims of Section~$3.3$; the \textit{Elk Bugling} panel appears in the main paper as
Figure~$4$. In each panel, coloured points are tokens from the
highlighted class; grey points are tokens from all other classes; circles are BEATs
tokens and triangles are CLAP tokens; gold and cyan stars are the corresponding class
centroids; the red diamond is the CLIP text embedding of the class label, and dashed
lines connect centroids to the text embedding. Two phenomena are visible across all
classes: (i) CLAP token clusters are consistently tighter and better separated;
(ii) the cyan (CLAP) centroid is farther from the text embedding than the gold (BEATs)
centroid in absolute terms, but pointing in a similar direction, exactly the
L2-vs-cosine pattern reported in Table~$1$.

\section{Inference-Time Cost Analysis}
\label{supp:runtime}

The mixed objective repairs cross-attention geometry during denoising rather than
at training time, so it introduces a wall-clock cost at every generation. We
report that cost as the time needed to produce one image, and express it as an
overhead relative to sampling without inference-time optimization. We separate
how this overhead scales with the number of audio sources from how it decomposes
across the individual loss terms.

\vspace{1mm}\noindent
\textbf{Measurement protocol.} We time a single call to the denoising pipeline,
synchronising the device immediately before and after so that no asynchronous
kernel launches are attributed to the wrong interval. The measured interval
therefore covers the complete $50$-step generation of one image, including the
attention-map collection and the latent updates, and excludes audio encoding and
the embedder forward pass. All runs use the same training conditions as mentioned earlier.

\begin{table}[!t]
\centering
\small
\setlength{\tabcolsep}{6pt}
\renewcommand{\arraystretch}{1.05}
\begin{tabular}{lccc}
\toprule
\textbf{Method} & \textbf{VGG2} ($n{=}2$) & \textbf{VGG3} ($n{=}3$) & \textbf{VGGn} ($n\!\in\!\{2,\dots,5\}$) \\
\midrule
AT (Clap), no inference loss & $4.082$ & $4.030$ & $4.032$ \\
\rowcolor{ourrow}
AT + $\Lmix$ (Clap), ours    & $11.408$ & $12.027$ & $12.356$ \\
\midrule
Relative overhead            & $\times 2.79$ & $\times 2.98$ & $\times 3.06$ \\
\bottomrule
\end{tabular}
\vspace{0.4em}
\caption{\textbf{Per-image inference cost as the number of audio sources grows.}
Wall-clock seconds per generated image, averaged over all groups of each
benchmark. The last row reports the ratio of the mixed objective to sampling
without inference-time optimization on the same benchmark.}
\label{tab:runtime_scaling}
\end{table}

\begin{table}[!t]
\centering
\small
\setlength{\tabcolsep}{6pt}
\renewcommand{\arraystretch}{1.05}
\begin{tabular}{lcc}
\toprule
\textbf{Configuration} & \textbf{Time / image (s)} & \textbf{Relative} \\
\midrule
AT (Clap)                           & $4.032$  & $\times 1.00$ \\
AT + $\Lastar$ (Clap)               & $7.272$  & $\times 1.80$ \\
\rowcolor{ourrow}
AT + $\Lmix$ (Clap),         & $12.356$ & $\times 3.06$ \\
\bottomrule
\end{tabular}
\vspace{0.4em}
\caption{\textbf{Cost decomposition by inference-time loss component}, measured on
VGGn under the same protocol as Table~\ref{tab:runtime_scaling}. Each row
corresponds to the matching row of the CLAP block of the ablation
(Table~\ref{tab:ablation_vggn}), so the accuracy of a configuration can be read
directly against its cost.}
\label{tab:runtime_losses}
\end{table}

\vspace{1mm}\noindent
\textbf{Scaling with the number of sources.} Table~\ref{tab:runtime_scaling}
reports the per-image cost of unoptimized sampling and of the full mixed
objective on the three compositional benchmarks. Unoptimized sampling stays flat
as $n$ grow while the mixed objective is close to three times slower, and that ratio barely moves between VGG2 and VGGn. Most of the added time goes into the backward pass used to
update the latent, and the cost of that pass is set by the size of the U-Net
rather than by how many tokens are being optimized. The overhead therefore
behaves like a constant multiplier, so the method stays practical as more sources
are added.

\vspace{1mm}\noindent
\textbf{Decomposition by loss component.} Table~\ref{tab:runtime_losses}
separates the two loss terms on VGGn. $\Lastar$ is much cheaper than the mixed
objective, so most of the cost comes from the activation term $\Lae$. This is
because of how $\Lae$ is applied rather than how it is computed. Following~\cite{attendexcite},
we run up to $20$ inner iterations at three of the denoising steps until every
token reaches its activation target, and each iteration costs another forward and
backward pass through the U-Net. $\Lastar$ instead applies a single latent update
per optimized step. Since the cost sits in only a few steps, it can be lowered
without retraining by reducing $T_{\max}$ or the number of inner iterations.

\section{Discussion}
\label{sec:discussion}

\vspace{1mm} \noindent
\textbf{Why does the geometric framing explain the results?}
The ablation results closely follow the three geometric properties introduced in Section~$3$. CLAP produces more compact clusters than BEATs at both levels of the clustering analysis (Table~$1$ and Figure~$4$), and this directly matches the improvement in AIC across both single-audio and compositional benchmarks. 

\noindent
Angular alignment with the text manifold also explains why the larger L2 distance of CLAP tokens does not hurt performance. Although CLAP tokens are farther from text embeddings in Euclidean distance, their angular alignment remains strong, which is what matters for cross-attention. 

\noindent
Finally, inter-token spatial separation controlled by $\Lastar$ improves metrics related to clean object separation, such as CLS and CCS. The ablation results confirm this behavior. Overall, the geometric framing yields clear, testable predictions that are consistently supported by the experiments.

\vspace{1mm} \noindent
\textbf{Why is the CLAP advantage smaller on IS3?}
The reversal on IS3 comes from how the two encoders are trained rather than from
the size of the embedder's training vocabulary. CLAP is optimized contrastively
to align each clip with a fixed set of language concepts and compresses it into a
single global embedding. This is what gives CLAP its tight, well-separated
clusters and its strong in-distribution performance, but the alignment only holds
for concepts seen during pretraining. When an IS3 clip belongs to a class outside
that vocabulary, its embedding drifts toward the nearest familiar concept, and the
projected token loses the discriminative detail needed to render the correct
object. 

\noindent
BEATs is instead trained by masked acoustic-token prediction over the
broad AudioSet distribution, with no commitment to any text vocabulary, and it
exposes a frame-level feature sequence rather than one compressed vector. Its
representations vary smoothly with acoustic content and degrade more gracefully on
unseen classes, which gives the embedder more grounded signal to map an
out-of-distribution sound onto a still-recognizable token. The CLAP advantage is
therefore largest when the test audio lies on the learned manifold and shrinks
once it moves off, as on IS3.

\vspace{1mm} \noindent
\textbf{Why does performance drop at larger $n$?}
As the number of audio sources increases ($n=4$ or $5$), the cross-attention at each spatial location must be shared across more tokens. Even when tokens are well separated in CLIP space, some tokens may still receive insufficient attention during generation. 

\noindent
The A$^{*}$ loss helps reduce overlap by spreading attention maps across different regions, but it cannot increase the total available attention. Higher image resolutions or explicit control of per-token attention allocation may help address this issue in future work.

\vspace{1mm} \noindent
\textbf{Why are CLS and CCS not always the highest for our method?}
CLS and CCS depend on BLIP-generated captions, which summarize the entire image into a single sentence. As scene complexity increases, these captions become less reliable at correctly describing every individual object. This can lower CLS and CCS scores even when the generated objects are visually accurate.

\noindent
Methods that generate simpler scenes may therefore obtain artificially better scores on these metrics. For this reason, we consider AIC the most reliable metric for compositional generation, as it directly assesses whether each object class is correctly recognized in the generated image.

\end{document}